\documentclass[twoside,11pt]{article}
\usepackage{amsmath,amsfonts,bm}

\def\eqref#1{equation~\ref{#1}}

\def\1{\bm{1}}

\def\va{{\bm{a}}}
\def\vb{{\bm{b}}}

\def\vd{{\bm{d}}}

\def\vg{{\bm{g}}}

\def\vm{{\bm{m}}}

\def\vu{{\bm{u}}}
\def\vv{{\bm{v}}}
\def\vw{{\bm{w}}}

\def\vz{{\bm{z}}}

\DeclareMathAlphabet{\mathsfit}{\encodingdefault}{\sfdefault}{m}{sl}
\SetMathAlphabet{\mathsfit}{bold}{\encodingdefault}{\sfdefault}{bx}{n}

\usepackage[preprint]{jmlr2e}

\usepackage{silence}
\usepackage[T1]{fontenc} 
\usepackage{algorithm}
\usepackage[noend]{algpseudocode}
\algrenewcommand\algorithmicindent{0.7em}
\makeatletter
\AtBeginDocument{%
  \providecommand{\theHALG@line}{}%
  \renewcommand{\theHALG@line}{\thealgorithm.\arabic{ALG@line}}%
}
\makeatother
\usepackage{xfrac}
\usepackage{subcaption}
\usepackage{wrapfig}
\usepackage{anyfontsize}
\usepackage{cleveref}

\allowdisplaybreaks
\usepackage{enumitem}
\usepackage{array,booktabs,multirow}
\usepackage{tikz}
\usetikzlibrary{arrows.meta,positioning,fit,calc}
\bibpunct{(}{)}{;}{a}{}{,}
\definecolor{Ink}{HTML}{202124}
\definecolor{Muted}{HTML}{5F6368}
\definecolor{Rule}{HTML}{D6D9DE}
\definecolor{BaseBG}{HTML}{F1F3F4}
\definecolor{BaseEdge}{HTML}{7A7F85}
\definecolor{SignalEdge}{HTML}{0072B2}
\definecolor{RepresentationEdge}{HTML}{009E73}
\definecolor{CreditEdge}{HTML}{E69F00}
\definecolor{UpdateEdge}{HTML}{CC79A7}
\colorlet{SignalBG}{SignalEdge!10!white}
\colorlet{RepresentationBG}{RepresentationEdge!10!white}
\colorlet{CreditBG}{CreditEdge!10!white}
\colorlet{UpdateBG}{UpdateEdge!10!white}
\hypersetup{colorlinks=true,linkcolor=black,urlcolor=black, citecolor=black}
\setlist{nosep,leftmargin=*}
\color{Ink}

\usepackage{lastpage}

\ShortHeadings{}{}

\firstpageno{1}

\begin{document}
\title{\resizebox{\textwidth}{!}{\LARGE\bfseries Streaming Deep Reinforcement Learning Finally Works}}
\author{\normalfont
\begin{tabular*}{\textwidth}{@{}l@{\extracolsep{\fill}}r@{}}
\textbf{Mohamed Elsayed}\textsuperscript{1,2} & {\small mohamedelsayed@ualberta.ca} \\[2pt]
\textbf{Elena Sorina Lupu}\textsuperscript{4} & {\small sorina.lupu@openmindresearch.org} \\[2pt]
\textbf{Gautham Vasan}\textsuperscript{1,2} & {\small vasan@ualberta.ca} \\[2pt]
\textbf{A.\ Rupam Mahmood}\textsuperscript{1,2,3} & {\small armahmood@ualberta.ca}
\end{tabular*}
\endgraf\vspace{8pt}
{\normalfont\small
\textsuperscript{1}\,Department of Computing Science, University of Alberta\qquad \textsuperscript{3}\,Canada CIFAR AI Chair\\[1pt]
\textsuperscript{2}\,Alberta Machine Intelligence Institute (Amii)\qquad \textsuperscript{4}\,Openmind Research Institute\endgraf}
}

\begingroup
\colorlet{purple}{RepresentationEdge}
\colorlet{brown}{RepresentationEdge}
\colorlet{blue}{SignalEdge}
\colorlet{orange}{SignalEdge}
\colorlet{teal}{UpdateEdge}
\colorlet{gray}{Muted}

\thispagestyle{empty}
\maketitle

\begin{abstract}%
Learning from a stream of experience as it arrives, also known as streaming learning, is a core part of natural learning.
However, reliable streaming learning has remained a persistent challenge in modern deep reinforcement learning (RL).
Instead, most deep RL algorithms learn from old experience by storing past interactions in a buffer---an approach that can be restrictive in resource-constrained and privacy-sensitive applications.
We show that both classical streaming RL, such as Q-learning and actor-critic, when used with deep neural networks, and batch deep RL, such as PPO, SAC, and DQN, when adapted to the streaming setting, often fail to learn.
Across $58$ Atari games and $50$ continuous-control tasks, we find that these methods, in aggregate, perform close to random policies despite extensive task-specific hyperparameter searches.
We call this pattern \emph{stream barrier}.
Here, we introduce Stream-X, a shared recipe for streaming deep RL algorithms that combines signal normalization, representation stabilization, and controlled parameter updates. 
By applying Stream-X to several base streaming RL algorithms, we provide the first family of deep RL algorithms to overcome the stream barrier. 
Using one prescribed hyperparameter configuration per algorithm across tasks, Stream-X substantially improves aggregate performance, often on par with batch RL algorithms.
Beyond these benchmarks, we demonstrate learning with Stream-X algorithms under nonstationarity and resource constraints.
Stream-AC, one of the Stream-X algorithms, repeatedly recovers performance across alternating floor-friction regimes in simulation, outperforming the evaluated PPO and SAC baselines.
It also learns a heading tracking task on a robot using proprioceptive and visual features from the on-board camera in a naturally changing laboratory environment.
Stream-Q learns a Pong game from pixels directly on an ESP32-S3 microcontroller, a device with limited compute and memory.
Together, these results demonstrate that Stream-X achieves effective learning without storing past transitions and enables adaptation and on-device learning.\footnote{Code is available at \url{https://github.com/mohmdelsayed/streaming-drl}}
\end{abstract}

\section{Introduction}
A long-term vision of Artificial Intelligence (AI) is to build agents that continue to learn indefinitely as they interact with their environments~\citep{kudithipudi2022biological}. 
A household robot, a personal assistant, or a planetary rover cannot encounter all the useful experiences before deployment, nor can it assume that the world remains fixed once deployed.
Such agents must learn from live interactions as they arrive, under limits on memory, computation, and communication.
Reinforcement learning (RL), inspired by natural learning from live interactions \citep{sutton2018rl,schultz1997neural}, is a framework suited to this setting.
Modern deep reinforcement learning, which combines reinforcement learning with deep neural networks, has enabled many successes in artificial intelligence, including game-playing agents \citep{silver2017go,berner2019dota}, scientific control systems \citep{degrave2022tokamak,bellemare2020balloons}, and post-training of large language models \citep{guo2025deepseek}.
In all these applications, deep reinforcement learning follows a specific training procedure: past interaction transitions are stored in a buffer, and network weights are updated using batches of these transitions.
This batch-learning procedure fits the broader machinery of deep learning, where batch updates exploit dense matrix multiplications on large GPU clusters.
However, in all these settings, training completes before deployment.
The agent learns from pre-collected or synthetic interactions generated in simulation, rather than from the live stream of experience in the real-world setting where it ultimately operates.

Applying batch learning to learn from real-world live interactions exposes three limitations and points to an alternative, which we call \textit{streaming learning}.
First, in batch learning, interaction samples are not used when they arrive; they are placed in a buffer, and their influence is delayed until they are sampled in a later batch.
Second, batch learning requires storing past samples, which can be large in practical deep RL, where observations often include high-dimensional sensory data such as images.
This can be difficult to facilitate when agents are operating under limited memory, computation, communication, or privacy constraints, such as on edge devices or embodied robots.
Third, batch training departs from models of biological memory, which store transformed representations of experience rather than literal copies \citep{bartlett1932remembering, neisser1967cognitive, hinton1986distributed, nagy2025adaptive}.
In streaming learning, by contrast, the agent learns from each interaction as it arrives, without storing raw interactions (\Cref{fig:figure1}a).
Early RL algorithms such as Temporal Difference (TD), Q-learning, and SARSA were designed to learn this way.
Similarly, adaptive control systems~\citep{slotine1991applied} update their parameters online, using each new interaction to continuously refine the controller.
Streaming learning is natural and well-motivated, yet its realization with deep networks, \textit{streaming deep RL}, was found to be unstable in early deep RL work \citep{mnih2015human}.
We call this phenomenon \textit{stream barrier}; overcoming it remains an open problem, which we address in this work.

We present \textit{Stream-X}, a recipe that overcomes the stream barrier in the evaluated value-based and actor-critic algorithms with a single shared hyperparameter configuration per algorithm.
En route, we show that the stream barrier is pervasive across the evaluated deep RL algorithms.
Our demonstration of stream barrier spans both deep RL algorithms and deep learning versions of classical streaming RL algorithms.
Across the RL benchmarks studied, when the evaluated classical streaming RL algorithms are run with deep networks or the evaluated deep RL algorithms are run without batch updates or buffers, they show substantial learning difficulties.
These results show that the evaluated deep RL algorithms rely heavily on batching and buffering, masking instabilities that emerge once these supports are removed.
To address them, we introduce Stream-X, a recipe that mitigates these instabilities by combining signal normalization, representation stabilization, and controlled parameter updates, which we apply to a varied class of RL algorithms for value-based control and actor-critic methods.
Each Stream-X algorithm uses a single prescribed hyperparameter configuration across its evaluated benchmark tasks. 

Stream-X algorithms enable two long-sought capabilities that are missing in batch RL: fast adaptation to nonstationary environments (\Cref{sec:fast_adapt_streaming_learning}) and on-device learning on resource-constrained devices (\Cref{sec:resource_constrained_learning}).
Using a nonstationary Ant task adapted from prior continual learning work, we show that the evaluated PPO and SAC baselines, which learn from buffered experience that can become stale as conditions change, fail to match Stream-AC’s sustained recovery and improvement across repeated changes in floor friction.
On a real Ant robot, Stream-AC can learn to perform heading tracking using proprioceptive and visual features in a naturally changing laboratory environment over many hours. Stream-Q, another instance of Stream-X, learns pixel-based Pong directly on an ESP32-S3 microcontroller, achieving effective learning without batching or buffering.

Overall, Stream-X revitalizes streaming deep RL as a natural learning regime by resolving a long-standing obstacle and brings fast adaptation and on-device learning within reach.

\section{Stream barrier in deep reinforcement learning}

We define stream barrier as the phenomenon in which deep RL algorithms fail to learn stably, even after hyperparameter tuning, when learning directly from a stream of experience without storing raw interactions.
Prior work contains hints of this phenomenon. 
The most prominent evidence is provided in the work that popularized modern deep RL and introduced the Deep Q-Network (DQN) algorithm \citep{mnih2015human}.
The authors observed that when DQN is used in a streaming manner, it ``could get stuck in a poor local minimum, or even diverge catastrophically,'' whereas batch learning led to ``smoothing out learning and avoiding oscillations or divergence.''
Instead of overcoming the instability of streaming learning, modern deep RL largely bypasses it through batch learning.
Deep RL took off partly due to the massive parallelization capability of batch learning but left streaming learning largely unexplored.

\begin{figure*}[ht]
    \centering
    \includegraphics[width=\textwidth]{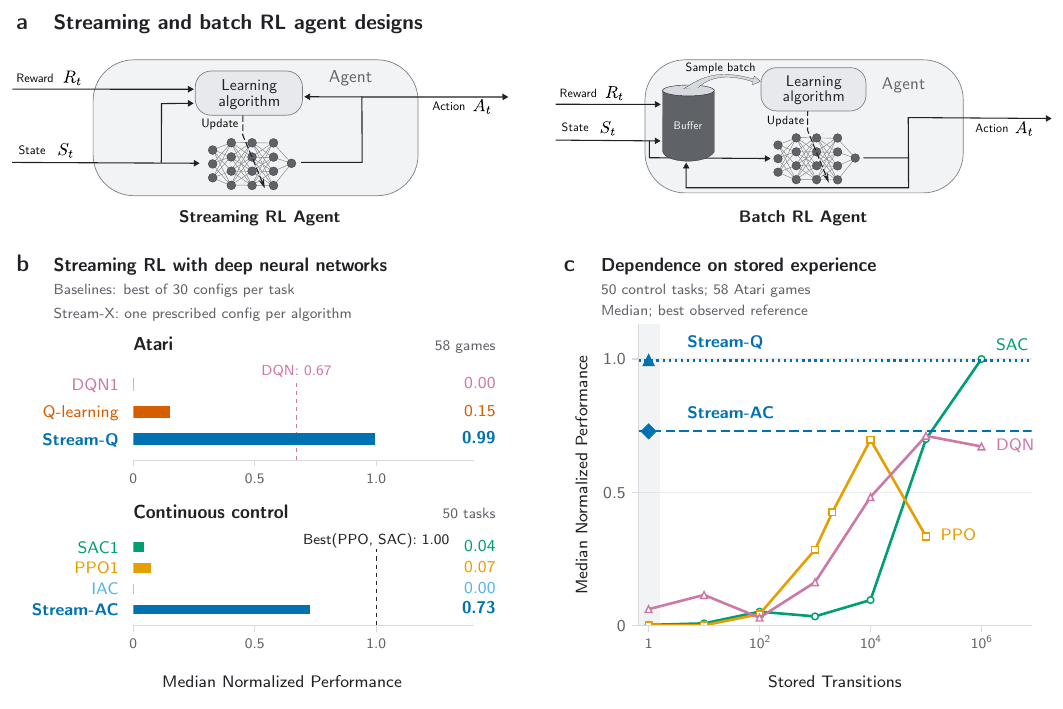}
\caption{\textbf{Stream barrier across algorithms and benchmarks.}
    \textbf{(a)} Streaming and batch RL agent designs. The streaming agent updates its network from each incoming interaction; the batch agent stores interactions in a buffer and updates its network from sampled batches.
    \textbf{(b)} Median normalized performance over $58$ discrete-control Atari games (DQN1, Q-learning, Stream-Q) or $50$ continuous-control tasks (SAC1, PPO1, IAC, Stream-AC). Each Atari agent is trained for $200$M frames and each continuous-control agent is trained for $5$M steps. The suffix `1' denotes a one-transition buffer. Returns are averaged across runs, normalized per task, then aggregated by the median. Zero denotes random-policy performance or worse. One denotes the best average return among all methods including buffer settings. Baselines use the best of $30$ configurations per task; each Stream-X algorithm uses one prescribed configuration across tasks. Blue bars identify Stream-X. Vertical dashed lines show the median score of $10^6$-transition DQN and per-task best of $10^6$-transition SAC and $2048$-transition PPO.
    \textbf{(c)} Median normalized performance versus stored transitions for DQN on the same $58$ Atari games and PPO and SAC on the same $50$ continuous-control tasks, using the same per-task references as panel b. Stored transitions denote PPO rollout length or DQN/SAC replay experience. Blue dotted and dashed lines denote buffer-free Stream-Q and Stream-AC, respectively, where the shaded strip marks the one-transition condition.
    }
    \label{fig:figure1}
\end{figure*}

We show that stream barrier is pervasive across  the evaluated RL benchmarks and across both studied value-based and policy-based algorithms (\Cref{fig:figure1}b).
We evaluate these algorithms on two main benchmark domains: Atari games \citep{bellemare2013ale} and robot control tasks from the DeepMind Control Suite \citep{tunyasuvunakool2020dmc} and MuJoCo \citep{todorov2012mujoco}.
Within these domains, we test representative modern deep RL algorithms: DQN \citep{mnih2015human} for value-based learning, tested on Atari games, and SAC \citep{haarnoja2018soft} and PPO \citep{schulman2017proximal} for policy-based algorithms, tested on robot control tasks.
These algorithms cover a range of algorithmic designs.
However, all perform batch learning over raw interactions stored in a buffer, whether it is a short rollout of recent trajectories, as in PPO, or a large replay buffer spanning hundreds of thousands to millions of steps, as in SAC and DQN.
To make these algorithms streaming, we reduce the experience buffer to one transition and enforce an immediate update: each transition is used when it is experienced and is not stored for later learning.
For each algorithm, we evaluate a shared set of hyperparameter configurations across tasks and select a configuration separately for each task. The searches vary step sizes and, for DQN1, PPO1, and SAC1, Adam parameters together with target-update frequency, the PPO clipping coefficient, or the SAC target-network averaging coefficient, respectively. Appendix~\ref{appendix:protocol} specifies the search spaces and selection procedure.

Under the evaluated best streaming configurations found with extensive tuning, these modern deep RL algorithms attain low aggregate performance.
\Cref{fig:figure1}b reports the aggregate performance of streaming DQN, SAC, and PPO, which we name DQN1, SAC1, and PPO1.
For each task, we average returns across runs and normalize relative to a random-policy baseline and the best average return among all methods and buffer settings, including Stream-X algorithms. We then take the median across the $58$ Atari games or $50$ continuous-control tasks. Scores lie in $[0,1]$ where scores at or below random are zero.
Appendix~\ref{appendix:normalization} describes the full setup.
The streaming baseline algorithms remain close to the level of a random policy.
These failures persist even after selecting each algorithm's best-performing hyperparameter configuration separately for each task from a search over $30$ configurations.
The same algorithms learn successfully once their buffers and batch updates are restored (see \Cref{fig:figure1}c), isolating the streaming setting, rather than the algorithm or benchmark, as the source of failure.
This breakdown occurs for both a value-based algorithm, DQN, and policy-based algorithms, SAC and PPO, showing that stream barrier is not confined to a single algorithm class.

We further test whether stream barrier is merely an artifact of modern deep RL algorithms reduced to streaming learning. 
We evaluate two classical streaming RL algorithms: Q-learning \citep{watkins1989} and incremental actor-critic \citep{barto1983neuronlike}, originally developed for linear function approximation and here extended to deep neural networks.
These algorithms were designed to learn from each interaction as it arrives. 
Yet, with deep networks, they too fail to learn effective policies across both domains. We add their aggregate performance to the same plot in \Cref{fig:figure1}b.
Stream barrier therefore arises across both evaluated deep RL and classical streaming algorithms, and across both value-based and policy-based methods, establishing it as a systematic phenomenon rather than a property of any single algorithm or benchmark.

To test whether stored experience is what allows modern deep RL to learn stably, we sweep the buffer size of standard deep RL algorithms, namely, PPO, SAC, and DQN (\Cref{fig:figure1}c).
If stream barrier is caused by removing stored experience, then performance should be poor at small buffer sizes and recover as algorithms regain access to larger archives of past interactions.
Across all three algorithms, median normalized performance is low at small buffer sizes and improves at larger settings, although the dependence is not always monotonic.
This shows that stream barrier is not an isolated failure at a buffer size of one: poor median performance persists across small and moderate buffers, and stable learning emerges only once the buffer is large enough.

\section{Overcoming the stream barrier with Stream-X algorithms}

We introduce our Stream-X recipe for effective streaming reinforcement learning. Because streaming deep RL agents must process one sample at a time, every learning update must be stable. That is, the agent must withstand the instabilities masked by batching and buffering.

 \begin{figure}[ht]
\centering
\begin{tikzpicture}[
  x=1mm, y=1mm,
  font=\sffamily\footnotesize, text=Ink,
  every path/.style={line cap=round, line join=round},
  panellabel/.style={font=\fontfamily{cmss}\bfseries\fontsize{11}{13}\selectfont, text=Ink, anchor=north west, inner sep=0pt},
  paneltitle/.style={font=\sffamily\bfseries\footnotesize, text=Ink, anchor=base west, inner sep=0pt},
  boxlabel/.style={font=\sffamily\bfseries\scriptsize, text=Muted, inner sep=0pt},
  note/.style={font=\sffamily\scriptsize, text=Muted, inner sep=0pt},
  container/.style={draw=BaseEdge, line width=0.6pt, rounded corners=1.8mm, fill=white},
  proc/.style={draw=BaseEdge, line width=0.6pt, fill=BaseBG, rounded corners=1.2mm,
               align=center, inner sep=1.4mm},
  stage/.style={proc, fill=white, font=\sffamily\fontsize{6.6}{7.5}\selectfont,
                text width=29mm, minimum height=14mm},
  unit/.style={circle, fill=BaseEdge, inner sep=0pt, minimum size=1.5mm},
  synapse/.style={draw=BaseEdge!55, line width=0.28pt},
  flow/.style={-{Latex[length=1.9mm,width=1.5mm]}, line width=0.7pt, draw=Muted},
  paramflow/.style={-{Latex[length=1.7mm,width=1.4mm]}, line width=0.7pt,
                    draw=RepresentationEdge, dash pattern=on 1mm off 0.8mm},
  ing/.style={-{Latex[length=1.7mm,width=1.4mm]}, line width=0.7pt, draw=#1},
  card/.style 2 args={draw=#1, fill=#2, line width=0.6pt, rounded corners=1.4mm,
                      text width=40mm, align=center, inner sep=2mm, minimum height=22mm}
]

\begin{scope}[yshift=51mm]
\node[panellabel] at (0,-1.5) {a};
\node[paneltitle] at (6.5,-3.5) {The base streaming learning algorithm};
\begin{scope}[every path/.style={}]
\clip (0,-36) rectangle (6,-7);
\node[container, fill=BaseBG!35, rounded corners=1.2mm, anchor=south west,
      minimum width=10mm, minimum height=27mm]
  (env) at (-5,-35) {};
\node[font=\sffamily\bfseries\scriptsize, rotate=90] at (2.5,-21.5) {Env};
\end{scope}
\node[container, fill=BaseBG!35, anchor=south west, minimum width=126mm, minimum height=27mm]
  (agent) at (24,-35) {};
\node[boxlabel, fill=BaseBG!35, inner xsep=1.4mm, inner ysep=0.4mm] at (34,-8) {Agent};

\node[stage, text width=26mm] (s1) at (39.5,-21)
  {{\bfseries\fontsize{7.3}{8.1}\selectfont Observe Transition}\\[1pt]
   \mbox{$S_t,A_t,R_{t+1},S_{t+1}$}};
\node[stage, text width=20mm] (s2) at (68.5,-21)
  {{\bfseries\fontsize{7.3}{8.1}\selectfont Use Networks}\\[1pt]
   \mbox{e.g., $\hat{v}$, $\hat{q}$, and $\pi$}};
\node[stage, text width=28mm] (s3) at (99.5,-21)
  {{\bfseries\fontsize{7.3}{8.1}\selectfont\mbox{Compute the increment}}\\[1pt]
    e.g., $\vu_t = \delta \vz_t$, with\\
    $\vz_t=\gamma\lambda \vz_{t-1}+\nabla_{\vw_t} \hat{q}$
  };
\node[stage, text width=28mm] (s4) at (133.3,-21)
  {{\bfseries\fontsize{7.3}{8.1}\selectfont Update Parameters}\\[1pt]
   e.g., \mbox{$\vw_{t+1}=\vw_{t}+\Delta,$}\\
   with \mbox{$\Delta=\alpha\,\mathrm{Opt}(\vu_t)$}
   };
\draw[flow] (s1) -- (s2);
\draw[flow] (s2) -- (s3);
\draw[flow] (s3) -- (s4);
\foreach \s/\c in {s1/SignalEdge, s2/RepresentationEdge, s3/CreditEdge, s4/UpdateEdge}
  \node[fill=\c, rounded corners=0.7mm, minimum width=8mm, minimum height=1.4mm,
        inner sep=0pt] at ([xshift=9mm]\s.north west) {};
\draw[flow, rounded corners=1.5mm] (s4.north) -- ++(0,3.5) -| (s1.north);
\node[note, fill=BaseBG!35, inner xsep=1mm, inner ysep=0.5mm] at (87,-10.5) {next time step};

\draw[flow] (5.4,-17.5) -- ([yshift=3mm]s1.west);
\node[note, font=\sffamily\fontsize{6.4}{7.4}\selectfont,
  fill=white, inner xsep=0.7mm, inner ysep=0.3mm] at (15.2,-17.5)
  {$S_{t+1},\,R_{t+1}$};
\draw[flow] (24,-25.5) -- (5.4,-25.5);
\node[note, font=\sffamily\fontsize{6.4}{7.4}\selectfont,
  fill=white, inner xsep=0.7mm, inner ysep=0.3mm] at (14.7,-25.5) {$A_t$};

\begin{scope}[yshift=-10mm]
\node[panellabel] at (0,-31.5) {b};
\node[card={SignalEdge}{SignalBG}, minimum height=22mm]                 (sig)  at (22,-52) {};
\node[card={RepresentationEdge}{RepresentationBG}, minimum height=22mm] (rep)  at (75,-52) {};
\node[card={UpdateEdge}{UpdateBG}, minimum height=22mm]                 (ctrl) at (128,-52) {};

\foreach \i in {1,...,3} \coordinate (ra\i) at ({57.5},{-45.5+(\i-2)*2.5});
\foreach \i in {1,...,4} \coordinate (rb\i) at ({63},{-45.5+(\i-2.5)*1.7});
\foreach \i in {1,...,3} \coordinate (rc\i) at ({68.5},{-45.5+(\i-2)*2.5});
\foreach \i/\j in {1/1,1/3,2/2,2/4,3/1,3/4}
  \draw[draw=RepresentationEdge, line width=0.35pt] (ra\i) -- (rb\j);
\foreach \i/\j in {1/1,2/2,3/1,4/3}
  \draw[draw=RepresentationEdge, line width=0.35pt] (rb\i) -- (rc\j);
\foreach \i in {1,...,3} \fill[RepresentationEdge] (ra\i) circle (0.55);
\foreach \i in {1,...,4} \fill[RepresentationEdge] (rb\i) circle (0.55);
\foreach \i in {1,...,3} \fill[RepresentationEdge] (rc\i) circle (0.55);

\draw[draw=RepresentationEdge!25!white, line width=0.4pt]
  (72,-42.4) -- (72,-48.6);
\draw[draw=Muted!45!white, line width=0.4pt] (74.5,-46.3) -- (81,-46.3);
\draw[draw=RepresentationEdge!45!white, line width=0.4pt,
      dash pattern=on 0.6mm off 0.5mm] (74.5,-44.74) -- (81,-44.74);
\foreach \position/\activation in {75.5/-0.4,77/2,78.5/0.4,80/2.8} {
  \draw[draw=Muted!85!white, line width=0.9pt]
    (\position,-46.3) -- (\position,{-46.3+1.3*\activation});
  \fill[Muted] (\position,{-46.3+1.3*\activation}) circle (0.4);
}
\draw[ing=RepresentationEdge, line width=0.7pt] (82,-46.3) -- (86,-46.3);
\node[font=\sffamily\bfseries\fontsize{6.2}{7}\selectfont,
      text=RepresentationEdge!80!Ink, inner sep=0pt] at (84,-43.7) {LN};
\draw[draw=Muted!45!white, line width=0.4pt]
  (87.5,-46.3) -- (94,-46.3);
\foreach \position/\activation in {88.5/-0.4,90/2,91.5/0.4,93/2.8} {
  \pgfmathsetmacro{\normalized}{1.3*(\activation-1.2)/sqrt(1.6)}
  \draw[draw=RepresentationEdge, line width=0.9pt]
    (\position,-46.3) -- (\position,{-46.3+\normalized});
  \fill[RepresentationEdge] (\position,{-46.3+\normalized}) circle (0.4);
}

\draw[draw=Rule, line width=0.4pt] (10,-48.3) -- (35,-48.3);
\draw[draw=SignalEdge, line width=0.6pt, fill=SignalBG]
  (11,-48.3) .. controls (14,-48.3) and (13,-45.5) .. (16,-45.5)
  .. controls (19,-45.5) and (18,-48.3) .. (21,-48.3) -- cycle;
\draw[draw=SignalEdge, line width=0.6pt, fill=SignalBG]
  (28,-48.3) .. controls (30,-48.3) and (29.5,-43.1) .. (31,-43.1)
  .. controls (32.5,-43.1) and (32,-48.3) .. (34,-48.3) -- cycle;
\draw[ing=SignalEdge] (22.5,-46.3) -- (26.6,-46.3);

\begin{scope}[xshift=-0.5mm]
\draw[draw=Rule, line width=0.4pt] (129.7,-45.5) -- (145.3,-45.5);
\foreach \boundheight in {-42.8,-48.2}
  \draw[draw=UpdateEdge!60!white, line width=0.35pt, dash pattern=on 0.8mm off 0.6mm]
    (129.7,\boundheight) -- (145.3,\boundheight);
\node[note, anchor=west, font=\sffamily\fontsize{5.5}{6.2}\selectfont]
  at (145.8,-42.8) {$\alpha$};
\node[note, anchor=west, font=\sffamily\fontsize{5.5}{6.2}\selectfont]
  at (145.8,-48.2) {$-\alpha$};
\draw[ing=UpdateEdge, line width=0.65pt] (125,-45.5) -- (128.5,-45.5);
\def\controlincrements{109/2.4,111/-1.3,113/5.0,115/-2.8,117/1.5,119/-4.2,121/1.2,123/-1.6}
\def\runningmax{0}
\foreach \position/\increment in \controlincrements {
  \pgfmathsetmacro{\magnitude}{abs(\increment)}
  \pgfmathsetmacro{\nextmax}{max(0.82*\runningmax,\magnitude)}
  \ifnum\position>109
    \path[fill=UpdateEdge!17!white]
      plot[domain=0:1, samples=12, variable=\fraction]
        ({\position-2+2*\fraction},{-45.5+0.54*\runningmax*pow(0.82,\fraction)})
      -- plot[domain=1:0, samples=12, variable=\fraction]
        ({\position-2+2*\fraction},{-45.5-0.54*\runningmax*pow(0.82,\fraction)})
      -- cycle;
    \foreach \side in {-1,1}
      \draw[draw=UpdateEdge!80!white, line width=0.6pt]
        plot[domain=0:1, samples=12, variable=\fraction]
          ({\position-2+2*\fraction},{-45.5+\side*0.54*\runningmax*pow(0.82,\fraction)})
        -- (\position,{-45.5+\side*0.54*\nextmax});
  \fi
  \xdef\runningmax{\nextmax}
}
\draw[draw=Rule, line width=0.4pt] (108.2,-45.5) -- (123.8,-45.5);
\def\runningmax{0}
\foreach \position/\increment in \controlincrements {
  \pgfmathsetmacro{\nextmax}{max(0.82*\runningmax,abs(\increment))}
  \draw[draw=Muted, line width=0.75pt]
    (\position,-45.5) -- (\position,{-45.5+0.54*\increment});
  \draw[draw=UpdateEdge!85!Ink, line width=0.75pt]
    ({\position+21.5},-45.5) -- ({\position+21.5},{-45.5+2.7*\increment/(\nextmax+0.00000001)});
  \xdef\runningmax{\nextmax}
}
\end{scope}

\node[anchor=north, text width=40mm, align=center] at (22,-50.3)
  {{\bfseries\color{Ink}Normalize signals}\\[1pt]
  {\scriptsize observation normalization\\reward scaling}};
\node[anchor=north, text width=40mm, align=center] at (75,-50.3)
  {{\bfseries\color{Ink}Stabilize representations}\\[1pt]
  {\scriptsize sparse initialization\\layer normalization}};
\node[anchor=north, text width=40mm, align=center] at (128,-50.3)
  {{\bfseries\color{Ink}Control the update}\\[1pt]
  {\scriptsize StreamingOptimizer}};
\draw[ing=SignalEdge, rounded corners=1.5mm] (sig.north) -- ++(0,2.5) -| (s1.south);
\draw[ing=RepresentationEdge, rounded corners=1.5mm] (rep.north) -- ++(0,2.5) -| (s2.south);
\draw[ing=UpdateEdge, rounded corners=1.5mm] (ctrl.north) -- ++(0,2.5) -| (s4.south);
\node[paneltitle, fill=white, inner xsep=1.2mm, inner ysep=0.8mm] at (6.5,-34.5)
  {Stream-X recipe};
\end{scope}
\end{scope}
\end{tikzpicture}
\caption{\textbf{The base streaming learning algorithm and the Stream-X recipe.}
\textbf{(a)} The streaming base shared by all Stream-X algorithms. The agent takes action $A_t$ in observation $S_t$, receives $S_{t+1}$ and $R_{t+1}$, and evaluates its value and, where applicable, policy networks to compute the learning update. The increment $\vu_t$ is used to update the parameters. With $\mathrm{Opt}$ the identity and without the Stream-X additions, this gives the streaming base algorithms.
 \textbf{(b)} Stream-X has three additions: stabilize representations, normalize signals, and control each update. Connections identify the stage on which each addition acts.
}
\label{fig:Stream-X-skeleton}
\end{figure}

The Stream-X recipe places a number of ingredients around a streaming learning base algorithm. This base algorithm computes an increment $\vu_t \in \mathbb{R}^d$ based on the current transition and updates the parameters by $\vw\leftarrow \vw+\alpha\,\mathrm{Opt}(\vu_t)$, where $\alpha>0$ is the step size and $\mathrm{Opt}$ maps the increment to a parameter update (\Cref{fig:Stream-X-skeleton}a).
For the TD-error-based methods, the increment is $\vu_t=\delta_t\vz_t$: the temporal-difference error $\delta_t\in\mathbb{R}$ measures how far the current value estimate sits from its bootstrapped target, and the eligibility trace $\vz_t=\gamma\lambda\vz_{t-1}+\vg_t$ accumulates the recent gradients $\vg_t$ of that estimate, with $\gamma\in[0,1]$ the discount factor and $\lambda\in[0,1]$ setting how far credit spreads backward in time \citep{sutton1988learning,sutton2018rl}. Setting $\mathrm{Opt}$ to the identity recovers the classical streaming algorithms.
Stream-Q based on Q-learning \citep{watkins1989}, Stream-SARSA based on SARSA \citep{rummery1994sarsa}, and Stream-AC based on actor-critic \citep{barto1983neuronlike} share this base and differ in which estimate they learn and how they obtain a policy\footnote{
For example, the Q-learning base maintains a parameterized estimate of action values, $\hat{q}_\vw$, with parameters $\vw\in\mathbb{R}^d$. It computes a temporal-difference error $\delta_t = R_{t+1} + \gamma \max_{a}\hat{q}(S_{t+1}, a;\vw) - \hat{q}(S_t, A_t;\vw)$, and the gradient of the estimate at the visited state, $\vg_t=\nabla_{\vw}\hat{q}(S_t,A_t;\vw)\in\mathbb{R}^d$.
}; Stream-RAC, however, uses it for its critic, while its actor follows a reparameterized policy gradient \citep{mohamed2020monte} without eligibility traces.

The ingredients of the Stream-X recipe are signal normalization, representation stabilization, and controlled parameter updates (\Cref{fig:Stream-X-skeleton}b). We present them here in detail.

\textbf{1) Normalize the signals.} A batch update averages over samples of differing magnitudes, whereas under streaming learning each update is exposed to the raw scale of a single state and a single reward, so both the network inputs and the temporal-difference error can vary by orders of magnitude across tasks and over time within a task. Signal normalization removes task-specific scale from the learning signal.
We maintain running estimates of the mean and variance of each input coordinate, standardize the input entries $O_i, \forall i$ before it enters each network, and scale the reward $R$ using the running variance of a discounted reward trace
\begin{align}
    \bar{O}_i \doteq \frac{O_i-\mu_i}{\sqrt{\sigma_i^2+\varepsilon}}, \qquad \bar{R} \doteq \frac{R}{\sqrt{\sigma_r^2+\varepsilon}},
\end{align}
where $\mu_i$ and $\sigma_i^2$ are the running mean and variance of the $i$th input coordinate, $\sigma_r^2$ is the running variance of the discounted reward trace $U_t\doteq\gamma U_{t-1}+r_t$, which is reset at episode boundaries, and $\varepsilon$ is a small number used for numerical stability (see {\color{blue}NormalizeObservation} (\Cref{alg:obs-norm}) and {\color{orange}ScaleReward} (\Cref{alg:reward-norm})).
Scaling by the return would be more natural, as the temporal-difference target is governed by the scale of discounted returns rather than of one-step rewards, but the return depends on future rewards and is therefore unavailable to a streaming update.
The reward trace is a causal counterpart: 
the two have the same variance under some assumptions (see \Cref{appendix:var-equivalence}). 

\textbf{2) Stabilize the representations.} When an agent learns from a temporally correlated stream of experience, its ability to learn can degrade as the network representation distribution shifts \citep{lyle2025disentangling}. Layer normalization \citep{ba2016layer} is one technique for giving each layer's pre-activations similar distributions. We apply layer normalization before each hidden-layer activation, standardizing the pre-activation vector $\va$ entries by subtracting their mean and dividing by their standard deviation. Specifically, the {\color{purple} LayerNorm} normalization $\phi$ we use is given by
\begin{align}
    \phi(\va) &\doteq \frac{\va-\mu}{\sqrt{\sigma^2+\varepsilon}}, \quad \text{where } \mu\doteq \frac{1}{n} \sum_{i=1}^{n} a_i \text{  and  } \sigma^2\doteq \frac{1}{n} \sum_{i=1}^{n} (a_i-\mu)^2,
\end{align}
where $n$ is the dimensionality of $\va$ and $\epsilon$ is a small number used for numerical stability.

Furthermore, we initialize linear and convolutional layers sparsely, so that each unit begins with a limited number of nonzero incoming connections (see {\color{brown}SparseInit} (\Cref{alg:sparse_init})). Sparse connectivity localizes each update, which can reduce interference between dissimilar inputs at early time steps.

\textbf{3) Control the update.} 
Without batching, every transition acts directly on the parameters, so an unusually large increment is not averaged with other samples and can dominate the update. We introduce a new optimizer, {\color{teal}\textit{StreamingOptimizer}}, which replaces $\mathrm{Opt}$ from the base algorithm and controls the size of every parameter update.
For each parameter vector, StreamingOptimizer maintains an element-wise decaying running maximum of the raw increment, initialized to zero,
\begin{align}
\vv_t=\max\!\left(\beta \vv_{t-1},\lvert \vu_t \rvert\right),\qquad
\vw_{t+1}=\vw_t+\alpha\frac{\vu_t}{\vv_t+\varepsilon},
\end{align}
where $\beta\in[0,1)$ is the retained fraction, $\alpha>0$ is the step size, $\epsilon>0$ a small constant for numerical stability, and all operations are element-wise.
Because $v_{t,i}\geq\lvert u_{t,i} \rvert$ for every coordinate $i$, the per-coordinate displacement is bounded, $\lvert\Delta_i \rvert<\alpha$, independent of the magnitude of $u_{t,i}$.
The maximum responds immediately to a large increment and relaxes through the decay, limiting the spike itself and reducing subsequent updates relative to this recent scale.
We analyze these properties and contrast StreamingOptimizer with RMSProp in the~\Cref{appendix:streaming-optimizer}.

Together, these components allow Stream-X algorithms to succeed across diverse environments under a \textit{single} prescribed hyperparameter configuration per algorithm without any task-specific tuning.
We focus in the main text on Stream-AC and Stream-Q and leave the Stream-SARSA and Stream-RAC results in the Appendix. Appendix~\ref{appendix:pseudocodes} gives the complete pseudocode of the four Stream-X algorithms: Stream-AC (Algorithm~\ref{alg:stream-ac}), Stream-Q (Algorithm~\ref{alg:stream-q}), Stream-RAC (Algorithm~\ref{alg:stream-rac}), and Stream-SARSA (Algorithm~\ref{alg:stream-sarsa}), showing where each ingredient enters the base algorithm.

\begin{wrapfigure}{r}{0.42\textwidth}
  \centering
  \includegraphics[width=0.42\textwidth]{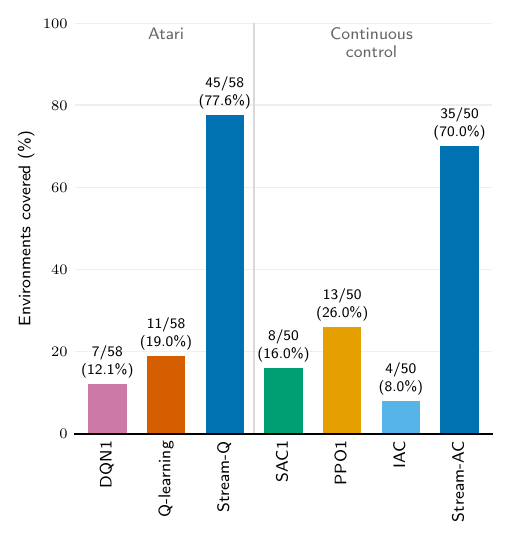}
  \caption{\textbf{Task coverage with a single hyperparameter configuration.} 
  }
  \label{fig:single-config-coverage}
\end{wrapfigure}
To quantify this breadth, we compare the fraction of tasks covered by a single configuration at matched training budgets (\Cref{fig:single-config-coverage}). 
Training budgets are $12.5$ million frames on Atari and $2$ million steps on continuous control. 
A task is considered covered when its normalized score reaches at least $0.25$, corresponding to one quarter of the improvement from random to the best observed return in this comparison. 
For each streaming baseline, we select the one configuration among the $30$ tested that covers the most tasks. Stream-X algorithms use their single prescribed configuration. 
Normalization uses the random-policy return and the best observed mean return across all configurations and methods in this matched-budget comparison, including batch DQN on Atari and batch SAC and PPO on continuous control.
Unlike \Cref{fig:figure1}b, this comparison selects one configuration per algorithm across tasks rather than separately for each task.
We observe that Stream-AC and Stream-Q can cover most of the environments given this training budget and coverage definition.
\vspace{0.5cm}

\begingroup
\setlength{\intextsep}{4pt}
\setlength{\floatsep}{4pt}
\begin{algorithm}[H]
\footnotesize
\caption{Stream-Q}\label{alg:stream-q}
\begin{algorithmic}[1]
\State \textbf{Given} Action-value network $\hat{q}(s, a;\vw)$ with {\color{purple} LayerNorm} and weights $\vw$  initialized with {\color{brown} SparseInit}
\State \textbf{Initialize} 
\State \quad Discount factor $\gamma$ and eligibility traces parameter $\lambda$ 
\State \quad Exploration schedule $\epsilon_\text{start}$, $\epsilon_\text{end}$, exploration fraction $f$, and budget of $N$ time steps 
\State \quad Statistics $\mu_S, \mu_r, p_S, p_r$, reward trace $U$, time $t$, and small number $\varepsilon$
\State \quad Optimizer state $\vv_{\vw}$, step size $\alpha$, and retained fraction $\beta$
\State \quad Exploration probability $\epsilon\leftarrow\epsilon_\text{start}$
\For{each episode}
\State $\vz_{\vw}\leftarrow \mathbf{0}$, $U\leftarrow 0$
\State Initialize $S$ \Comment{{\color{gray} First state of the episode}}
\State {\color{blue} $S, \mu_S, p_S \leftarrow$  NormalizeObservation($S, \mu_S, p_S, t$)}
\State Choose $A$ from $S$ $\epsilon$-greedily w.r.t. $\hat{q}(S, \cdot, \vw)$
\For{each step in the episode}
\State $t \leftarrow t + 1$
\State Take action $A$, observe $S^\prime$, $R$, $T$ \Comment{{\color{gray} $T$ indicates whether $S^\prime$ is a terminal state}}
\State {\color{blue} $S^\prime, \mu_S, p_S \leftarrow$  NormalizeObservation($S^\prime, \mu_S, p_S, t$)}
\State {\color{orange} $R, U, \mu_r, p_r \leftarrow$  ScaleReward($R, \gamma, U, \mu_r, p_r, T, t$)}
\State $\epsilon \leftarrow \max\left(\epsilon_\text{start} + (\epsilon_\text{end} - \epsilon_\text{start})\frac{t}{f N},\; \epsilon_\text{end}\right)$ \Comment{{\color{gray} Linearly annealed exploration}}
\State Choose $A^\prime$ from $S^\prime$ $\epsilon$-greedily w.r.t. $\hat{q}(S^\prime, \cdot, \vw)$
\State $\delta \leftarrow R + \gamma \max_a \hat{q}(S^\prime, a, \vw) - \hat{q}(S, A, \vw)$ \Comment{{\color{gray} If $S^\prime$ is terminal, then $\hat{q}(S^\prime,\cdot, \vw)\doteq 0$}}
\State $\vz_{\vw} \leftarrow \gamma \lambda \vz_{\vw} + \nabla_{\vw} \hat{q}(S, A, \vw)$
\State {\color{teal}$\vd, \vv_{\vw} \gets \text{StreamingOptimizer}(\delta\vz_\vw, \vv_{\vw}, \beta, \varepsilon)$}
\State $\vw \gets \vw + \alpha\vd$
\If {$A^\prime \neq \arg\max_a \hat{q}(S^\prime, a, \vw)$}
\State $\vz_{\vw}\leftarrow \mathbf{0}$
\EndIf
\State $S \leftarrow S^\prime$,\; $A \leftarrow A^\prime$
\EndFor
\EndFor
\end{algorithmic}
\end{algorithm}
\begin{algorithm}[H]
\footnotesize
\caption{Stream-AC}\label{alg:stream-ac}
\begin{algorithmic}[1]
\State \textbf{Given} State-value network $\hat{v}(s;\vw)$ with {\color{purple} LayerNorm} and weights $\vw$ initialized with {\color{brown} SparseInit}, and policy network $\pi(a|s; \boldsymbol\theta)$ with {\color{purple} LayerNorm} and weights $\boldsymbol\theta$ initialized with {\color{brown} SparseInit}
\State \textbf{Initialize} 
\State \quad Discount factor $\gamma$ and eligibility traces parameter $\lambda$ 
\State \quad Statistics $\mu_S, \mu_r, p_S, p_r$, reward trace $U$, time $t$, and small number $\varepsilon$
\State \quad Optimizer state $\vv_{\vw}, \vv_{\boldsymbol\theta}$, value step size $\alpha_{\hat{v}}$, policy step size $\alpha_{\pi}$, and retained fraction $\beta$
\State \quad Entropy coefficient $\tau$
\For{each episode}
\State $\vz_{\vw}\leftarrow \mathbf{0}, \vz_{\boldsymbol\theta}\leftarrow \mathbf{0}$, $U\leftarrow 0$
\State Initialize $S$ \Comment{{\color{gray} First state of the episode}}
\State {\color{blue} $S, \mu_S, p_S \leftarrow$  NormalizeObservation($S, \mu_S, p_S, t$)}
\For{each step in the episode}
\State $t \leftarrow t + 1$
\State $A \sim \pi(\cdot|S, \boldsymbol\theta)$
\State Take action $A$, observe $S^\prime$, $R$, $T$ \Comment{{\color{gray} $T$ indicates whether $S^\prime$ is a terminal state}}
\State {\color{blue} $S^\prime, \mu_S, p_S \leftarrow$  NormalizeObservation($S^\prime, \mu_S, p_S, t$)}
\State {\color{orange} $R, U, \mu_r, p_r \leftarrow$ ScaleReward($R, \gamma, U, \mu_r, p_r, T, t$)}
\State $\delta \leftarrow R + \gamma \hat{v}(S^\prime, \vw) - \hat{v}(S, \vw)$ \Comment{{\color{gray} If $S^\prime$ is terminal, then $\hat{v}(S^\prime, \vw)\doteq 0$}}
\State $\vz_{\vw} \leftarrow \gamma \lambda \vz_{\vw} + \nabla_{\vw} \hat{v}(S,\vw)$
\State $\vz_{\boldsymbol\theta} \leftarrow \gamma \lambda \vz_{\boldsymbol\theta} + \nabla_{\boldsymbol\theta} \left(\log \pi(A|S, \boldsymbol\theta) + \tau\text{sign}(\delta) H(\cdot|S, \boldsymbol\theta)\right)$ 
\Comment{{\color{gray} $H(\cdot|S, \boldsymbol\theta)$ is the entropy}}
\State {\color{teal}$\vd_\pi, \vv_{\boldsymbol\theta} \gets \text{StreamingOptimizer}(\delta\vz_{\boldsymbol\theta}, \vv_{\boldsymbol\theta}, \beta, \varepsilon)$}
\State $\boldsymbol\theta \gets \boldsymbol\theta + \alpha_{\pi}\vd_\pi$
\State {\color{teal}$\vd_{\hat{v}}, \vv_{\vw} \gets \text{StreamingOptimizer}(\delta\vz_\vw, \vv_{\vw}, \beta, \varepsilon)$}
\State $\vw \gets \vw + \alpha_{\hat{v}}\vd_{\hat{v}}$
\State $S \leftarrow S^\prime$
\EndFor
\EndFor
\end{algorithmic}
\end{algorithm}
\endgroup

\section{Fast adaptation with streaming learning}\label{sec:fast_adapt_streaming_learning}

Streaming learning agents learn from the most recent observation without relying on a stored replay buffer of past experiences, which makes them well-suited in nonstationary environments where previously collected data can be stale, e.g., due to changes in the environment. 
We examine this adaptation hypothesis in two complementary settings: simulated locomotion with abrupt, repeated changes in floor friction, and vision-based navigation on a physical robot in a naturally changing laboratory over many hours.

\subsection{Locomotion with an alternating-friction floor in simulation}

We used an alternating-friction version of the Ant environment \citep{todorov2012mujoco}, adapted from the nonstationary locomotion setting of \citet{dohare2024lop}. The coefficient of friction between the robot's feet and the floor alternated between $2.0$ (high-traction) and $0.02$ (slippery) every $2$ million time steps, resulting in ten regimes over $20$ million steps (\Cref{fig:alternating-friction-ant}a). Each change occurred instantaneously with no explicit signal to the agent, requiring it to adapt to the changed contact dynamics from its ongoing stream of experience. 
The alternating friction therefore tests whether the agent can recover performance after repeated changes in the environment.

\begin{figure}[ht]
    \centering
    \includegraphics[width=\textwidth]{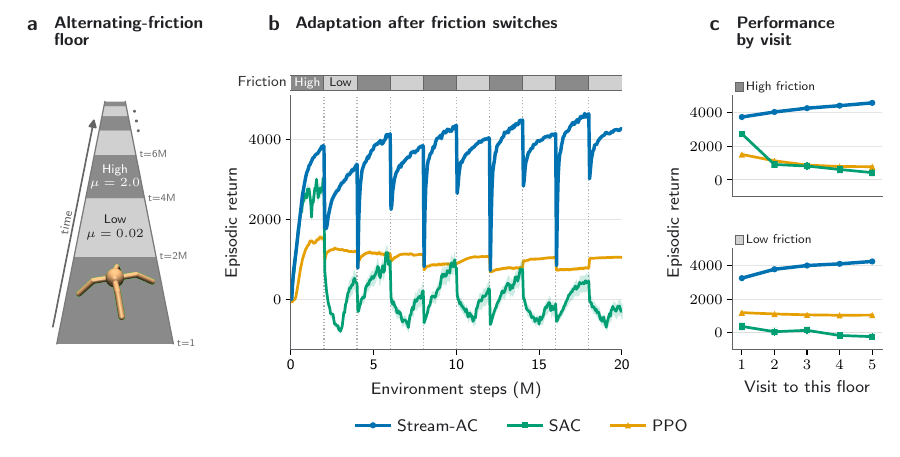}
    \caption{\textbf{Fast adaptation in a nonstationary environment.}
    \textbf{(a)} Schematic of the alternating floor friction task, where the floor friction alternates between high ($\mu=2.0$) and low ($\mu=0.02$) every $2$ million time steps.
    \textbf{(b)} Episodic return of Stream-AC, PPO, and SAC over $20$ million time steps across ten alternating regimes. The top bar indicates the current floor friction. Curves show the mean over $60$ runs, and shaded regions denote the standard error.
    \textbf{(c)} Return over the final quarter steps of each regime across successive visits to high-friction and low-friction floors.}
    \label{fig:alternating-friction-ant}
\end{figure}

We compare Stream-AC with the batch algorithms PPO and SAC in a nonstationary environment (\Cref{fig:alternating-friction-ant}b). SAC and PPO represent two standard forms of stored experience: a replay buffer of one million transitions and a rollout buffer of $2048$ transitions, respectively.

Stream-AC recovered performance across the repeated alternations (\Cref{fig:alternating-friction-ant}b). Its average return over the final quarter of each regime reached or exceeded that of the preceding visit to the same friction across all eight revisits. From the first to the fifth visit, these average returns increased by about $850$ on high-friction and $1000$ on low-friction floors (\Cref{fig:alternating-friction-ant}c). These results show improved performance across repeated exposures, without identifying whether recovery reflects retained representations, relearning, or both.

The evaluated batch baselines do not show comparable recovery. SAC learns the first high-friction regime well, but its return collapses after the first switch and remains low over the remaining training period. Immediately after a switch, most transitions in its million-transition replay buffer come from the previous regime, making replay staleness a plausible contributor to the initial disruption. That mismatch alone does not explain persistent failure after the buffer has been replenished. PPO avoids a comparable collapse but does not show Stream-AC's pattern of recovery and improvement. Its much shorter rollout limits the duration of stored-data mismatch. Loss of plasticity, documented in related settings by \citet{dohare2024lop}, is a possible explanation, but the return curves do not establish it as the cause here. A replay-capacity sweep shows that smaller buffers help SAC in some regimes, but do not restore sustained adaptation (Appendix~\ref{appendix:ant-replay-buffers}).

These results demonstrate repeated recovery by Stream-AC without explicit change detection or switch-triggered resets. They support streaming learning as a practical approach to adaptation in this setting, while leaving open how much of the advantage arises from avoiding replay, from the Stream-X ingredients, or from their interaction.

\subsection{Vision-based navigation using a legged robot in a nonstationary environment}\label{sec:vision_based_navigation}

We validated our method on a vision-based navigation task on-board the physical Open Ant robot~\citep{lupu2026openantrobotplatform}, a physical variant of the commonly used Gymnasium Ant environment.
The robot was allowed to learn this task autonomously over multiple hours in a cluttered robot laboratory environment, without any artificial modifications or scripted changes to the environment (\Cref{fig:ant-heading}a) using Stream-AC (\Cref{alg:stream-ac}). 
Consequently, the visual observations naturally varied over time as the laboratory evolved, as shown in~\Cref{fig:ant-heading}b.
For example, other robots were moved throughout the workspace, employees entered and interacted with the environment, and the appearance of the windows changed as outdoor conditions transitioned from day to night.
This setup provided a natural test of the method's ability to continue learning under the distribution shifts that arise during ordinary operation, rather than relying on manually constructed changes to the environment.

\begin{figure}[ht]
    \centering\includegraphics[width=\linewidth, height=\textheight, keepaspectratio, 
    trim=20 0 20 400, 
    clip
]{figures/images.pdf}
    \caption{\textbf{Heading tracking on the physical Open Ant robot.} Stream-AC combines a history of proprioceptive inputs with camera features extracted by a frozen encoder and processed by an MLP network.
    \textbf{(a)} The robot in the laboratory arena initialized at different orientations and positions.
    \textbf{(b)} Example images from the onboard camera showing nonstationarity.
    \textbf{(c)} Example heading tracking (actual heading from the magnetometer versus desired heading) within each episode at four different points in the learning process. The polar plots show starting, intermediate, desired, and final headings.}
    \label{fig:ant-heading}
\end{figure}

The task is a heading-tracking navigation task in which the agent was rewarded for aligning its instantaneous heading with a target direction of $0$ degrees. The task is episodic and the duration of each episode was 120 seconds. After 120 seconds, the robot was automatically reset at a random orientation and position within the space (\Cref{fig:ant-heading}a).
At each timestep, the reward was computed as the cosine similarity between the current heading vector, estimated from onboard magnetometer readings, and the desired heading vector, yielding a bounded reward in $\left[-1, 1\right]$.
The observation vector combined $24$ dimensions of proprioceptive state: eight joint positions, eight joint velocities, the desired heading vector, and six-axis inertial measurements (linear acceleration and angular velocity), with a visual feature vector derived from an onboard RGB camera. 
Visual features were extracted from a frozen, self-supervised DINO ViT-S/16 \citep{caron2021dino} vision transformer and passed through an MLP. 
The output of the MLP was flattened and concatenated with the proprioceptive information. Details on the network sizes and inputs can be seen in~\Cref{appendix:architecture}. 

\Cref{fig:ant-heading}c shows a fixed-target heading demonstration. We selected four representative points during the 11-hour learning period and plot the actual heading, measured by the on-board magnetometer, against the desired heading of $0^\circ$. Early in learning, the robot exhibits substantial tracking error. After 6 hours, tracking has improved, although a persistent offset remains. By 10 hours, the offset is nearly eliminated, demonstrating progressively more accurate heading tracking with Stream-AC.

\section{On-device streaming learning with resource-constrained devices}\label{sec:resource_constrained_learning}

\begin{figure}[ht]
  \centering
  \includegraphics[width=\textwidth]{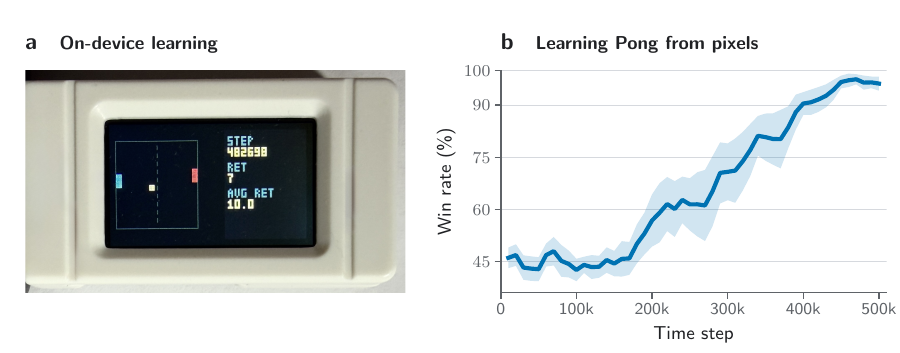}
  \vspace{-0.8cm}
  \caption{\textbf{Stream-Q learns Pong from pixels on a microcontroller.}
  \textbf{(a)} The agent plays and learns on a $4.5\,\mathrm{cm}\times 2.5\,\mathrm{cm}$ ESP32-S3 board with a $240$\,MHz processor and $512$\,KB of SRAM. Simulation, rendering, action selection, and learning all run on the device.
  \textbf{(b)} Win rate against the scripted opponent over $500{,}000$ agent steps. The agent processes two stacked $32\times 32$ image frames with a convolutional neural network and selects among three actions: no-op, up, and down. The curve is the mean over $5$ independent runs and the shaded region indicates the standard error.}
  \label{fig:esp32}
\end{figure}

We demonstrate effective on-device learning from pixels under tight memory constraints with Stream-Q (\Cref{alg:stream-q}).
Large experience buffers are impractical on such devices: a million byte-valued $32\times32$ frames require about $1$\,GB, before accounting for other transition data.
A smaller buffer may fit, with $100$ such frames occupying about $100$\,KB, but  as shown in~\Cref{fig:figure1}c, reducing memory requirements does not produce reliable learning with the evaluated batch RL methods.
Stream-Q avoids this dependence on buffering by learning directly from each incoming transition.

We used Stream-Q to solve Pong played from pixels on an ESP32-S3 microcontroller (\Cref{fig:esp32}a). The agent observes the grayscale game board as two stacked frames, chooses among three paddle actions (up, down, and no-op), and plays to $21$ points against a scripted opponent. Appendix~\ref{appendix:esp32} details the full setup. Stream-Q learns to play Pong from pixels (\Cref{fig:esp32}b). Over the first $50{,}000$ steps, the agent wins about $45\%$ of its games against the scripted opponent. Over the last $50{,}000$ steps of the $500{,}000$-step run, it wins about $97\%$. This establishes the feasibility of pixel-based deep RL on the microcontroller, without storing past transitions for learning.

\section{Ablation study}
\label{section:ablation}

We remove each Stream-X ingredient in turn, leaving the base RL algorithm, the remaining components, and the single prescribed hyperparameter configuration untouched (\Cref{fig:ablation}; all ten tasks are presented in Appendix~\ref{appendix:full-ablation}). Each ingredient addresses a distinct challenge, and its impact varies across environments. On continuous control, layer normalization and observation normalization are critical on Humanoid, while StreamingOptimizer prevents severe instability and performance collapse on the Ant environment. On Atari, Frostbite degrades severely without layer normalization, observation normalization, reward scaling, or eligibility traces. Conversely, on Enduro, almost all ablations, except for the StreamingOptimizer, achieve comparable final performance.

\begin{figure}[ht]
    \centering
    \includegraphics[width=\textwidth]{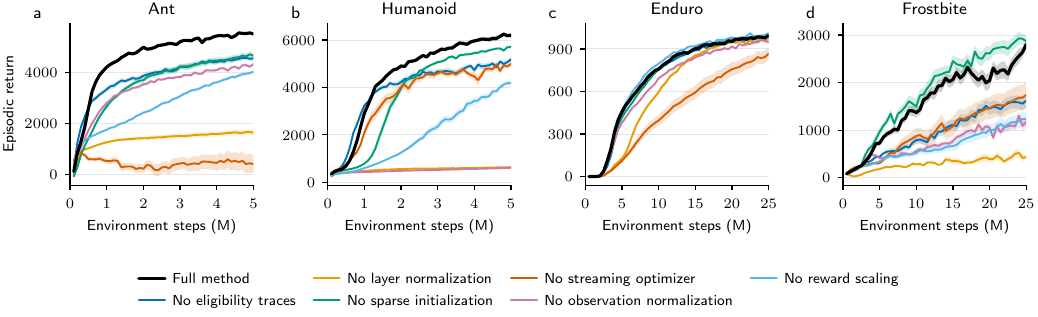}
    \caption{Ablation on the ingredients of Stream-AC (\textbf{a}, \textbf{b}) and Stream-Q (\textbf{c}, \textbf{d}). Variants remove exactly one ingredient from the full method while keeping the rest fixed. Curves are means over $30$ runs, and the shaded area represents the standard error.}
    \label{fig:ablation}
\end{figure}

These results reveal task-dependent sensitivity to the ingredients of the Stream-X recipe. An ingredient whose removal has little effect on one task can substantially affect performance on another (Appendix~\ref{appendix:full-ablation}). These leave-one-out comparisons support combining these ingredients for performance across diverse tasks, but do not establish that every ingredient is necessary or rule out effective alternative combinations.

\section{Related work}

\textbf{Continual learning.}
The goal of continual learning research is to develop algorithms that allow agents to keep learning indefinitely in a nonstationary environment. 
The two obstacles in continual learning are loss of plasticity~\citep{dohare2024lop} and catastrophic forgetting~\citep{McCloskey1989,hetherington1989}.
Loss of plasticity reduces the agent's ability to learn gradually over time, while catastrophic forgetting prevents the agent from retaining and utilizing past memories, ultimately hindering its performance improvement.
Some works focus on maintaining plasticity primarily in nonstationary supervised learning \citep[e.g.,][]{dohare2024lop,kumar2023maintaining,Lewandowski2023curvature,elsayed2024upgd,lewandowski2024learning,lee2024hare}, whereas others focus on reinforcement learning setting~\citep[e.g.,][]{delfosse2024adaptive,xu2024DrM,ma2024revisiting,lyle2024disentangling,lyle2024normalization,lyle2023understanding,elsayed2024weight}. On the other hand, a few works \citep[e.g.,][]{elsayed2024upgd} address catastrophic forgetting and loss of plasticity at the same time. In particular, \citet{elsayed2024upgd} addresses catastrophic forgetting by protecting useful weights from drastic change when using gradient-based updates. Another approach to address forgetting is to promote sparse representations. Sparse representations \citep[e.g.,][]{lan2023elephant} address forgetting since their gradient-based updates make significant changes to the weights of fewer connections than dense representations.

Our work evaluates Stream-AC in two nonstationary environments: varying friction in a simulated Ant (\Cref{sec:fast_adapt_streaming_learning}) and a vision-based navigation task on-board the Open Ant robot in naturally changing laboratory conditions (\Cref{sec:vision_based_navigation}).
Apart from these two nonstationary Ant environments, the other environments we studied are stationary, yet there are commonalities between instability issues in single-task learning and continual learning issues such as loss of plasticity~\citep{dohare2023overcoming}.
For example, LayerNorm, recently shown effective against loss of plasticity~\citep{lyle2023understanding}, is also found beneficial against instability in our work.
Moreover, StreamingOptimizer was found effective against learning instability.
Our update control may also be beneficial for addressing catastrophic forgetting to some extent, as it may alleviate interference by bounding the update.

\textbf{Leaky maximum  normalization.}
AdaGrad~\citep{duchi2011adagrad}, RMSProp~\citep{tielman2012lecture}, and Adam~\citep{kingma2015adam} normalize updates using accumulated or exponentially averaged squared gradients. In contrast, a \emph{leaky maximum}, $v_t=\max(\beta v_{t-1},|u_t|)$, where $\beta\in[0,1)$ and $u_t\in\mathbb{R}$, tracks an upper envelope: it captures peaks immediately, forgets them gradually, and always dominates the current $u_t$. AdaMax~\citep[Section~7.1]{kingma2015adam} derives this recurrence as the infinity-norm limit of exponentially weighted $p$-norm accumulators and uses it to normalize Adam's bias-corrected first moment. An earlier precedent is AutoStep \citep{mahmood2012tuning}, which uses maximum-based normalization in step-size adaptation to control the response to sudden changes in scale. Its normalizer captures peaks immediately but relaxes toward the current sample, rather than decaying toward zero at a fixed rate as is the case in our optimizer.

StreamingOptimizer applies the leaky maximum to the full TD increment $|\delta_t z_t|$ without an additional first moment. Matching the normalized increment to the quantity tracked by the leaky maximum bounds each coordinate's displacement by $\alpha$, as shown in Appendix~\ref{appendix:streaming-optimizer}, independently of the decay rate. Instead, AdaMax normalizes a first moment by a leaky maximum of gradient magnitudes, so its update is not guaranteed to be in the interval $(-\alpha, \alpha)$ regardless of the decay rates. In contrast, applying the leaky maximum to the bias-corrected first moment instead of the gradients gives the displacement bound, $\Delta\in(-\alpha,\alpha)$,  in a general-purpose optimizer (see Algorithm~\ref{alg:bounded-optimizer} in Appendix~\ref{appendix:streaming-optimizer}).

\noindent\textbf{Streaming reinforcement learning.}
Most algorithms introduced by \citet{sutton2018rl} are streaming. However, their usage remained limited to tabular and linear approximation. Later, several attempts were made to use neural networks with those streaming methods. There are a few works that target the problem setting of streaming reinforcement learning. For example, \citet{elfwing2018sigmoid} proposed a new activation function that gave more stability and showed promise with deep Q-learning but still with limited performance. Similarly, \citet{young2019minatar} demonstrated the effectiveness of the approach by \citet{elfwing2018sigmoid} on their MinAtar benchmark, a simplified version of the Atari benchmark \citep{bellemare2013ale}. \citet{deAsis2020incremental} introduced an incremental version of REINFORCE that works well in small problems \citep[see also][]{kimura1995reinforcement}. \citet{javed2024swift} introduced the SwiftTD algorithm for prediction problems by applying it to the last linear layer of a deep network and TD$(\lambda)$ to all other layers. \citet{modayil2023towards} introduced a streaming method for prediction that learns from unstructured observations.

Concurrently with the first version of this work, \citet{vasan2024avg} introduced a policy-gradient method with deep neural networks that learns without batch updates, target networks, or replay buffers.
The main differences are twofold: 1) their method is specific to reparameterized policy gradient, whereas we provide a general recipe that spans value-based and policy-based control, and 2) unlike their work, our results use a single prescribed hyperparameter configuration per algorithm, held fixed across 58 Atari and 50 continuous control tasks, as well as on hardware.
Our work is a comprehensive step toward reviving streaming deep RL by providing a class of algorithms for both value-based and policy-based control, showing success on most commonly used complex benchmark tasks.

\noindent\textbf{Adaptive control.} 
Adaptive control algorithms~\citep{slotine1991applied, ioannou1996robust, krstic1995nonlinear, narendra2012stable}, such as Model Reference Adaptive Control (MRAC) and Self-tuning Control (STC), can also be considered forms of streaming learning.
They update the controller parameters incrementally after each new observation, without storing or replaying past interactions in a buffer.
This allows the controller to continuously adapt to changes in the  environment.
However, adaptive control uses a known parametric structure of the system, with parameters that appear linearly, whereas Stream-X considers deep reinforcement learning with neural network policies and value functions.
Examples of streaming learning using adaptive control on hardware include NASA's X-15 crewed hypersonic rocket-powered aircraft~\citep{x15_adapt_control},
remote-controlled airplanes~\citep{Shi2020Adaptive}, quadcopters~\citep{O_Connell_2022}, and ground vehicles~\citep{10706036}. 

\noindent\textbf{Resource-constrained learning.} Learning under tight memory and computational constraints connects our work to TinyML, which brings machine learning to microcontrollers and other resource-constrained devices, including tiny robots \citep{neuman2022tiny} and edge devices \citep{lin2022ondevice,zhu2022ondevice}. Recent work has made such learning more practical by reducing the memory and computational costs of backpropagation \citep{cai2020tinytl,profentzas2022minilearn,patil2022poet,lin2023tiny}. 
Deep RL introduces an additional challenge: replay buffers and rollout batches store past interactions, adding memory demands beyond those of the network and its updates. Stream-X addresses this algorithmic challenge by learning from each transition without storing past transitions for learning. 
Our Stream-Q experiment demonstrates this connection concretely: the agent learns Pong from pixels directly on a microcontroller (\Cref{fig:esp32}; Appendix~\ref{appendix:esp32}).
Prior work \citep{eschmann2024rltools} demonstrated learning using SAC \citep{haarnoja2018soft} directly on a microcontroller for a pendulum simulated environment using low-dimensional state observations. Our contribution on resource-constrained learning is not the first deep RL learning on a microcontroller, nor a claim that such devices cannot fit experience buffers. 
Rather, vision-based learning or more complex observation vectors make stored experience expensive. Our small buffer results from~\Cref{fig:figure1}c show why fitting a small buffer is not sufficient for reliable learning in the evaluated methods.

\section{Limitations and future work}
Although we have explored a few representative streaming RL algorithms, our approach is compatible with many other algorithms, such as double Q-learning \citep{vanhasselt2010doubleQ}, dueling Q-learning networks \citep{wang2016dueling}, noisy networks Q-learning \citep{fortunato2018noisy}, or even in the continuing setting \citep[e.g.,][]{naik2024reward}.  Additionally, we have considered only simple MLP and CNN architectures, leaving the exploration of more advanced modern architectures like residual networks, also including those designed specifically for streaming learning (e.g., \citealt{modayil2023towards}), to future work. 
Our paper focuses on model-free methods, which are less sample-efficient compared to model-based ones; thus, a promising direction is to discover how the agent can incrementally learn a model of the environment to improve sample efficiency following the success of batch model-based methods \citep[e.g.,][]{hafner2025mastering,samsami2024mastering,liu2024locality}. Another promising direction is to combine our approach with real-time recurrent learning \citep{williams1989learning} to handle partial observability, especially with the recent scalable approaches \citep[e.g.,][]{irie2024exploring,zucchet2024online,elelimy2024real,javed2023scalable}. In addition, we focus mainly on on-policy methods (except for Stream-Q) and leave comprehensive experimentation with our approach to a broader range of off-policy methods \citep[e.g.,][]{sutton2016emphatic,he2023lcetd} for future work.

\section{Conclusion}
Deep reinforcement learning agents can learn effectively from a stream of experience without replay or batch updates as demonstrated by Stream-X. Across the benchmarks studied in this work, streaming variants of representative deep RL algorithms and classical streaming algorithms with deep networks show substantial learning difficulties despite the reported hyperparameter searches. Stream-X algorithms overcome this empirical pattern, which we call stream barrier, through a recipe shared across value-based and actor--critic learning.

Stream-X combines representation stabilization, signal normalization, and bounded parameter updates with streaming algorithms that use eligibility traces. The resulting value-based and actor--critic methods learn effectively across diverse tasks using a single prescribed hyperparameter configuration per algorithm without per-task tuning.

The experiments illustrate the practical capabilities of streaming learning. Stream-AC recovers performance across repeated changes in floor friction more effectively than the evaluated PPO and SAC baselines. Stream-Q learns pixel-based Pong on a microcontroller, and the Open Ant study evaluates learning with proprioceptive and visual features in a nonstationary laboratory environment. These demonstrations motivate streaming RL for adaptive, resource-constrained agents. By making deep RL effective across these benchmark domains without storing past transitions, Stream-X establishes a practical foundation for agents that learn directly from ongoing interaction.

\section*{Acknowledgment}
We gratefully acknowledge funding from the Canada CIFAR AI Chairs program, the Reinforcement Learning and Artificial Intelligence (RLAI) laboratory, the Alberta Machine Intelligence Institute (Amii), and the Natural Sciences and Engineering Research Council (NSERC) of Canada. We would also like to thank the Digital Research Alliance of Canada for providing the computational resources needed.
We would like to thank NASA Jet Propulsion Laboratory, California Institute of Technology for access to laboratory space for the Open Ant experiments. We acknowledge the use of AI tools for copyediting and improving figures presentation.

\bibliographystyle{plainnat}
\bibliography{references}

\newpage

\appendix

\section{Streaming Optimizer}
\label{appendix:streaming-optimizer}

Under streaming learning, individual updates can be noisy or large. \textit{StreamingOptimizer} introduced in~\Cref{alg:streaming-optimizer} normalizes the full increment $\vu_t$ by a decaying element-wise running maximum $\vv_t$.
The optimizer takes as parameters a retained fraction $\beta$ and a small $\varepsilon$ for numerical stability.

\begin{algorithm}[ht]
\small
\caption{StreamingOptimizer($\vu, \vv, \beta, \varepsilon$)}\label{alg:streaming-optimizer}
\begin{algorithmic}[1]
\State \textbf{Require:} 
increment $\vu$,  leaky maximum $\vv$, retained fraction $\beta$, and small number $\varepsilon$
\vspace{0.3em}
\State $\vv \leftarrow \max(\beta \vv, |\vu|)$ \Comment{{\color{gray} Decaying element-wise running max}} \label{alg_line:update_for_v}
\vspace{0.3em}
\State \textbf{Return:} $\frac{\vu}{\vv+\varepsilon}$, $\vv$
\end{algorithmic}
\end{algorithm}

We establish two properties: (1) a per-coordinate displacement bound and (2) an asymmetric accumulator response that incorporates a spike immediately, while forgetting the spike gradually. 
We then state a general-purpose variant, \textit{BoundedOptimizer} in~\Cref{alg:bounded-optimizer}.

\subsection{Per-coordinate displacement bound}

Assume finite increments of $\vu$, a positive step-size $\alpha>0$, $\varepsilon>0$, and initial leaky-max $\vv_0=\mathbf{0}$. By construction, per dimension, \Cref{alg_line:update_for_v} in~\Cref{alg:streaming-optimizer} becomes
\[
v_{t,i}=\max(\beta v_{t-1,i},|u_{t,i}|)\geq |u_{t,i}|.
\]
Hence, the magnitude of the weights change becomes
\[
|\Delta_{t,i}|= 
\left| \alpha\frac{u_{t,i}}{v_{t,i}+\varepsilon}\right| = 
\alpha\frac{|u_{t,i}|}{v_{t,i}+\varepsilon}
\leq\alpha\frac{|u_{t,i}|}{|u_{t,i}|+\varepsilon}<\alpha.
\]
The bound holds for every coordinate and every finite increment sequence.

For comparison, consider uncentered RMSProp~\citep{tielman2012lecture} without momentum, with accumulator $q_{t,i}=\rho q_{t-1,i}+(1-\rho)g_{t,i}^2$, $q_{0,i}=0$, and $\rho\in[0,1)$. Since $q_{t,i}\geq(1-\rho)g_{t,i}^2$, its update obeys
\begin{equation}\label{eq:rms_prop_bound}
|\Delta_{t,i}|=\alpha\frac{|g_{t,i}|}{\sqrt{q_{t,i}}+\varepsilon}
\leq\frac{\alpha}{\sqrt{1-\rho}},
\end{equation}
where $g_{t, i}$ is the per-dimension gradient.
After a long sequence of small gradients, a sufficiently large gradient approaches this upper bound. With $\rho=0.999$, the bound is approximately $31.6\alpha$. Thus RMSProp has a finite bound for every fixed $\rho<1$, but no bound independent of $\rho$ as $\rho\to1$. StreamingOptimizer instead bounds the displacement by the nominal step size itself, removing the coupling between the accumulator decay rate and the step size.

\subsection{Asymmetric accumulator response}

The leaky maximum incorporates a new increment immediately, whenever $|u_{t,i}|>\beta v_{t-1,i}$. In the absence of a new maximum, it decays by $\beta$ per step. For $0<\beta<1$, its half-life is $\ln(2)/[-\ln(\beta)]$, approximately $13{,}863$ steps at the prescribed $\beta=0.99995$.

In contrast, RMSProp uses the same exponential averaging rule for increases and decreases: only a fraction $1-\rho$ of the current squared gradient enters the accumulator. After a prolonged sequence of small gradients, this leads to the potentially large normalized step quantified in Eq.~\ref{eq:rms_prop_bound}.

\subsection{Illustrative increment and update time series}
\label{appendix:illustrative-time-series}
Figure~\ref{fig:optimizer-spikes} expands the optimizer illustration in Figure~\ref{fig:Stream-X-skeleton}b using controlled scalar increment sequences. RMSProp and StreamingOptimizer receive exactly the same signed increments $u_t$. StreamingOptimizer uses $v_t=\max(\beta v_{t-1},|u_t|)$. RMSProp uses $q_t=\rho q_{t-1}+(1-\rho)u_t^2$, without momentum or bias correction. We match the decay rates of the normalization statistics by setting $\rho=\beta^2$: StreamingOptimizer's $\beta=0.9$ and $0.99$ are paired with RMSProp's $\rho=0.81$ and $0.9801$, respectively. These illustrative rates make the response visible over $250$ steps. With zero increments after time $t$, the accumulators are given by
\[
v_{t+k}=\beta^k v_t,\qquad
\sqrt{q_{t+k}}=\rho^{k/2}\sqrt{q_t}=\beta^k\sqrt{q_t}.
\]
Thus, the statistics have the same zero-input decay factor and half-life, while retaining their different responses to incoming increments. This matching does not equate their magnitudes or trajectories under nonzero input: the leaky maximum immediately incorporates a new maximum, whereas RMSProp incorporates only a fraction $1-\rho$ of the current squared increment.

We initialize both accumulators to zero and use $\varepsilon=10^{-8}$. The displayed signed updates are $\Delta_t/\alpha=u_t/(v_t+\varepsilon)$ and $u_t/(\sqrt{q_t}+\varepsilon)$, respectively. Mirroring the positive normalization statistics about zero allows direct comparison with signed increments. Isolated spikes show the immediate incorporation of a new maximum and its subsequent decay (see \Cref{fig:optimizer-spikes}).

\begin{figure}[p]
  \centering
  \includegraphics[width=\linewidth]{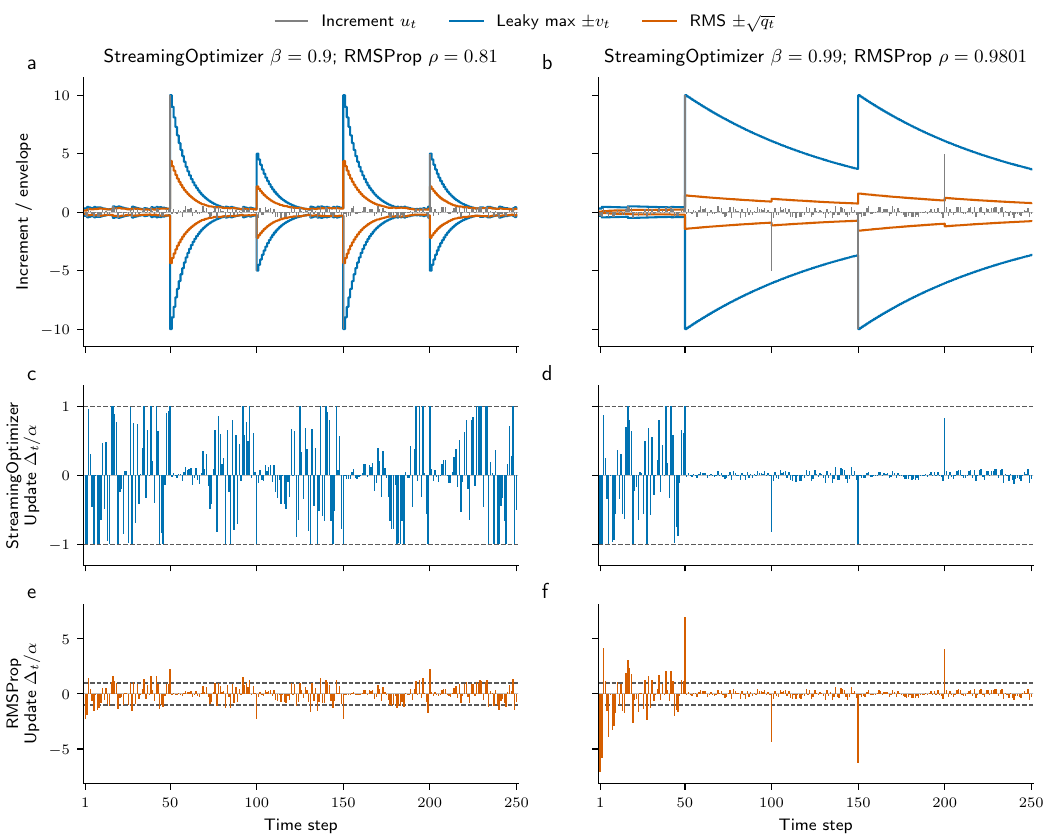}
  \caption{\textbf{Optimizer responses to isolated increment spikes.}
  A fixed random sequence has independent magnitudes uniform on $[0,0.5]$ and equiprobable signs, with increments replaced by $10$ at step $50$, $-5$ at step $100$, $-10$ at step $150$, $5$ at step $200$.
  Columns use $\beta=0.9$ and $0.99$, respectively, with RMSProp retention $\rho=\beta^2$ ($0.81$ and $0.9801$), matching the zero-input decay of $v_t$ and $\sqrt{q_t}$. Dashed horizontal lines in the update panels mark $\pm1$; vertical scales are shared within each row but differ between the two optimizers so that StreamingOptimizer's smaller updates remain visible. Both accumulators start at zero, so RMSProp's first nonzero update also displays its initialization transient.
  \textbf{(a, b)} Gray bars show the signed increments; blue and orange step curves show the mirrored leaky maximum $\pm v_t$ and square root of the second moment $\pm\sqrt{q_t}$. Curves change only when the corresponding increment arrives.
  \textbf{(c, d)} StreamingOptimizer updates divided by $\alpha$. The updates are bounded by $\alpha$ regardless of the increment sequence.
  \textbf{(e, f)} Corresponding RMSProp updates.   Despite matched zero-input decay rates, RMSProp's partial incorporation of a spike permits larger transient updates, whereas StreamingOptimizer bounds each update by $\alpha$.}
  \label{fig:optimizer-spikes}
\end{figure}

\subsection{BoundedOptimizer: a general-purpose variant}
\label{appendix:bounded-optimizer}

Neither property established above depends on the streaming setting. Both follow from taking a decaying running maximum of the very quantity that is then divided by it, and that construction transfers unchanged to ordinary gradient-based learning.
Combining the two gives \textit{BoundedOptimizer}, stated as Algorithm \ref{alg:bounded-optimizer}. From the per-dimension gradient $g_{t,i}$ it forms the bias-corrected first moment $\widehat{m}_{t,i}=m_{t,i}/(1-\beta_1^{t})$ of Adam, with $m_{t,i}=\beta_1 m_{t-1,i}+(1-\beta_1)g_{t,i}$, accumulates the running maximum of that same quantity, $v_{t,i}=\max(\beta_2 v_{t-1,i},|\widehat{m}_{t,i}|)$, and updates $\Delta_{t,i}=\alpha\,\widehat{m}_{t,i}/(v_{t,i}+\varepsilon)$, all operations element-wise. Since $v_{t,i}\ge|\widehat{m}_{t,i}|$ holds by construction, the displacement argument given above applies verbatim: $|\Delta_{t,i}|<\alpha$ for every coordinate, every gradient sequence, and every pair of decay rates. The accumulator is likewise asymmetric, so a gradient spike is capped on the step it occurs and cannot be re-amplified on the steps that follow.

\begin{algorithm}[ht]
\small
\caption{BoundedOptimizer($\vg, \vm, \vv, \beta_1, \beta_2, t,\varepsilon$)}\label{alg:bounded-optimizer}
\begin{algorithmic}[1]
\State \textbf{Require:}
gradient $\vg$, first moment $\vm$, leaky maximum $\vv$, first-moment decay $\beta_1$, running-max decay $\beta_2$, and time step $t$
\vspace{0.5em}
\State $t \leftarrow t + 1$
\State $\vm \leftarrow \beta_1 \vm + (1-\beta_1)\vg$
\State $\widehat{\vm} \leftarrow \vm/(1-\beta_1^t)$
\State $\vv \leftarrow \max(\beta_2 \vv, |\widehat{\vm}|)$ \Comment{{\color{gray} Decaying element-wise running max}}
\State \textbf{Return:} $\frac{\widehat{\vm}}{\vv+\varepsilon}, \vm, \vv, t$
\end{algorithmic}
\end{algorithm}

The closest existing construction is AdaMax \citep{kingma2015adam}, which differs in exactly one place: it takes the maximum of $|g_{t,i}|$ rather than of $|\widehat{m}_{t,i}|$, so its numerator and denominator are formed from different quantities. How large the update is depends on both decay rates. 
Consider a coordinate $i$. For $0<\beta_2<1$, the recurrence $u_{t,i}=\max(\beta_2u_{t-1,i},|g_{t,i}|)$ implies, for every $1\le k\le t$,
\[
u_{t,i}\ge\beta_2u_{t-1,i}\ge\cdots\ge\beta_2^{t-k}u_{k,i}
\ge\beta_2^{t-k}|g_{k,i}|.
\]
Dividing by the positive factor $\beta_2^{t-k}$ gives
\[
|g_{k,i}|\le\beta_2^{-(t-k)}u_{t,i}.
\]
With $m_{0,i}=0$, we have $m_{t,i}=(1-\beta_1)\sum_{k=1}^{t}\beta_1^{t-k}g_{k,i}$. For $u_{t,i}>0$, bias correction, the triangle inequality, and the bound above give
\[
\begin{aligned}
\frac{|\widehat{m}_{t,i}|}{u_{t,i}}
&=\frac{|m_{t,i}|}{(1-\beta_1^t)u_{t,i}}\\
&=\frac{1-\beta_1}{(1-\beta_1^t)u_{t,i}}
\left|\sum_{k=1}^{t}\beta_1^{t-k}g_{k,i}\right|\\
&\le\frac{1-\beta_1}{(1-\beta_1^t)u_{t,i}}
\sum_{k=1}^{t}\beta_1^{t-k}|g_{k,i}| &&\left(\substack{\text{using the triangle inequality }
\left|\sum_k a_k \right|\le\sum_k|a_k|}\right)\\
&\le\frac{1-\beta_1}{(1-\beta_1^t)u_{t,i}}
\sum_{k=1}^{t}\beta_1^{t-k}\beta_2^{-(t-k)}u_{t,i}\\
&=\frac{1-\beta_1}{1-\beta_1^t}
\sum_{k=1}^{t}\left(\frac{\beta_1}{\beta_2}\right)^{t-k}.
\end{aligned}
\]
When $0\le\beta_1<\beta_2$, this is a geometric sum with ratio $\beta_1/\beta_2<1$
\[
\sum_{k=1}^{t}\left(\frac{\beta_1}{\beta_2}\right)^{t-k}
=\sum_{j=0}^{t-1}\left(\frac{\beta_1}{\beta_2}\right)^j
=\frac{1-(\beta_1/\beta_2)^t}{1-\beta_1/\beta_2}
\le\frac{1}{1-\beta_1/\beta_2}
=\frac{\beta_2}{\beta_2-\beta_1}.
\]
Consequently,
\[
\frac{|\widehat{m}_{t,i}|}{u_{t,i}}
\le\frac{(1-\beta_1)\,\beta_2}{(1-\beta_1^t)\,(\beta_2-\beta_1)},
\qquad 0\le\beta_1<\beta_2,
\]
with strict inequality when $\beta_1>0$.
As $t\to\infty$, the bound approaches $(1-\beta_1)\beta_2/(\beta_2-\beta_1)$. With $\beta_2=0.999$ fixed, this is approximately $1.009$, $1.11$, and $1.25$ for $\beta_1=0.9$, $0.99$, and $0.995$, respectively. In contrast, normalizing the corrected first moment by its own running maximum, as in BoundedOptimizer, bounds each coordinate's displacement by the step size independently of both decay rates.


\section{Algorithms pseudocodes}
\label{appendix:pseudocodes}

We give the pseudocode of the four Stream-X control algorithms and the Stream-TD prediction algorithm below, followed by their shared subroutines.
In each, {\color{purple} LayerNorm} and {\color{brown} SparseInit} stabilize the representation, {\color{blue} NormalizeObservation} and {\color{orange} ScaleReward} normalize the signals, and {\color{teal} StreamingOptimizer} controls the update.

\begin{algorithm}[ht]
\small
\caption{Stream-RAC}\label{alg:stream-rac}
\begin{algorithmic}[1]
\State \textbf{Given} Action-value network $\hat{q}(s, a;\vw)$ with {\color{purple} LayerNorm} and weights $\vw$ initialized with {\color{brown} SparseInit}, and policy network $\pi(a|s; \boldsymbol\theta)$ parameterizing a tanh-normal distribution with {\color{purple} LayerNorm} and weights $\boldsymbol\theta$ initialized with {\color{brown} SparseInit}
\State \textbf{Initialize} 
\State \quad Discount factor $\gamma$ and eligibility traces parameter $\lambda$ 
\State \quad Statistics $\mu_S, \mu_r, p_S, p_r$, reward trace $U$, time $t$, and small number $\varepsilon$
\State \quad Optimizer state $\vv_{\vw}, \vv_{\boldsymbol\theta}$, critic step size $\alpha_{\hat{q}}$, policy step size $\alpha_{\pi}$, and retained fraction $\beta$
\State \quad Entropy coefficient $\alpha_\text{ent}$ and dimensionality of action space $n$
\For{each episode}
\State $\vz_{\vw}\leftarrow \mathbf{0}$, $U\leftarrow 0$
\State Initialize $S$ \Comment{{\color{gray} First state of the episode}}
\State {\color{blue} $S, \mu_S, p_S \leftarrow$  NormalizeObservation($S, \mu_S, p_S, t$)}
\State $A \sim \pi(\cdot|S, \boldsymbol\theta)$ \Comment{{\color{gray} $A = \tanh(\tilde{A})$, $\tilde{A} \sim \mathcal{N}(\mu_{\boldsymbol\theta}(S),\, \sigma_{\boldsymbol\theta}(S))$}}
\For{each step in the episode}
\State $t \leftarrow t + 1$
\State Take action $A$, observe $S^\prime$, $R$, $T$ \Comment{{\color{gray} $T$ indicates whether $S^\prime$ is a terminal state}}
\State {\color{blue} $S^\prime, \mu_S, p_S \leftarrow$  NormalizeObservation($S^\prime, \mu_S, p_S, t$)}
\State {\color{orange} $R, U, \mu_r, p_r \leftarrow$  ScaleReward($R, \gamma, U, \mu_r, p_r, T, t$)}
\State $A^\prime \sim \pi(\cdot|S^\prime, \boldsymbol\theta)$ \Comment{{\color{gray} Sampled without gradients; it is also the action executed next}}
\State $\delta \leftarrow R + \gamma \left(\hat{q}(S^\prime, A^\prime, \vw) - \frac{\alpha_\text{ent}}{n} \log \pi(A^\prime|S^\prime, \boldsymbol\theta)\right) - \hat{q}(S, A, \vw)$ \Comment{{\color{gray} If $T$, bootstrapped target $\doteq 0$}}
\State $A_{\boldsymbol\theta} \sim \pi(\cdot|S, \boldsymbol\theta)$ \Comment{{\color{gray} Reparameterized sample, gradients flow through $\boldsymbol\theta$}}
\State $\mathcal{J}_\pi \leftarrow \hat{q}(S, A_{\boldsymbol\theta}, \vw) - \frac{\alpha_\text{ent}}{n} \log \pi(A_{\boldsymbol\theta}|S, \boldsymbol\theta)$
\State {\color{teal}$\vd_\pi, \vv_{\boldsymbol\theta} \gets \text{StreamingOptimizer}(\nabla_{\boldsymbol\theta} \mathcal{J}_\pi, \vv_{\boldsymbol\theta}, \beta, \varepsilon)$} \Comment{{\color{gray} Using no eligibility traces}}
\State $\boldsymbol\theta \gets \boldsymbol\theta + \alpha_{\pi}\vd_\pi$
\State $\vz_{\vw} \leftarrow \gamma \lambda \vz_{\vw} + \nabla_{\vw} \hat{q}(S, A, \vw)$
\State {\color{teal}$\vd_{\hat{q}}, \vv_{\vw} \gets \text{StreamingOptimizer}(\delta\vz_\vw, \vv_{\vw}, \beta, \varepsilon)$}
\State $\vw \gets \vw + \alpha_{\hat{q}}\vd_{\hat{q}}$
\State $S \leftarrow S^\prime$,\; $A \leftarrow A^\prime$
\EndFor
\EndFor
\end{algorithmic}
\end{algorithm}

\begin{algorithm}[ht]
\small
\caption{Stream-SARSA}\label{alg:stream-sarsa}
\begin{algorithmic}[1]
\State \textbf{Given} Action-value network $\hat{q}(s, a;\vw)$ with {\color{purple} LayerNorm} and weights $\vw$ initialized with {\color{brown} SparseInit}
\State \textbf{Initialize} 
\State \quad Discount factor $\gamma$ and eligibility traces parameter $\lambda$ 
\State \quad Exploration schedule $\epsilon_\text{start}$, $\epsilon_\text{end}$, exploration fraction $f$, and budget of $N$ time steps 
\State \quad Statistics $\mu_S, \mu_r, p_S, p_r$, reward trace $U$, time $t$, and small number $\varepsilon$
\State \quad Optimizer state $\vv_{\vw}$, step size $\alpha$, and retained fraction $\beta$
\State \quad Exploration probability $\epsilon\leftarrow\epsilon_\text{start}$
\For{each episode}
\State $\vz_{\vw}\leftarrow \mathbf{0}$, $U\leftarrow 0$
\State Initialize $S$ \Comment{{\color{gray} First state of the episode}}
\State {\color{blue} $S, \mu_S, p_S \leftarrow$  NormalizeObservation($S, \mu_S, p_S, t$)}
\State Choose $A$ from $S$ $\epsilon$-greedily w.r.t. $\hat{q}(S, \cdot, \vw)$
\For{each step in the episode}
\State $t \leftarrow t + 1$
\State Take action $A$, observe $S^\prime$, $R$, $T$ \Comment{{\color{gray} $T$ indicates whether $S^\prime$ is a terminal state}}
\State {\color{blue} $S^\prime, \mu_S, p_S \leftarrow$  NormalizeObservation($S^\prime, \mu_S, p_S, t$)}
\State {\color{orange} $R, U, \mu_r, p_r \leftarrow$  ScaleReward($R, \gamma, U, \mu_r, p_r, T, t$)}
\State $\epsilon \leftarrow \max\left(\epsilon_\text{start} + (\epsilon_\text{end} - \epsilon_\text{start})\frac{t}{f N},\; \epsilon_\text{end}\right)$ \Comment{{\color{gray} Linearly annealed exploration}}
\State Choose $A^\prime$ from $S^\prime$ $\epsilon$-greedily w.r.t. $\hat{q}(S^\prime, \cdot, \vw)$
\State $\delta \leftarrow R + \gamma \hat{q}(S^\prime, A^\prime, \vw) - \hat{q}(S, A, \vw)$ \Comment{{\color{gray} If $S^\prime$ is terminal, then $\hat{q}(S^\prime,\cdot, \vw)\doteq 0$}}
\State $\vz_{\vw} \leftarrow \gamma \lambda \vz_{\vw} + \nabla_{\vw} \hat{q}(S, A, \vw)$
\State {\color{teal}$\vd, \vv_{\vw} \gets \text{StreamingOptimizer}(\delta\vz_\vw, \vv_{\vw}, \beta, \varepsilon)$}
\State $\vw \gets \vw + \alpha\vd$
\State $S \leftarrow S^\prime$,\; $A \leftarrow A^\prime$
\EndFor
\EndFor
\end{algorithmic}
\end{algorithm}

\begin{algorithm}[ht]
\small
\caption{Stream-TD}\label{alg:stream-td}
\begin{algorithmic}[1]
\State \textbf{Given} State-value network $\hat{v}(s;\vw)$ with {\color{purple} LayerNorm} and weights $\vw$ initialized with {\color{brown} SparseInit}
\State \textbf{Initialize} 
\State \quad Discount factor $\gamma$ and eligibility traces parameter $\lambda$ 
\State \quad Statistics $\mu_S, \mu_r, p_S, p_r$, reward trace $U$, time $t$, and small number $\varepsilon$
\State \quad Optimizer state $\vv_{\vw}$, step size $\alpha$, and retained fraction $\beta$
\For{each episode}
\State $\vz_{\vw}\leftarrow \mathbf{0}$, $U\leftarrow 0$
\State Initialize $S$ \Comment{{\color{gray} First state of the episode}}
\State {\color{blue} $S, \mu_S, p_S \leftarrow$  NormalizeObservation($S, \mu_S, p_S, t$)}
\For{each step in the episode}
\State $t \leftarrow t + 1$
\State Observe $S^\prime$, $R$, $T$ \Comment{{\color{gray} $T$ indicates whether $S^\prime$ is a terminal state}}
\State {\color{blue} $S^\prime, \mu_S, p_S \leftarrow$  NormalizeObservation($S^\prime, \mu_S, p_S, t$)}
\State {\color{orange} $R, U, \mu_r, p_r \leftarrow$  ScaleReward($R, \gamma, U, \mu_r, p_r, T, t$)}
\State $\delta \leftarrow R + \gamma \hat{v}(S^\prime, \vw) - \hat{v}(S, \vw)$ \Comment{{\color{gray} If $S^\prime$ is terminal, then $\hat{v}(S^\prime, \vw)\doteq 0$}}
\State $\vz_{\vw} \leftarrow \gamma \lambda \vz_{\vw} + \nabla_{\vw} \hat{v}(S,\vw)$
\State {\color{teal}$\vd, \vv_{\vw} \gets \text{StreamingOptimizer}(\delta\vz_\vw, \vv_{\vw}, \beta, \varepsilon)$}
\State $\vw \gets \vw + \alpha\vd$
\State $S \leftarrow S^\prime$
\EndFor
\EndFor
\end{algorithmic}
\end{algorithm}

\clearpage

\subsection{Shared subroutines}

The shared subroutines are given below, together with TrackMeanVar, the estimator used by both signal-normalization subroutines. Each is itself streaming: it keeps a constant amount of state per parameter or per observation coordinate and never revisits a past transition.

\textbf{\color{brown} SparseInit} (Algorithm~\ref{alg:sparse_init}) allocates $(1-s)\sum_l n_l$ nonzero weights across layers following \citet{yildirim2024continual}, with $s=0.95$. Here, $n_l$ counts layer $l$'s weights and $m_l$ its incoming weights per unit. Allocations are proportional to $\phi_l=\texttt{fan\_in}+\texttt{fan\_out}$ for linear layers or $C_{\text{in}}+C_{\text{out}}+k_h+k_w$ for convolutions, capped at $n_l$, with excess redistributed. Convolutional kernels are viewed as rows of $m_l=C_{\text{in}}k_hk_w$ weights per output channel. Uniformly initialized weights are randomly zeroed per row; biases are zero and all weights remain trainable.

\textbf{TrackMeanVar} (Algorithm~\ref{alg:track-mean-var}) is the estimator underlying both signal-normalization subroutines. It maintains a running mean and variance for a scalar or vector stream in a single pass. We used Welford's algorithm \citep{welford1962}, which accumulates the corrected sum of squares $p$ and reports the unbiased sample variance. It weights every observation equally and converges to the true moments of a stationary stream. Although we track the mean and variance using sample estimates based on Welford's algorithm in this work, we also tried relying on exponential moving averages that track the mean and variance with a decay of $0.99999$ and observed very similar performance.

\textbf{\color{blue} NormalizeObservation} (Algorithm~\ref{alg:obs-norm}) applies TrackMeanVar to the observation vector, coordinate by coordinate, and returns the standardized observation that is fed to the network. The statistics are updated with the observation before it is used, so the network never sees an unstandardized input. For frame-stacked image observations the statistics are shared across the stacked frames, since the frames are samples of the same image distribution.

\textbf{\color{orange} ScaleReward} (Algorithm~\ref{alg:reward-norm}) divides the reward by the running standard deviation of the discounted reward trace $U_t=\gamma U_{t-1}+r_t$ without subtracting a mean. The trace is reset at episode boundaries, but its accumulated statistics are retained.

\begin{algorithm}[ht]
\small
\caption{\color{brown} SparseInit}\label{alg:sparse_init}
\begin{algorithmic}
\State \textbf{Require:} network $f$ with at least one linear or convolutional layer, target sparsity $s\in[0,1)$
\State $B\gets(1-s)\sum_l n_l$, $\mathcal{D}\gets\emptyset$
\Repeat
\State $\epsilon \gets \left(B-\sum_{l\in\mathcal{D}} n_l\right)\big/\sum_{l\notin\mathcal{D}}\phi_l$ \Comment{$\phi_l$ is allocation score for layer $l$}
\State $\mathcal{D}\gets\mathcal{D}\cup\{l\notin\mathcal{D} : \epsilon\phi_l\ge n_l\}$
\Until{$\mathcal{D}$ stops growing or contains every layer}
\State $d_l\gets\min\left(1,\sfrac{\epsilon\phi_l}{n_l}\right)$ for each layer $l$
\For{each linear or convolutional layer $l$}
\State Sample each weight $W^l_{i,j}$ independently from $U[-1/\sqrt{m_l},\,1/\sqrt{m_l}]$
\State $k_l\gets\lceil(1-d_l)m_l\rceil$
\For{each row $i$}
\State Draw $\mathcal{I}_i\subseteq\{1,\ldots,m_l\}$ uniformly from all subsets of size $k_l$
\State $W^l_{i,j}\gets0$ for every $j\in\mathcal{I}_i$
\EndFor
\State $\vb^l\gets\mathbf{0}$
\EndFor
\State \textbf{Return:} initialized network $f$
\end{algorithmic}
\end{algorithm}

\noindent\begin{minipage}{.49\textwidth}
\begin{algorithm}[H]
\small
\caption{{\color{blue} NormalizeObservation}}\label{alg:obs-norm}
\begin{algorithmic}
\State \textbf{Require:} $S, \mu, p, n$, small number $\varepsilon$
\State $\mu, p, \sigma^2, n \leftarrow$ TrackMeanVar($S, \mu, p, n$)
\State \textbf{Return:}  $\frac{S-\mu}{\sqrt{\sigma^2 + \varepsilon}}$, $\mu$, $p$
\State
\State
\vspace{0.07cm}
\end{algorithmic}
\end{algorithm}
\end{minipage}\hfill
\begin{minipage}{.49\textwidth}
\begin{algorithm}[H]
\small
\caption{{\color{orange} ScaleReward}}\label{alg:reward-norm}
\begin{algorithmic}
\State \textbf{Require:} $r, \gamma, U, \mu, p, T, n$, small number $\varepsilon$
\State $U \leftarrow \gamma U + r$
\State $\mu, p,\sigma^2, n \leftarrow$ TrackMeanVar($U, \mu, p, n$)
\State $\bar{r} \gets \frac{r}{\sqrt{\sigma^2 + \varepsilon}}$
\State \textbf{Return:} $\bar{r}$, $U$, $\mu$, $p$
\end{algorithmic}
\end{algorithm}
\end{minipage}

\begin{algorithm}[ht]
\small
\caption{TrackMeanVar (Welford 1962)}\label{alg:track-mean-var}
\begin{algorithmic}
\State \textbf{Require}: input $x$, mean $\mu$, statistic $p$, and counter $n$, with $\mu, p, n$ initialized to zero
\State $n \gets n + 1$
\State $\bar{\mu} \gets \mu + \frac{1}{n}(x-\mu)$
\State $p \gets p + (x-\mu)(x-\bar{\mu})$
\State $\sigma^2\gets \frac{p}{n-1}$ if $n\geq 2$, otherwise $\sigma^2\gets 1$
\State \textbf{Return:} $\bar{\mu}, p, \sigma^2, n$
\end{algorithmic}
\end{algorithm}

\section{Relationship to the earlier version}
\label{appendix:relationship-to-earlier-work}

This manuscript builds on \citet{elsayed2024streaming}, released as arXiv:2410.14606 on 18 October 2024. The comparison here refers to its revised version of 6 December 2024 (v2). That work introduced the term \emph{stream barrier} and the Stream-X approach, combining eligibility traces, signal normalization, layer normalization, sparse initialization, and update control. It presented Stream-AC, Stream-Q, Stream-SARSA, and Stream-TD, with experiments in continuous control, MinAtar, Atari, and time-series prediction. Here, we discuss how the current manuscript differs from the earlier version.

\paragraph{Revised update control.} The earlier algorithms used Overshooting-bounded Gradient Descent (ObGD), which scales an update using the trace's $L_1$ norm and the TD-error magnitude, motivated by control of the effective step size. Its adjusted step size is
\[
\alpha_t^{\rm ObGD}=\frac{\alpha}{\max\left(1,\alpha\kappa\max(1,|\delta_t|)\|\vz_t\|_1\right)},
\]
where $\kappa$ is the overshooting-control parameter. This global scaling couples every parameter's update to the summed trace magnitude across the network. As a network grows, an increase in $\|\vz_t\|_1$ makes the shared step size smaller. For example, with $d$ nonzero trace coordinates of equal magnitude, each coordinate's displacement in this regime is $|\delta_t|/[\kappa\max(1,|\delta_t|)d]$, decreasing as $1/d$. Thus, ObGD can become increasingly conservative per parameter in larger networks. Instead, StreamingOptimizer divides each coordinate of the full increment (e.g., $\vu_t=\delta_t\vz_t$ for TD-based updates) by its own decaying running maximum, avoiding this network-wide normalization. Appendix~\ref{appendix:streaming-optimizer} establishes a per-coordinate displacement bound and characterizes the accumulator's response to large increments; it also introduces the general-purpose BoundedOptimizer variant.

\paragraph{Layer-aware sparse initialization.} The earlier initialization applied the same sparsity fraction $s$ to every layer. Its implementation zeros $\lceil s m\rceil$ incoming weights per unit, where $m$ is the layer's fan-in. If $(1-s)m<1$, this rounds up to all $m$ weights, making the entire weight matrix zero. In a fully connected first layer, high sparsity or a low-dimensional observation can therefore remove every input connection at initialization: for example, $s=0.95$ and $m=10$ leave no nonzero weights. With zero biases, the first-layer pre-activations are then zero for every observation, so observation information cannot propagate into the network at initialization. This is an initial loss of input dependence, not necessarily permanent disconnection, because the zeroed weights remain trainable. Instead, the current initialization based on \cite{yildirim2024continual} allocates a global weight budget across layers, allowing low-fan-in layers to receive higher densities rather than imposing the same sparsity everywhere (Algorithm~\ref{alg:sparse_init}). This addresses the uniform-allocation limitation, although budget allocation and rounding alone do not guarantee nonzero connections at arbitrarily high sparsity.

\paragraph{Broader algorithmic and empirical evaluation.} The current family adds Stream-RAC, applying the recipe to reparameterized actor--critic learning. The benchmark study now includes explicit $30$-configuration baseline searches, per-task configuration selection, and PPO/SAC/DQN buffer-size sweeps (Figure~\ref{fig:figure1}). Figure~\ref{fig:figure1} reports median all-best-normalized performance over $50$ continuous-control tasks and $58$ Atari games under the protocol in Appendix~\ref{appendix:normalization}. Atari streaming-method reporting uses $30$ runs rather than the earlier $10$, and continuous-control reporting uses a $5$M-step budget rather than $20$M. We also extend the ingredient ablations to both Stream-AC and Stream-Q and report network diagnostics. Because the algorithms, budgets, and reporting procedures differ between versions, differences in their reported scores are not a controlled comparison of the two versions.

\paragraph{Adaptation and on-device learning.} The earlier work studied learning in stationary benchmark environments and motivated resource-constrained and continual learning applications. This manuscript presents a continual learning experiment through the alternating-friction Ant setup, including a SAC replay-capacity comparison (Figures~\ref{fig:alternating-friction-ant} and~\ref{fig:ant-replay-buffers}). It also adds a robot heading-tracking study with proprioceptive and visual features, and pixel-based Pong learning on a microcontroller (Figures~\ref{fig:ant-heading} and~\ref{fig:esp32}). These studies extend the experimental scope from benchmark learning to adaptation and resource-constrained learning.

\paragraph{Development and reporting.} Appendix~\ref{appendix:protocol} now makes the development protocol explicit: the recipe was developed on five MuJoCo Gym tasks, while discrete-control hyperparameters were selected on MinAtar before the first Atari evaluation without algorithm changes or retuning. This documents how configurations were obtained. The current reporting specifies all-best normalization with median aggregation across all tasks in Figure~\ref{fig:figure1}. Divergence is treated per run, rather than assigning zero to an entire method--task result when any run diverges as in the earlier version.


\section{Variance equivalence of returns and reward traces}
\label{appendix:var-equivalence}

A return-based normalization scheme is appealing because TD targets are governed by the scale of discounted returns rather than by the scale of one-step rewards. The return, however, is \textit{non-causal}: computing it at time $t$ requires rewards that have not yet been observed, making it unavailable in a strict streaming update. A natural causal substitute is the discounted reward trace, which depends only on present and past rewards. Here, we show that, under a second-order stationarity assumption, the return and reward trace have the same variance when measured jointly over fixed-horizon episodes and uniformly sampled timesteps.

\noindent Consider episodes of fixed horizon $T$. Let episode $m$ produce rewards $r_1^{(m)},\ldots,r_T^{(m)}$. At each timestep $t\in\{1,\ldots,T\}$, define
\[
G_t^{(m)}
=
\sum_{k=0}^{T-t}\gamma^k r_{t+k}^{(m)},
\qquad
U_t^{(m)}
=
\sum_{k=0}^{t-1}\gamma^k r_{t-k}^{(m)}.
\]
\noindent Under our assumption, there exist a constant $\mu$ and a lag-covariance function $c$ such that
\[
\mathbb{E}_m\left[r_t^{(m)}\right]=\mu,
\qquad
\operatorname{Cov}_m
\left(r_i^{(m)},r_j^{(m)}\right)
=
c(j-i)
\]
for all $i,j\in\{1,\ldots,T\}$. Since covariance is symmetric,
$c(\ell)=c(-\ell)$.

\begin{proposition}
Let $\tau$ be uniformly distributed over $\{1,\ldots,T\}$ and independent of the episode index $m$. Under the assumption above,
\[
\operatorname{Var}_{m,\tau}
\left(G_\tau^{(m)}\right)
=
\operatorname{Var}_{m,\tau}
\left(U_\tau^{(m)}\right).
\]
\end{proposition}

\begin{proof}
For $L\in\{1,\ldots,T\}$, define
\[
a_L
=
\mu\sum_{k=0}^{L-1}\gamma^k,
\qquad
v_L
=
\sum_{i=0}^{L-1}
\sum_{j=0}^{L-1}
\gamma^{i+j}c(j-i).
\]
By the stationarity assumption,
\[
\mathbb{E}_m\left[G_t^{(m)}\right]=a_{T-t+1},
\qquad
\operatorname{Var}_m\left(G_t^{(m)}\right)=v_{T-t+1},
\]
whereas
\[
\mathbb{E}_m\left[U_t^{(m)}\right]=a_t,
\qquad
\operatorname{Var}_m\left(U_t^{(m)}\right)=v_t.
\]
For $U_t^{(m)}$, the covariance terms involve $c(i-j)$. Since covariance is symmetric, $c(i-j)=c(j-i)$, and therefore $\operatorname{Var}_m\left(U_t^{(m)}\right)=v_t$.
\\
\\
\noindent Because $\tau$ is uniform over $\{1,\ldots,T\}$, the random variables $\tau$ and $T-\tau+1$ have the same distribution. Therefore, by the law of total variance,
\begin{align*}
\operatorname{Var}_{m,\tau}
\left(G_{\tau}^{(m)}\right)
&=
\mathbb{E}_{\tau}
\left[
\operatorname{Var}_{m}
\left(G_{\tau}^{(m)} \mid \tau\right)
\right]
+
\operatorname{Var}_{\tau}
\left(
\mathbb{E}_{m}
\left[G_{\tau}^{(m)} \mid \tau\right]
\right)
\\
&=
\mathbb{E}_{\tau}
\left[v_{T-\tau+1}\right]
+
\operatorname{Var}_{\tau}
\left(a_{T-\tau+1}\right)
\\
&=
\mathbb{E}_{\tau}
\left[v_{\tau}\right]
+
\operatorname{Var}_{\tau}
\left(a_{\tau}\right)
\\
&=
\mathbb{E}_{\tau}
\left[
\operatorname{Var}_{m}
\left(U_{\tau}^{(m)} \mid \tau\right)
\right]
+
\operatorname{Var}_{\tau}
\left(
\mathbb{E}_{m}
\left[U_{\tau}^{(m)} \mid \tau\right]
\right)
\\
&=
\operatorname{Var}_{m,\tau}
\left(U_{\tau}^{(m)}\right).
\end{align*}
\end{proof}
The equality holds jointly over episodes and uniformly sampled timesteps. It does not generally hold at each fixed timestep, since $\operatorname{Var}_m\left(G_t^{(m)}\right)=v_{T-t+1}$ and $\operatorname{Var}_m\left(U_t^{(m)}\right)=v_t$.


\section{Experimental Details}
\label{appendix:experimental-details}

We implemented the algorithms in Python and used PyTorch \citep{paszke2017auto} for automatic differentiation to backpropagate gradients in neural networks. In addition, we use the Gymnasium \citep{towers2026gymnasium} framework for environment implementations.

We reset all trace memories, e.g., eligibility or reward traces, when a reset signal is provided by the environment, as a reset indicates a discontinuity in experience.

Unless specified otherwise, benchmark learning curves report the average over $30$ runs. The alternating-friction experiment uses $60$ runs per method, and the microcontroller experiment uses $5$ runs. Each streaming baselines tuning configuration (see \Cref{appendix:protocol}) is based on $3$ runs. Each Stream-X algorithm uses a single prescribed hyperparameter configuration across tasks, as specified below, without per-task tuning.

\paragraph{Stream-X development process.} We developed the Stream-X recipe through experiments on five MuJoCo Gym environments: Ant, HalfCheetah, Humanoid, Walker2d, and Hopper. Once the recipe produced stable learning on these development environments, we fixed it and evaluated it on the remaining continuous-control environments without further changes to the recipe or per-task tuning. The continuous-control aggregates include both the five development environments and the subsequent evaluation environments.

For value-based algorithms, we selected the Stream-Q hyperparameters using MinAtar environments \citep{young2019minatar}. We fixed these settings before evaluating the algorithms on the full Atari benchmark suite for the first time, without modifying the algorithms or retuning their hyperparameters on Atari. The Atari results therefore assess transfer from MinAtar to a separate benchmark suite, whereas the continuous-control results assess transfer beyond the five development tasks within the reported suite.

\subsection{Streaming baselines tuning and reporting}
\label{appendix:protocol}

\paragraph{Configuration generation and budgets.} For each baseline, we construct $30$ configurations by including the defaults specified below and sampling the remaining configurations independently from Table~\ref{tab:baseline-search}. Each configuration is evaluated with $3$ independent runs, where we picked the configuration that achieved best average performance for the longer runs in \Cref{fig:figure1}. IAC, PPO1, and SAC1 are tuned on $50$ continuous-control environments for $2$M steps per run. DQN1, Q-learning are tuned on $58$ Atari games for $3{,}125{,}000$ agent steps per run ($12.5$M frames with action repeat four). 

\begin{table}[ht]
\centering
\scriptsize
\caption{Hyperparameter search spaces for streaming baselines.}
\label{tab:baseline-search}
\renewcommand{\arraystretch}{1.08}
\begin{tabular}{@{}>{\raggedright\arraybackslash}p{0.14\linewidth}>{\raggedright\arraybackslash}p{0.40\linewidth}>{\raggedright\arraybackslash}p{\dimexpr0.46\linewidth-4\tabcolsep\relax}@{}}
\toprule
\textbf{Algorithm} & \textbf{Parameter} & \textbf{Sampling distribution / range} \\
\midrule
\multirow[c]{5}{*}{\textbf{DQN1}} & Learning rate & $\mathrm{LogU}[10^{-7},10^{-3}]$ \\
 & Adam $\beta_1$ & $U[0.7,0.99]$ \\
 & Adam $\beta_2$ & $U[0.9,0.9999]$ \\
 & Adam $\varepsilon$ & $\mathrm{LogU}[10^{-10},10^{-6}]$ \\
 & Target-update interval (steps) & $\{500,2000\}$ \\
\addlinespace[5pt]
\multirow[c]{5}{*}{\textbf{PPO1}} & Initial learning rate & $\mathrm{LogU}[10^{-6},10^{-3}]$ \\
 & Adam $\beta_1$ & $U[0.7,0.99]$ \\
 & Adam $\beta_2$ & $U[0.9,0.999]$ \\
 & Adam $\epsilon$ & $\mathrm{LogU}[10^{-10},10^{-6}]$ \\
 & Clipping coefficient & $U[0.05,0.4]$ \\
\addlinespace[5pt]
\multirow[c]{5}{*}{\textbf{SAC1}} & Critic learning rate $\alpha_Q$ & $\mathrm{LogU}[10^{-6},10^{-3}]$ \\
 & Adam $\beta_1$ & $U[0.7,0.99]$ \\
 & Adam $\beta_2$ & $U[0.9,0.9999]$ \\
 & Adam $\epsilon$ & $\mathrm{LogU}[10^{-10},10^{-6}]$ \\
 & Target averaging coefficient $\tau$ & $U[0.001,0.05]$ \\
\addlinespace[5pt]
\multirow[c]{3}{*}{\textbf{IAC}} & Actor step size & $\mathrm{LogU}[10^{-7},10^{-3}]$ \\
 & Critic step size (independent) & $\mathrm{LogU}[10^{-7},10^{-3}]$ \\
 & Shared trace parameter $\lambda$ & $\{0.5,0.8,0.9,0.95,0.99\}$ \\
\addlinespace[5pt]
\multirow[c]{1}{*}{\textbf{Q-learning}} & Step size & $\mathrm{LogU}[10^{-7},10^{-3}]$ \\
\bottomrule
\end{tabular}
\par\smallskip
\begin{minipage}{\linewidth}
\footnotesize\raggedright
$U[a,b]$ denotes continuous uniform sampling; $\mathrm{LogU}[a,b]$ denotes uniform sampling in $\log_{10}$ space. Finite sets and integer ranges are sampled uniformly. Default configurations are included separately and can lie outside the random-sampling ranges.
\end{minipage}
\end{table}

\paragraph{Included defaults.} DQN1 includes two configurations, with Adam $\beta_1=0$ or $0.9$. Both use learning rate $10^{-4}$, $\beta_2=0.999$, $\varepsilon_{\rm Adam}=10^{-8}$, and target-update interval $1000$. PPO1 includes two configurations with initial learning rate $3\times10^{-4}$, the same two $\beta_1$ choices, $\beta_2=0.999$, $\varepsilon_{\rm Adam}=10^{-5}$, and clipping coefficient $0.2$. SAC1 includes two configurations with $\alpha_Q=10^{-3}$, the same two $\beta_1$ choices, $\beta_2=0.999$, $\varepsilon_{\rm Adam}=10^{-8}$, and $\tau=0.005$. Thus, each of these searches adds $28$ random configurations. IAC includes one default with actor and critic step sizes both $10^{-4}$ and $\lambda=0$; Q-learning includes one default with step size $3\times10^{-4}$, followed by $29$ randomly sampled configurations, with $\lambda=0$ fixed throughout its search.

\paragraph{Selection and reporting.} A run's score is the average of episodic returns whose terminating steps lie in the final $10\%$ of the tuning budget: the last $200{,}000$ steps in continuous control and $312{,}500$ agent steps in Atari. We average these run scores ($3$ independent runs per configuration) and select the highest-scoring configuration separately for each environment. A recorded divergence in any tuning run assigns the configuration a score of $-\infty$. For continuous control, ties between configurations with score $-\infty$ retain the same order, selecting the first included default if every configuration diverges. Best-performing streaming-baseline configurations are run for $5$M steps in continuous control and $50$M agent steps ($200$M frames) in Atari with $30$ independent runs.

For the Atari streaming baselines, the random-action probability $\epsilon$ decreases linearly from $1$ to $0.01$ over the first $2.5$M agent steps and then remains at $0.01$. The discount factor is fixed at $\gamma=0.99$ for all searches.

\subsection{Performance normalization in \Cref{fig:figure1}}
\label{appendix:normalization}

Each run is scored by the average return of episodes terminating at or after $90\%$ of the reporting budget: $4.5$M steps in continuous control or $45$M agent steps in Atari. We then average equally across runs to obtain $R_{m,e}$ for method $m$ on task $e$. Diverged runs receive the task's random-policy score.

\paragraph{Median all-best-normalized performance in Figure~\ref{fig:figure1}.} Let $\mathcal{M}_d$ contain all methods and buffer settings displayed in panels b and c for domain $d$, and define $A_e=\max_{m\in\mathcal{M}_d}R_{m,e}$. Thus, $A_e$ is the best observed averaged return on task $e$, including Stream-X and every displayed buffer size. Every method and setting on that task uses the same score:
\[
z_{m,e}=\begin{cases}
\displaystyle\frac{\max(0,R_{m,e}-R_e^{\rm random})}{A_e-R_e^{\rm random}}, & A_e>R_e^{\rm random},\\[6pt]
0, & A_e\leq R_e^{\rm random}.
\end{cases}
\]
The best-observed references use the same final-window scoring as the evaluated methods. Scores lie in $[0,1]$: zero denotes random performance or worse, and one denotes the best observed performance on that task. The evaluated method may set its own denominator. The DQN and SAC sweeps use replay capacities $1$, $10$, $100$, $10^3$, $10^4$, $10^5$, and $10^6$; PPO uses rollout lengths $1$, $10$, $100$, $10^3$, $2048$, $10^4$, and $10^5$.

Each plotted bar or sweep point is $\operatorname{median}_{e\in\mathcal{E}_d}z_{m,e}$, where $\mathcal{E}_d$ is the task set for the corresponding domain. The median is taken across normalized task scores. For an even number of tasks, the median is the average of the two middle scores. .

Both panels retain all $58$ Atari games and $50$ continuous-control tasks ($40$ DM Control and $10$ MuJoCo Gym tasks) listed in Table~\ref{tab:benchmark-environments}. The vertical dashed comparison lines in panel b use these same $z_{m,e}$ scores: the median for DQN with $10^6$ replay transitions is $0.672$, and $\operatorname{median}_e\max(z_{\mathrm{SAC}(10^6),e},z_{\mathrm{PPO}(2048),e})=1.000$. The maximum is taken separately on each task before taking the median. The latter equals one because these two batch settings attain the all-best reference on $31$ of $50$ tasks, not because they win every task.

\paragraph{Supplemental baseline comparison in Figure~\ref{fig:aggregate-all-methods}.} This comparison uses the same all-best normalization rule and full task cohorts as Figure~\ref{fig:figure1}, but replaces Stream-Q with Stream-SARSA on Atari and Stream-AC with Stream-RAC on continuous control so that we do not have two Stream-X algorithms competing against each other. The replacement applies to both the displayed methods and the reference calculations.

\begin{table}[ht]
\centering
\scriptsize
\renewcommand{\arraystretch}{1.15}
\caption{Benchmark tasks in Figures~\ref{fig:figure1} and~\ref{fig:aggregate-all-methods}: all $50$ continuous-control tasks and $58$ Atari games. DM Control identifiers omit the common \texttt{-v0} suffix.}
\label{tab:benchmark-environments}
\begin{tabular}{@{}>{\raggedright\arraybackslash}p{0.18\linewidth}>{\raggedright\arraybackslash}p{\dimexpr0.82\linewidth-2\tabcolsep\relax}@{}}
\toprule
Domain & Environments \\
\midrule
MuJoCo Gym (10) &
Ant-v4; HalfCheetah-v4; Hopper-v4; Humanoid-v4; HumanoidStandup-v4; InvertedDoublePendulum-v4; InvertedPendulum-v4; Pusher-v4; Reacher-v4; Walker2d-v4. \\
\addlinespace
DM Control (40) &
acrobot-swingup; acrobot-swingup\_sparse; ball\_in\_cup-catch; cartpole-balance; cartpole-balance\_sparse; cartpole-swingup; cartpole-swingup\_sparse; cartpole-three\_poles; cartpole-two\_poles; cheetah-run; dog-fetch; dog-run; dog-stand; dog-trot; dog-walk; finger-spin; finger-turn\_easy; finger-turn\_hard; fish-swim; fish-upright; hopper-hop; hopper-stand; humanoid-run; humanoid-stand; humanoid-walk; manipulator-bring\_ball; manipulator-bring\_peg; pendulum-swingup; point\_mass-easy; point\_mass-hard; quadruped-fetch; quadruped-run; quadruped-walk; reacher-easy; reacher-hard; swimmer-swimmer15; swimmer-swimmer6; walker-run; walker-stand; walker-walk. \\
\addlinespace
Atari (58) &
Air Raid; Alien; Amidar; Assault; Asterix; Asteroids; Atlantis; Bank Heist; Battle Zone; Beam Rider; Berzerk; Bowling; Boxing; Breakout; Carnival; Centipede; Chopper Command; Crazy Climber; Defender; Demon Attack; Double Dunk; Enduro; Fishing Derby; Freeway; Frostbite; Gopher; Gravitar; H.E.R.O.; Ice Hockey; James Bond; Kangaroo; Krull; Kung Fu Master; Montezuma's Revenge; Ms. Pacman; Name This Game; Phoenix; Pitfall; Pong; Private Eye; Q*bert; River Raid; Road Runner; Robotank; Seaquest; Skiing; Solaris; Space Invaders; Star Gunner; Tennis; Time Pilot; Tutankham; Up'N Down; Venture; Video Pinball; Wizard of Wor; Yars' Revenge; Zaxxon. \\
\bottomrule
\end{tabular}
\end{table}

\subsection{Details on agents used in Atari experiments}
Stream-Q and Stream-SARSA use a network composed of a convolutional layer \citep{LeCun1998gradient} with $32$ filters of size $8\times 8$ and using a stride of $5$ followed by another convolutional layer with $64$ filters of size $4\times 4$ and using a stride of $3$ followed by another convolutional layer with $64$ filters of size $3\times 3$ and using a stride of $2$. The output of the last convolutional layer is flattened and is fed to a fully connected layer with $256$ hidden units and another one with $128$ hidden units. We used LeakyReLU activations \citep{maas2013rectifier} where LayerNorm \citep{ba2016layer} is added before each activation layer. Stream-Q and Stream-SARSA use $\lambda=0.8$ and $\gamma=0.99$. We used $\beta$ of $0.99995$ and a step size of $3\times 10^{-4}$ for StreamingOptimizer. Each agent experiences $50$M time steps or $200$M frames in total and uses an $\epsilon$-greedy policy where $\epsilon$ starts with $1$ and decreases to $0.01$ with a linear schedule that reaches $\epsilon=0.01$ at $5\%$ of the total time steps used. Each action taken by the agent is repeated  $4$ times. Lastly, we used sparse initialization with a sparsity ratio $s$ of $95\%$.

The Atari environments are set up the same way as the work by \citet{hessel2018rainbow} except for some key changes that make the environment harder. We downsampled the frames to $84 \times 84$; then, we converted each frame from RGB to grayscale. To address partial observability, we stacked $4$ frames. We made the agent take a random action at the start of the episode for environments that are fixed until the firing command is used where applicable. The episode is terminated on loss of life. Lastly, we made the agent take a random number of no-operation (no-op) actions (up to $30$) at the beginning of each episode. The main difference between our setup and \citet{hessel2018rainbow} is that we do not clip the rewards, nor do we divide the frame pixels by $255$.

DQN1 is based on the CleanRL implementation \citep{huang2022cleanrl}, with replay capacity and batch size both set to $1$, training frequency $1$, and \texttt{learning\_starts}$=1$. Target networks are retained, with hard-copy coefficient $\tau=1$ and the update interval searched as specified in Table~\ref{tab:baseline-search}. DQN1 uses three convolutional layers with $(32,8,5)$, $(64,4,3)$, and $(64,3,2)$ filters, kernel sizes, and strides, respectively, followed by a $256$-unit fully connected hidden layer and the action-value output layer. Hidden activations are ReLU, without LayerNorm or sparse initialization. Unlike Stream-Q's preprocessing, DQN1 retains reward-sign clipping $\{+1,0,-1\}$ and divides pixel values by $255$ in its forward pass.

Q-learning uses the one-step streaming base of Stream-Q without the Stream-X ingredients: the current temporal-difference error scales the current action-value gradient, and the update is applied directly by stochastic gradient descent, with no eligibility traces ($\lambda=0$), no StreamingOptimizer, no layer normalization, no sparse initialization, and no observation or reward normalization. It uses the same network architecture, the same $\epsilon$-greedy schedule, and the same $50$M-step budget as Stream-Q. Its step size was tuned per game following Appendix~\ref{appendix:protocol}, sampling it log-uniformly from $[10^{-7}, 10^{-3}]$, with $\lambda$ fixed to $0$.

\subsection{Details on agents used in MuJoCo Gym and DM Control tasks}\label{appendix:details_on_agent}

Stream-AC used a $128\times 128$ fully connected network with LeakyReLU activations \citep{maas2013rectifier} where LayerNorm \citep{ba2016layer} is added before each activation layer. In the last layer of the policy network, we used two heads: one for the actions mean and the other for actions standard deviation. We used separate networks for the policy and value functions. In continuous control, the standard deviation is parameterized by the SoftPlus function: $f(x)=\log(1+e^{x})$. For numerical stability, when the input to the function exceeds a threshold of $20$, we used a linear mapping of $y=x$. The actions are clamped to be in the range $[-1, 1]$. For discrete control, softmax policy parameterization can be used. Stream-AC uses $\lambda=0.8$ and $\gamma=0.99$. For StreamingOptimizer we used a retained fraction $\beta$ of $0.99995$ with a step size of $10^{-4}$ for the policy network and $2\times 10^{-4}$ for the value network, and an entropy coefficient $\tau$ of $0.01$. The agent experiences $5$M time steps in total. Lastly, we used sparse initialization with a sparsity ratio $s$ of $95\%$.

Stream-AC uses adaptive entropy regularization with coefficient $\tau\geq0$. The regularized policy gradient is given by
$\vg^{\mathrm{reg}}_t = \nabla_{\boldsymbol\theta}\log\pi(A_t\mid S_t;\boldsymbol\theta_t)
+ \tau\,\operatorname{sign}(\delta_t)\nabla_{\boldsymbol\theta}H(\cdot\mid S_t;\boldsymbol\theta_t)$,
with the actor trace $\vz^\theta_t = \gamma\lambda\vz^\theta_{t-1}+\vg^{\mathrm{reg}}_t$,
Thus, the TD error scales the accumulated trace, not the gradient before accumulation. For $\lambda=0$, the raw increment is the policy-gradient term plus $|\delta_t|\tau\nabla_{\boldsymbol\theta}H$; with traces, the current TD error also scales the retained past gradients, as in Algorithm~\ref{alg:stream-ac}.

Stream-RAC uses separate $256\times256$ fully connected network with LeakyReLU activations for policy and action-value networks with parameter-free LayerNorm and sparse initialization with sparsity ratio of $95\%$. We used two heads at the last policy's head: one for the actions mean and the
other for actions standard deviation.
Its tanh-squashed Gaussian policy uses standard deviation parameterized by the SoftPlus. The actor step size is $10^{-4}$, the critic step size is $3\times10^{-4}$, $\gamma=0.99$, $\lambda=0.8$, and the optimizer retention fraction is $0.99995$. The entropy coefficient is $0.025$ divided by the action dimension in both the actor objective and the bootstrapped target. The actor has no eligibility trace while the critic does.

PPO1 is based on CleanRL \citep{huang2022cleanrl}, with one environment, a rollout length of one transition, one minibatch, and one epoch. Actor and critic use separate $64\times64$ tanh networks with orthogonal initialization; the Gaussian policy has a learned state-independent log standard deviation. Observation normalization and reward normalization are retained, with normalized observations and rewards clipped to $[-10,10]$. The initial learning rate is annealed linearly over the training budget. GAE uses $\lambda_{\rm GAE}=0.95$; advantage normalization is disabled, value-loss clipping is enabled, the entropy coefficient is zero, the value-loss coefficient is $0.5$, and the gradient-norm limit is $0.5$.

SAC1 uses two critics, their target networks, and a tanh-squashed Gaussian actor, each with two $256$-unit ReLU hidden layers. The actor's log standard deviation is restricted to $[-5,2]$, and actions are rescaled to the environment bounds. Replay capacity and batch size are $1$, \texttt{learning\_starts}$=1$, and the actor and target-network update intervals are both equal to $1$. The updates begin when the zero-based step index exceeds the learning-start threshold. The actor learning rate is fixed at $0.3\alpha_Q$, and the automatically tuned entropy coefficient uses an Adam optimizer with learning rate $\alpha_Q$ and target entropy equal to minus the action dimension. Actor, critic, and entropy optimizers share the Adam hyperparameters. $\tau$ is the sampled target-network averaging coefficient. SAC1 does not apply observation or reward normalization.

AC($\lambda$), the incremental actor--critic reported as IAC in \Cref{fig:figure1}, is the same streaming base as Stream-AC run without the Stream-X ingredients: the actor and critic traces are applied directly by stochastic gradient descent, with no StreamingOptimizer, no layer normalization, no sparse initialization, and no observation or reward normalization. It uses the same $128\times 128$ networks and the same $5$M-step budget as Stream-AC. Its two step sizes and its trace parameter were tuned per environment, sampling each step size log-uniformly from $[10^{-7}, 10^{-3}]$ and $\lambda$ from $\{0.5, 0.8, 0.9, 0.95, 0.99\}$. Even at its per-environment best configuration it diverges on most tasks: runs were flagged as diverged on $35$ of the $50$ environments, which is why its aggregate in \Cref{fig:figure1} sits near the random-policy level.

\subsection{On-device Pong on the ESP32-S3}
\label{appendix:esp32}

The device we used is a Waveshare ESP32-S3-GEEK, a $4.5\,\mathrm{cm}\times 2.5\,\mathrm{cm}$ board carrying a dual-core Xtensa LX7 processor clocked at $240$\,MHz with $512$\,KB of on-chip SRAM and a $240\times 135$ LCD. Stream-Q, the tensor and neural network layers, the reverse-mode automatic differentiation engine, and the Pong environment are all implemented using TinyRL \citep{elsayed2026tinyrl}, a library for on-device learning. Simulation, rendering, action selection, and learning all run on the device; a host is used only to read the episode log off the board's flash over the serial link after training has finished. 

The game is played on a $42\times 42$ arena and downsampled to a $32\times 32$ grayscale observation; two consecutive frames are stacked, giving $2048$ inputs. The action space has three elements: no-op, up, and down. The reward is $+1$ on the step the agent scores, $-1$ on the step it concedes, and $0$ otherwise. An episode ends when either side reaches $21$ points. Each action is repeated for $4$ frames, so a $500{,}000$-step run corresponds to $2$ million game frames; sticky actions are disabled. The opponent is scripted at the medium difficulty setting: it moves toward the ball's height corrupted by zero-mean Gaussian noise of scale $15$ pixels, with a dead zone of $5$ pixels, at $0.7$ of paddle speed while the ball approaches it and $0.4$ while the ball recedes.

The network used has three convolutional layers with $8$, $16$, and $16$ channels, kernel sizes $5$, $3$, and $3$, and strides $3$, $2$, and $2$, which reduce the $2\times 32\times 32$ input to a $16$-dimensional feature vector, followed by a fully connected layer of $32$ units and a linear output layer of three action values. LayerNorm followed by LeakyReLU is applied after each convolution and after the hidden fully connected layer, for $4{,}539$ parameters in total. All layers are sparsely initialized with $s=0.95$ using the routine of Algorithm~\ref{alg:sparse_init}. Observation normalization and reward scaling are those of Algorithms~\ref{alg:obs-norm} and~\ref{alg:reward-norm}, with the observation statistics tracked per input coordinate. The step size is $3\times 10^{-4}$, $\gamma=0.99$, $\lambda=0.8$, and the StreamingOptimizer retained fraction is $\beta=0.99995$, all matching the prescribed configuration used elsewhere in this work. The $\epsilon$-greedy schedule anneals from $1$ to $0.01$ over the first half of the budget, that is, an exploration fraction of $f=0.5$. The persistent state carried between steps is the parameter vector, the eligibility trace, the optimizer's running maximum, and the observation and reward statistics, which together occupy about $70$\,KB. For scale, storing one million distinct $32\times32$ byte-valued frames requires about $1$\,GB; storing each two-frame observation separately requires at least $2$\,GB.

We report $5$ independent runs of $500{,}000$ agent steps. For every episode, we log its terminating step, the points scored, and the points conceded. Episodes are assigned to $10{,}000$-step bins by their terminating step and averaged within a bin for each run; \Cref{fig:esp32}b reports the mean and standard error of those per-run values across the five runs, smoothed with a centered three-bin moving average. The aggregate figures quoted in the main text are computed over the raw episodes falling in the first and last $50{,}000$ steps, averaged within a run and then across runs.

\subsection{Performance of Stream-RAC and Stream-SARSA}
\label{appendix:all-methods}

Figure~\ref{fig:aggregate-all-methods} complements Figure~\ref{fig:figure1}b by comparing Stream-SARSA with DQN1 and Q-learning on Atari, and Stream-RAC with SAC1, PPO1, and IAC on continuous control. Stream-SARSA and Stream-RAC replace Stream-Q and Stream-AC in both the displayed methods and the normalization reference calculation, demonstrating the recipe with additional base algorithms rather than ranking Stream-X algorithms against one another. Horizontal bars report median all-best-normalized performance.

\begin{figure}[H]
  \centering
  \includegraphics[width=0.9\textwidth]{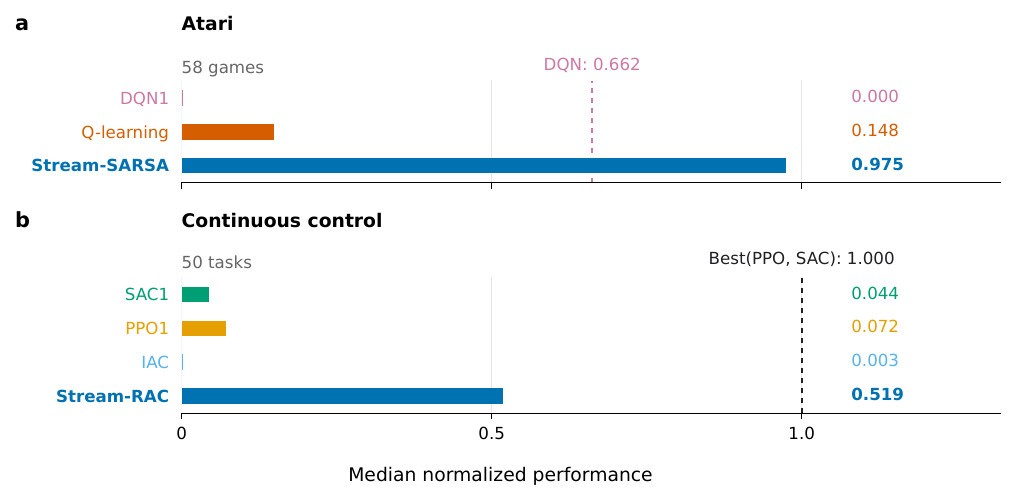}
  \caption{\textbf{Performance of Stream-RAC and Stream-SARSA.}
  \textbf{(a), top:} DQN1, Q-learning, and Stream-SARSA on $58$ Atari games.
  \textbf{(b), bottom:} SAC1, PPO1, IAC, and Stream-RAC on $50$ continuous-control tasks.
  Bars show median normalized performance across tasks, using final-$10\%$ returns averaged across runs. Zero denotes random performance or worse, and one denotes the best observed averaged return methods and buffer settings after replacing Stream-Q with Stream-SARSA on Atari and Stream-AC with Stream-RAC on continuous control. Vertical dashed lines show batch comparisons on the same scale: DQN with $10^6$ replay transitions ($0.662$), and the per-task better of SAC with $10^6$ replay transitions or PPO with a $2{,}048$-step rollout, aggregated by the median ($1.0$). Baselines use the best of $30$ configurations per task; each Stream-X algorithm uses one prescribed configuration across its domain. Blue bars and bold labels identify Stream-X.}
  \label{fig:aggregate-all-methods}
\end{figure}

All Atari methods are evaluated on the same $58$ games, and all continuous-control methods on the same $50$ tasks. Stream-SARSA attains a median score of $0.975$, compared with $0.148$ for Q-learning and less than $0.001$ for DQN1. Stream-RAC attains $0.519$, compared with $0.044$ for SAC1, $0.072$ for PPO1, and $0.003$ for IAC.

\subsection{Task-level performance profiles}
\label{appendix:performance-profiles}

Figure~\ref{fig:performance-profiles} shows the distribution of the task-level scores underlying Figure~\ref{fig:figure1}. At each threshold, a curve reports the percentage of tasks whose normalized score is at least that threshold. These profiles complement the median by showing performance across the full range of task scores, using the same all-best references and all $58$ Atari games and $50$ control tasks.

\begin{figure}[ht]
  \centering
  \includegraphics[width=\textwidth]{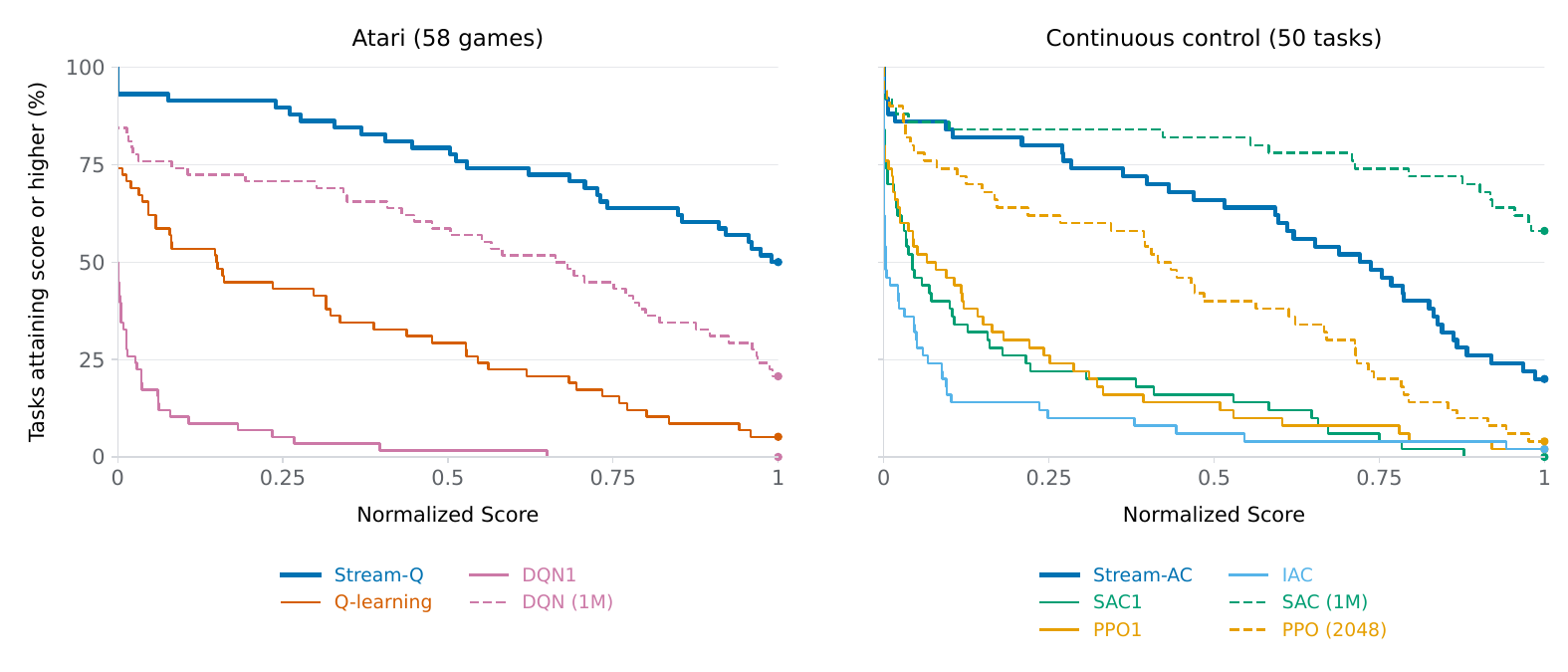}
  \caption{\textbf{Performance profiles across benchmark tasks.}
  \textbf{Left:} Atari ($58$ games). \textbf{Right:} Continuous control ($50$ tasks).
  Each curve gives the percentage of tasks attaining the normalized score on the horizontal axis or higher. Returns are averaged across runs before normalization using the same per-task all-best references as Figure~\ref{fig:figure1}; no median aggregation is applied to these curves. Solid lines denote streaming methods, with Stream-X highlighted in blue. Dashed lines denote DQN or SAC with a million-transition replay buffer and PPO with a $2{,}048$-step rollout. At score one, the endpoint gives the percentage of tasks on which a method matches the best observed reference.}
  \label{fig:performance-profiles}
\end{figure}

\subsection{Configuration coverage at matched short budgets}
\label{appendix:short-budget-coverage}

We show in Figure~\ref{fig:short-budget-coverage} how broadly a single hyperparameter configuration works across tasks at the streaming baselines' tuning budgets. We use all $58$ Atari games and $50$ continuous-control tasks in this experiment with $3{,}125{,}000$ agent steps on Atari ($12.5$M frames) and $2$M steps in continuous control. DQN1, Q-learning, SAC1, PPO1, and IAC each contribute all $30$ configurations from Appendix~\ref{appendix:protocol}. DQN and SAC with million-transition replay buffers, PPO with its default $2{,}048$-step rollout, and Stream-Q and Stream-AC each contribute one fixed configuration.

A run is scored by the mean return of episodes terminating within $[0.9B,B]$, where $B$ is the short budget. We average across runs: $3$ for tuning configuration and $30$ for batch or Stream-X method. A recorded divergence at or before $B$ assigns that run the task's random-policy score.

We apply the all-best normalization of Appendix~\ref{appendix:normalization}. A configuration covers a task when its normalized score is at least $0.25$, meaning at least one quarter of the best observed improvement above random.

\begin{figure}[ht]
  \centering
  \includegraphics[width=\textwidth]{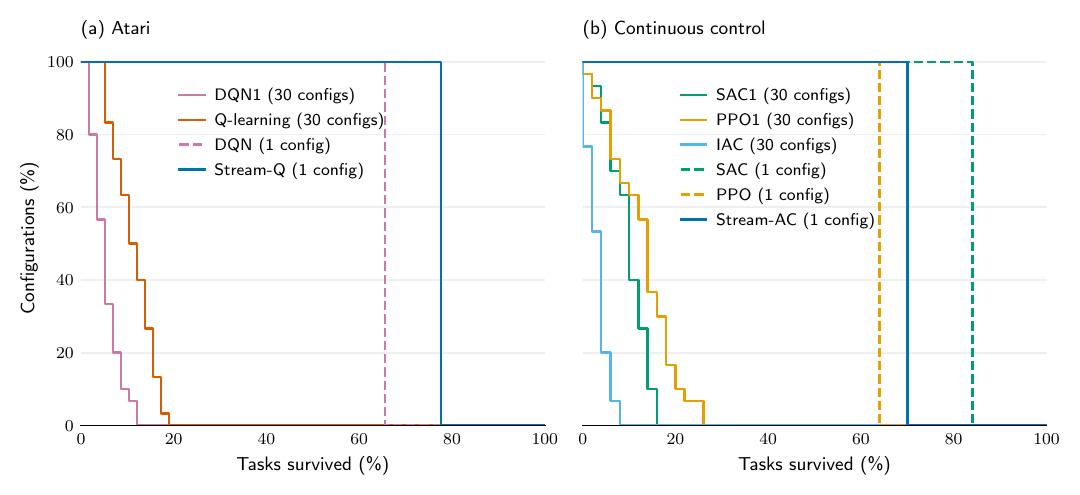}
  \caption{\textbf{Configuration coverage at matched short budgets.}
  \textbf{(a)} Atari ($58$ games; $3.125$M agent steps). \textbf{(b)} Continuous control ($50$ tasks; $2$M steps).
  A task is covered when its short-budget, averaged, all-best-normalized score is at least $0.25$. Each curve shows the percentage of configurations covering at least the percentage of tasks on the horizontal axis. Streaming baselines contribute $30$ configurations each; the batch methods and Stream-X contribute one each. Dashed lines identify the batch methods, and blue lines identify Stream-X. All methods in a panel share the same per-task references.}
  \label{fig:short-budget-coverage}
\end{figure}

\subsection{SAC's buffer-size sweep in nonstationary Ant}
\label{appendix:ant-replay-buffers}

To examine whether a smaller replay buffer alleviates SAC's difficulties after changes in floor friction, we compare replay capacities of $10^0, 10^1, 10^2, 10^3$, $10^4$, $10^5$, and $10^6$ transitions on the alternating-friction Ant task. Each configuration is evaluated over $60$ independent runs of $20$ million environment steps, with friction alternating between $\mu=2.0$ and $\mu=0.02$ every $2$ million steps. Figure~\ref{fig:ant-replay-buffers} repeats the learning-curve view of \Cref{fig:alternating-friction-ant}b, adding the smaller-buffer SAC runs and retaining Stream-AC and PPO as references.

The effect of reducing replay capacity is not monotonic. However, none of the tested capacities reproduces Stream-AC's sustained recovery and improvement across repeated visits. Thus, a smaller replay buffer can mitigate some of SAC's poor performance, but reducing capacity alone does not resolve it in these experiments.

\begin{figure}[ht]
  \centering
  \includegraphics[width=\textwidth]{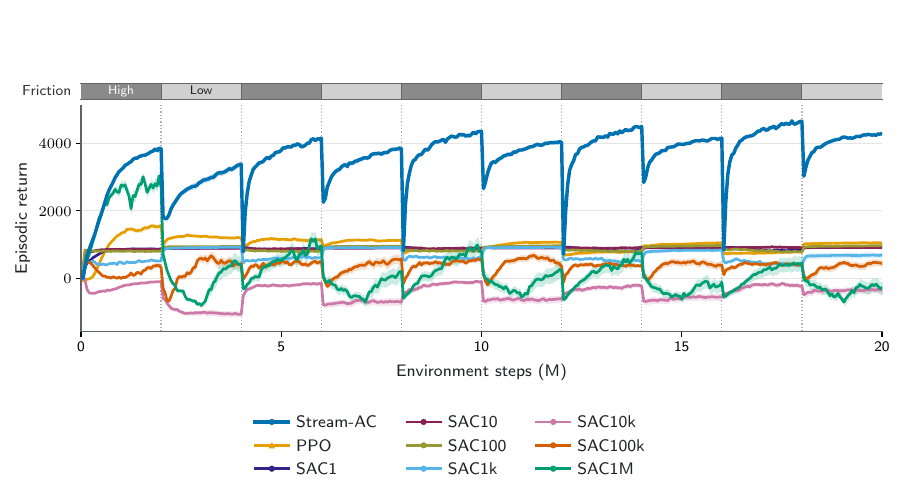}
  \caption{\textbf{SAC replay-buffer sweep in nonstationary Ant.} SAC is shown with replay capacities of  1, 10, 100, 1k, 10k, 100k, and 1M, alongside the same Stream-AC and PPO reference runs as in the main figure. Episodic return is plotted over $20$ million environment steps. The top strip indicates floor friction, which switches every $2$ million steps. Curves show means over $60$ independent runs per configuration; shading denotes the standard error.}
  \label{fig:ant-replay-buffers}
\end{figure}

\section{Diagnostics}
\label{appendix:diagnostics}

To examine learning dynamics, we show in \Cref{fig:learning_dynamics} three network statistics used by \citet{dohare2024lop}, measured for PPO, PPO1, and Stream-AC with the same procedure. These statistics characterize representations and weights, but they are not direct tests of the ability to learn a new task and do not by themselves establish preserved plasticity.

We log three quantities, separately for the policy network and the value network, following the reference implementation of \citet{dohare2024lop}.
\begin{itemize}[leftmargin=1.5em,itemsep=2pt,topsep=2pt]
    \item \textbf{Dormant units (\%)}: the percentage of hidden units that are strictly inactive on every state in the measurement batch. This is a meaningful notion of deadness only for ReLU-like activations. For this reason all three diagnostic agents use ReLU activations rather than the default Tanh activation of PPO/PPO1 implementations or the default LeakyReLU activation used for Stream-AC.
    \item \textbf{Effective rank}: the smallest number of singular values of the last hidden layer's activation matrix whose normalized sum reaches $0.99$. Because this is a count bounded by the layer width, and the agents use different widths, we report it divided by the width, i.e.\ as the fraction of directions retained.
    \item \textbf{Average weight magnitude}: the mean of $|w|$ over the linear mapping weights of the network, excluding biases and the policy's log-standard-deviation parameter. Its absolute scale is fixed by the fan-in dependent initialization, so we report it relative to each run's own value at the first checkpoint, which makes the plotted quantity the drift away from initialization.
\end{itemize}

Each agent keeps a rolling measurement buffer of the $2048$ most recent observations it has encountered, and the current networks are re-evaluated on that buffer every $10{,}000$ environment steps to produce the three statistics above. At each checkpoint, we pass states recently visited by the agent through the checkpoint network to compute the diagnostics. This captures the network’s responses to the agent’s recent state distribution. In contrast, \citet{dohare2024lop} accumulate activations during interaction, while the network is changing. We recompute activations at each checkpoint to apply the same diagnostic procedure to all three algorithms. Apart from the ReLU activation choice above, Stream-AC uses the $128\times 128$ networks and hyperparameters of Appendix~\ref{appendix:experimental-details}; PPO uses the CleanRL defaults ($64\times 64$ networks, $2048$-step rollouts, learning-rate annealing, gradient clipping); and PPO1 uses $64\times 64$ networks with one environment, a one-transition rollout, one minibatch, and one update epoch, with advantage normalization disabled. Each curve is the mean over $30$ runs with a $30$M-step budget. We show the plots smoothed with a centred $50$-checkpoint moving average, corresponding to approximately $0.5$M environment steps.

\begin{figure}[t]
    \centering
    \includegraphics[width=0.48\textwidth]{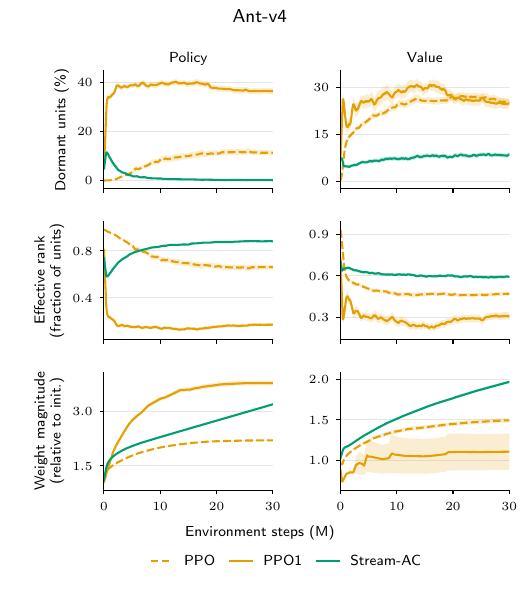}
    \includegraphics[width=0.48\textwidth]{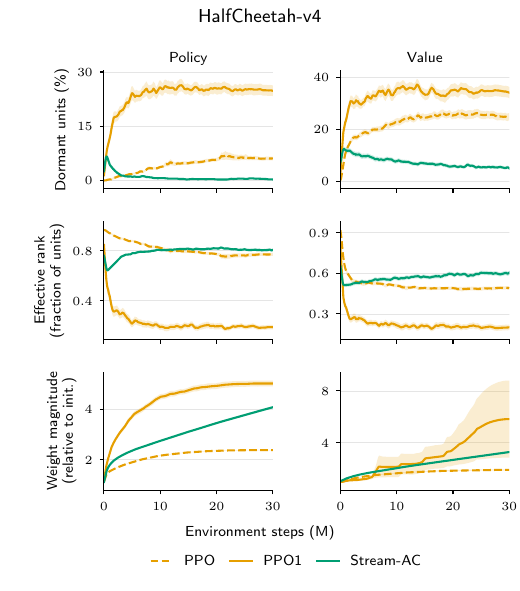}
    \includegraphics[width=0.48\textwidth]{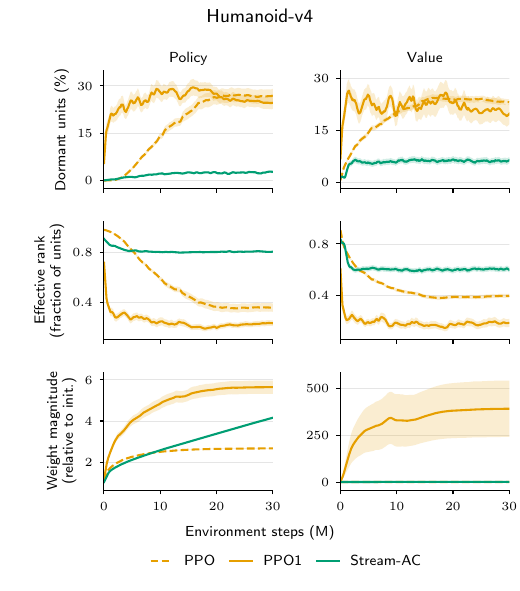}
    \includegraphics[width=0.48\textwidth]{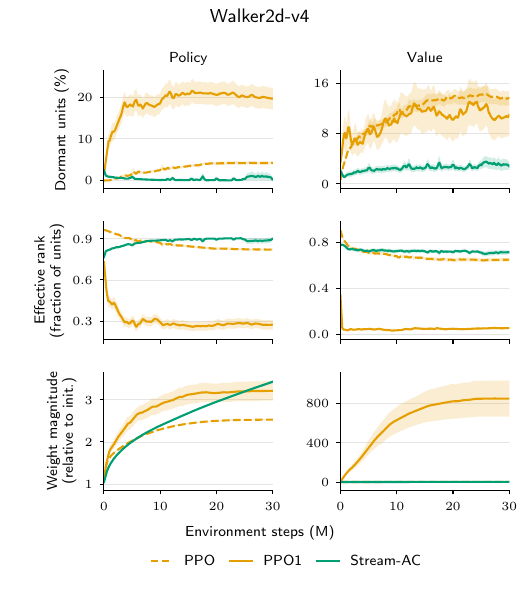}
    \caption{Learning-dynamics diagnostics of PPO, PPO1, and Stream-AC on Ant-v4, HalfCheetah-v4, Humanoid-v4, and Walker2d-v4. Each block shows three statistics---dormant units, effective rank, and average weight magnitude relative to initialization---for the policy network (left column) and the value network (right column) over a $30$M-step budget. Curves are means over $30$ runs per method, and the shaded area represents the standard error.}
    \label{fig:learning_dynamics}
\end{figure}

\clearpage

\section{Full ablation}
\label{appendix:full-ablation}

The grid has seven configurations: the full method and six variants, each removing one ingredient. Stream-AC is ablated on five MuJoCo environments over $5$M time steps and Stream-Q on five Atari games over $25$M time steps. All configurations use the prescribed hyperparameters from Appendix~\ref{appendix:experimental-details}, where each variant differs from the full method in exactly one respect:
\begin{itemize}[leftmargin=1.5em, topsep=0pt, noitemsep]
    \item \textbf{No LayerNorm}: each LayerNorm module is replaced by the identity.
    \item \textbf{No SparseInit}: every linear and convolutional layer uses the default dense initialization instead of the layer-dependent sparsities allocated with global target $s=0.95$. Dense weights follow $U[-1/\sqrt{\texttt{fan\_in}},\, 1/\sqrt{\texttt{fan\_in}}]$; the sparse routine also zeros biases.
    \item \textbf{No StreamingOptimizer}: the update becomes $\vw \leftarrow \vw + \alpha\,\delta_t \vz_t$, corresponding to plain stochastic gradient descent on the eligibility trace at the same step size, with no running maximum and hence no displacement bound.
    \item \textbf{No NormalizeObservation}: raw observations are provided to the network.
    \item \textbf{No RewardScaling}: the raw reward is used to compute the temporal-difference error.
    \item \textbf{No eligibility traces}: $\lambda$ is set to $0$, so the trace reduces to the current gradient and credit no longer propagates backward in time.
\end{itemize}

\begin{figure}[ht]
    \centering
    \includegraphics[width=\textwidth]{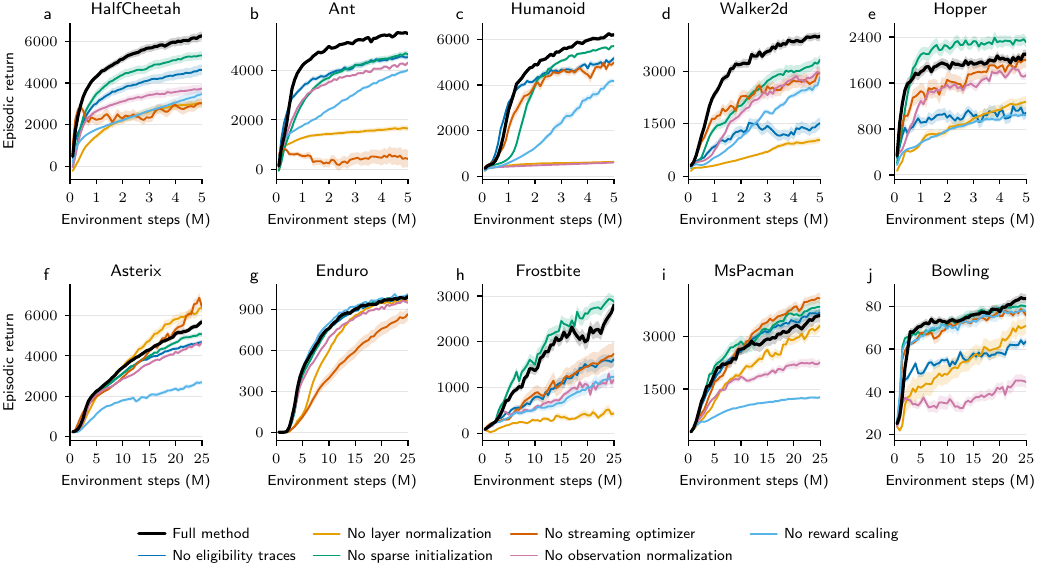}
    \caption{Ablation of Stream-AC and Stream-Q. \textbf{Top row} (\textbf{a}--\textbf{e}): five MuJoCo environments over $5$M time steps. \textbf{Bottom row} (\textbf{f}--\textbf{j}): five Atari games over $25$M time steps. Variants remove one component at a time. Curves are means with standard-error shading over $30$ runs.}
    \label{fig:ablation-full}
\end{figure}

\clearpage

\section{The nonstationary robot vision task details}
\label{appendix:architecture}

For the vision-based heading task, the agent receives histories of proprioceptive observations, actions, and visual features. Each proprioceptive observation is $24$-dimensional, comprising eight joint positions, eight joint velocities, a two-dimensional desired heading vector, and six IMU measurements. We stack the three most recent observations and actions, yielding a $72$-dimensional proprioceptive history and $24$-dimensional action history.

\begin{figure}[th]
    \centering
    \includegraphics[width=\linewidth]{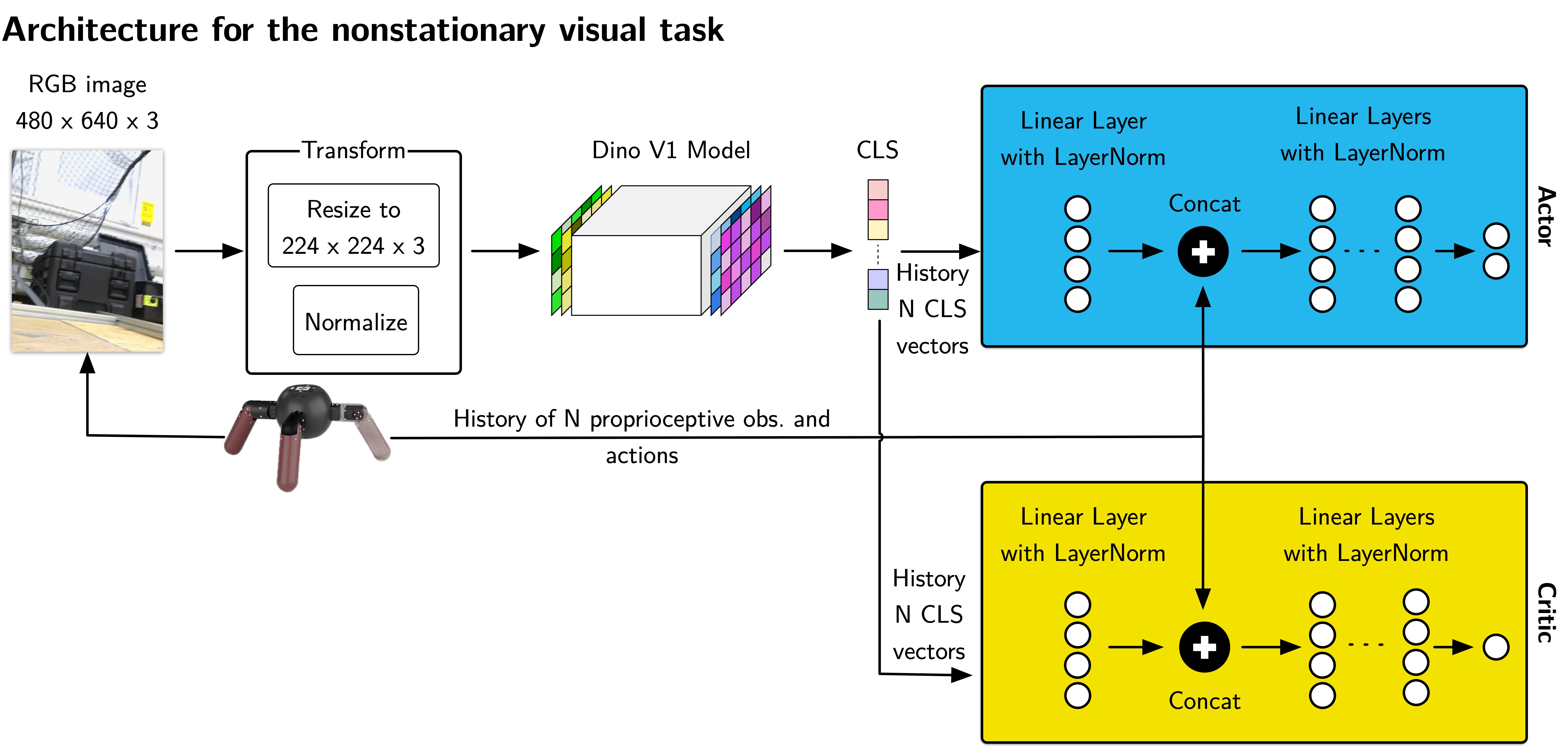}
    \caption{Actor and critic architecture: camera images are encoded by a frozen DINO ViT-S/16 backbone, projected to a lower-dimensional visual embedding, and concatenated with proprioceptive and action history before entering fully connected actor and critic heads.}
    \label{fig:architecture_vision}
\end{figure}

At each time step, the on-board RGB image is resized and center-cropped to $224\times224$ pixels and processed by a frozen DINO ViT-S/16 visual encoder. We extract the $384$-dimensional CLS token and $\ell_2$-normalize it. 
We use the CLS token rather than the spatial patch tokens since it already provides a compact, semantically-rich global summary of the scene. 
In earlier experiments, we also evaluated using the full set of DINO patch tokens rather than the CLS token alone. This achieved comparable final performance, but required more hardware time to learn, likely due to the larger input dimensionality of the visual projection.
The three most recent CLS embeddings are stacked, producing a $1152$-dimensional visual input. The visual history is passed through a learned linear projection from $1152$ to $128$ dimensions, followed by a ReLU and layer normalization. The resulting $128$-dimensional visual embedding is concatenated with the $96$-dimensional proprioceptive and action history.

Both actor and critic use the same MLP after fusion. The resulting $224$-dimensional input is mapped to two $512$-dimensional hidden layers, each followed by layer normalization and LeakyReLU. 
The policy outputs eight continuous actions in $[-1,1]$, corresponding to the eight actuated joints. These normalized actions are subsequently mapped to the physical joint ranges, with $\pm30^\circ$ for the hip joints and $\pm20^\circ$ for the knee joints. The DINO backbone is kept fixed, while the visual projection and actor--critic networks are learned online using Stream-AC.

\clearpage
\section{Learning curves}
\label{appendix:learning-curves}

\begin{figure}[H]
\centering
\includegraphics[width=\textwidth,height=0.8\textheight,keepaspectratio]{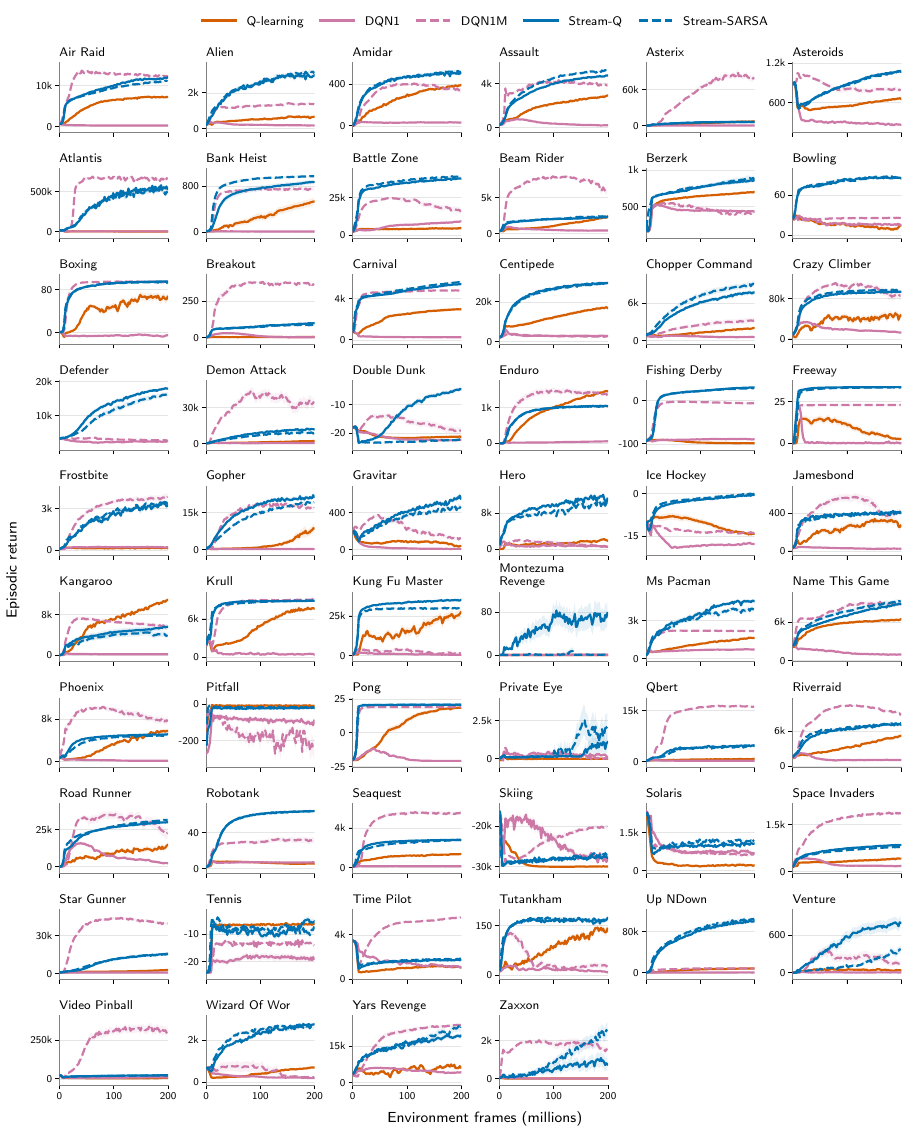}
\caption{\textbf{Atari learning curves across all $58$ games.} Q-learning, DQN1, DQN with a million-transition replay buffer, Stream-Q, and Stream-SARSA are evaluated on $58$ games. Each run's episodic returns are averaged in $100$ nonoverlapping bins of $500{,}000$ agent steps ($2$M frames). All curves are averages over $30$ independent runs, with the shaded area denoting one standard error. Returns are unnormalized, and each panel uses its own vertical scale.}
\end{figure}

\begin{figure}[H]
\centering
\includegraphics[width=\textwidth,height=\textheight,keepaspectratio]{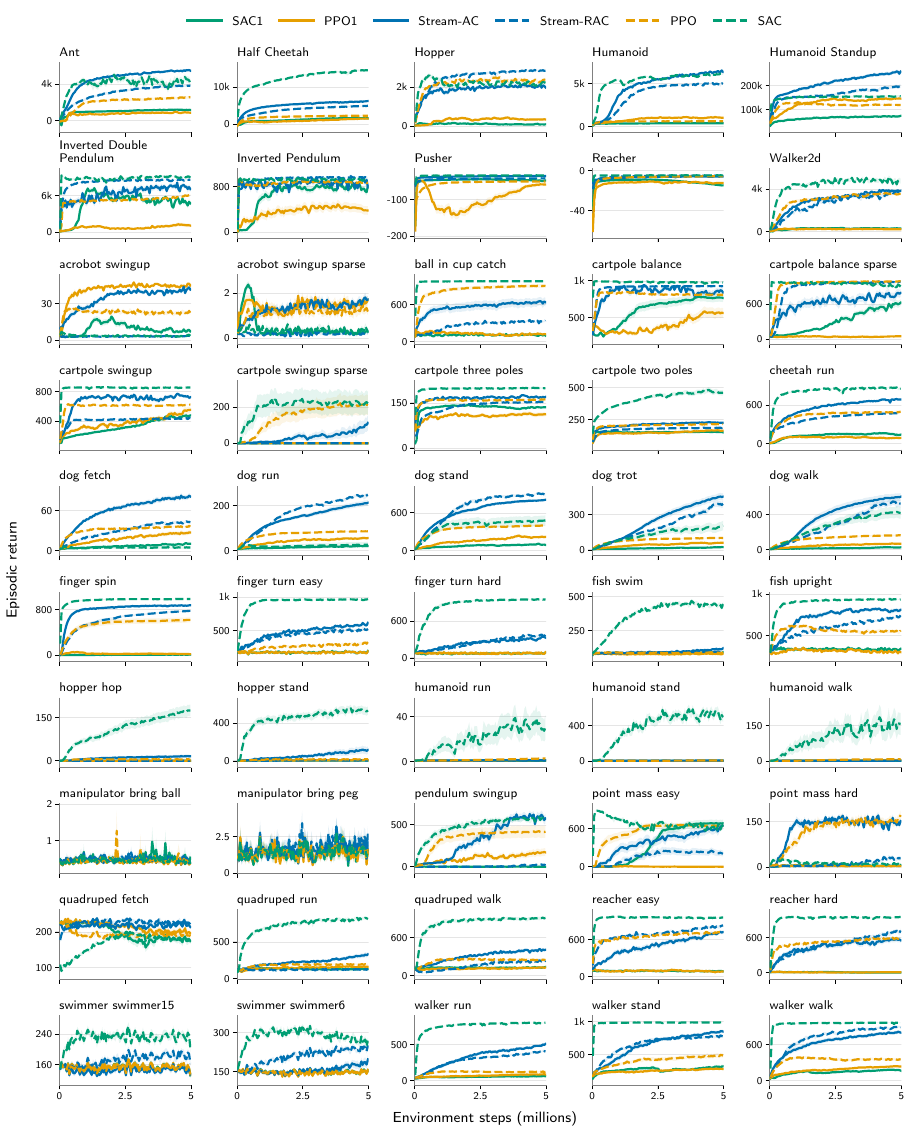}
\caption{\textbf{Learning curves across all continuous-control tasks.} SAC1, PPO1, Stream-AC, Stream-RAC, PPO, and SAC on all $50$ tasks. Returns are averaged within each run over $100$ nonoverlapping bins of $50{,}000$ environment steps. All curves are averages over $30$ independent runs, with the shaded area denoting one standard error. Returns are not normalized, and each panel has its own vertical scale.}
\end{figure}

\begin{figure}[H]
\centering
\includegraphics[width=\textwidth,height=\textheight,keepaspectratio]{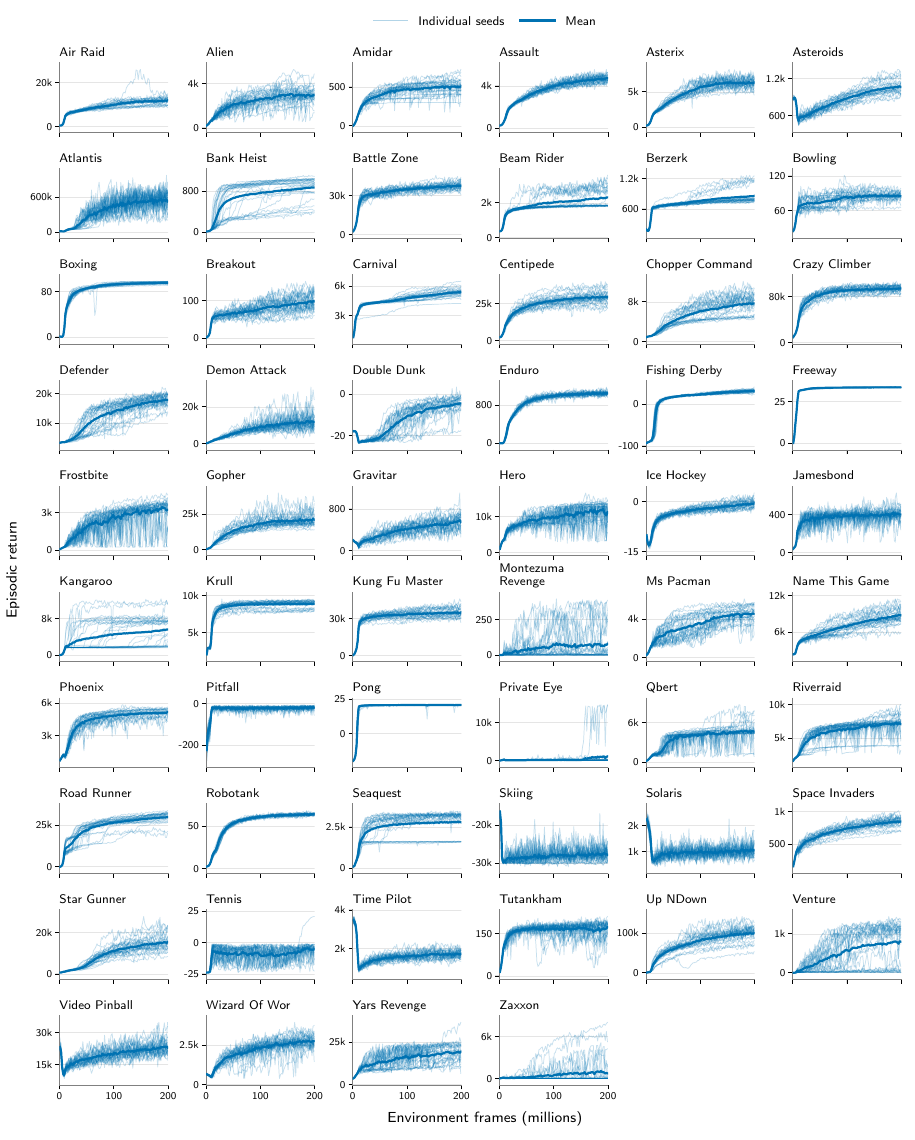}
\caption{\textbf{Stream-Q:  individual learning curves.} Thin translucent lines show the $30$ individual runs, and the thick line shows their average. Each run is averaged in $500{,}000$-agent-step bins ($2$M frames). Returns are unnormalized and vertical scales are panel-specific.}
\end{figure}

\begin{figure}[H]
\centering
\includegraphics[width=\textwidth,height=\textheight,keepaspectratio]{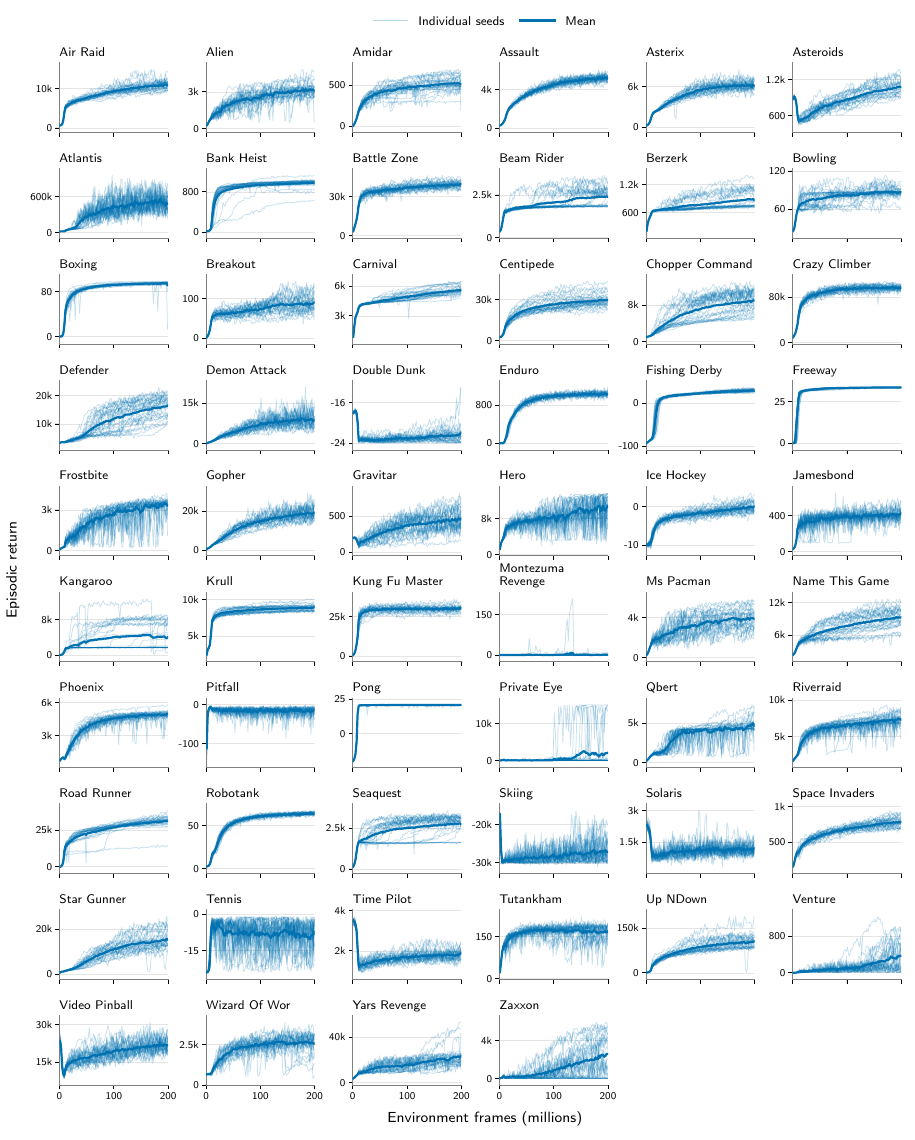}
\caption{\textbf{Stream-SARSA:  individual learning curves.} Thin translucent lines show the $30$ individual runs, and the thick line shows their average. Each run is averaged in $500{,}000$-agent-step bins ($2$M frames). Returns are unnormalized and vertical scales are panel-specific.}
\end{figure}

\begin{figure}[H]
\centering
\includegraphics[width=\textwidth,height=\textheight,keepaspectratio]{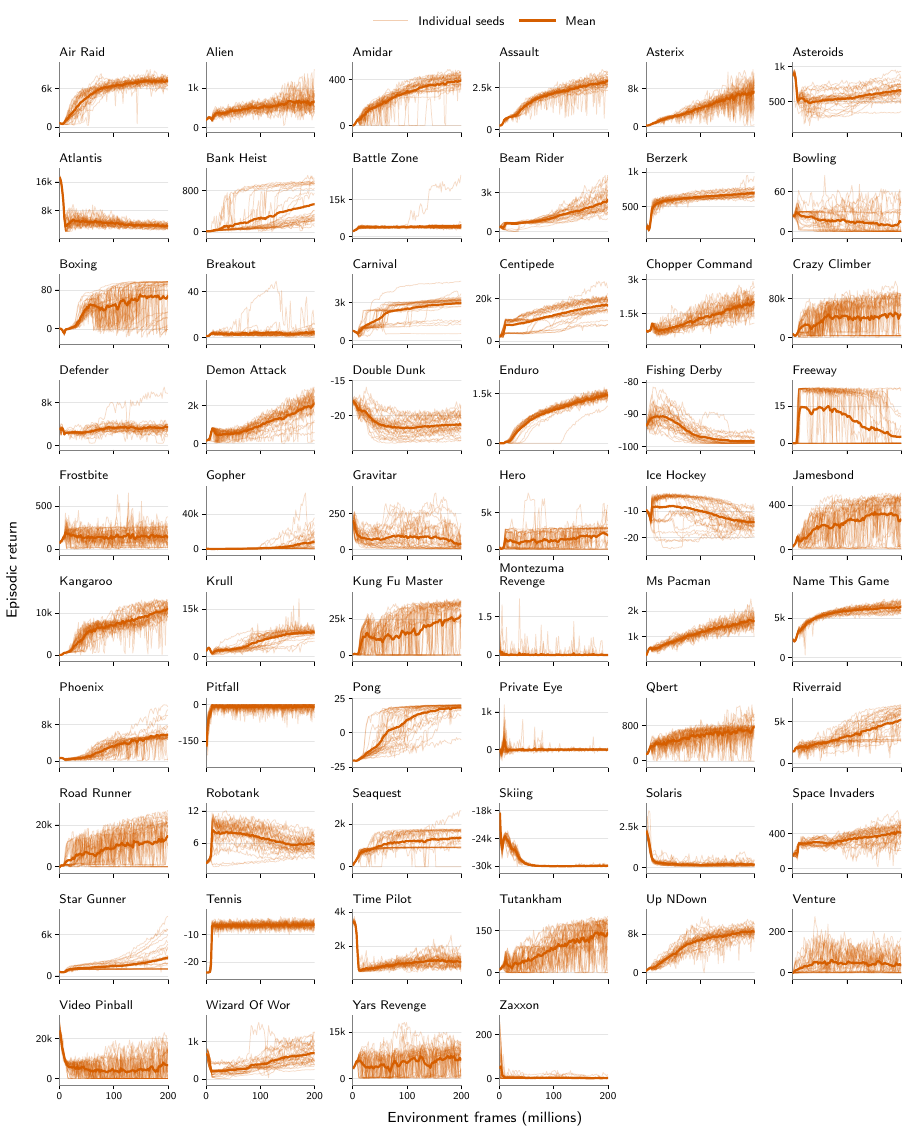}
\caption{\textbf{Q-learning: individual learning curves.} Thin translucent lines show the $30$ individual runs, and the thick line shows their average. Each run is averaged in $500{,}000$-agent-step bins ($2$M frames). Returns are unnormalized and vertical scales are panel-specific.}
\end{figure}

\begin{figure}[H]
\centering
\includegraphics[width=\textwidth,height=\textheight,keepaspectratio]{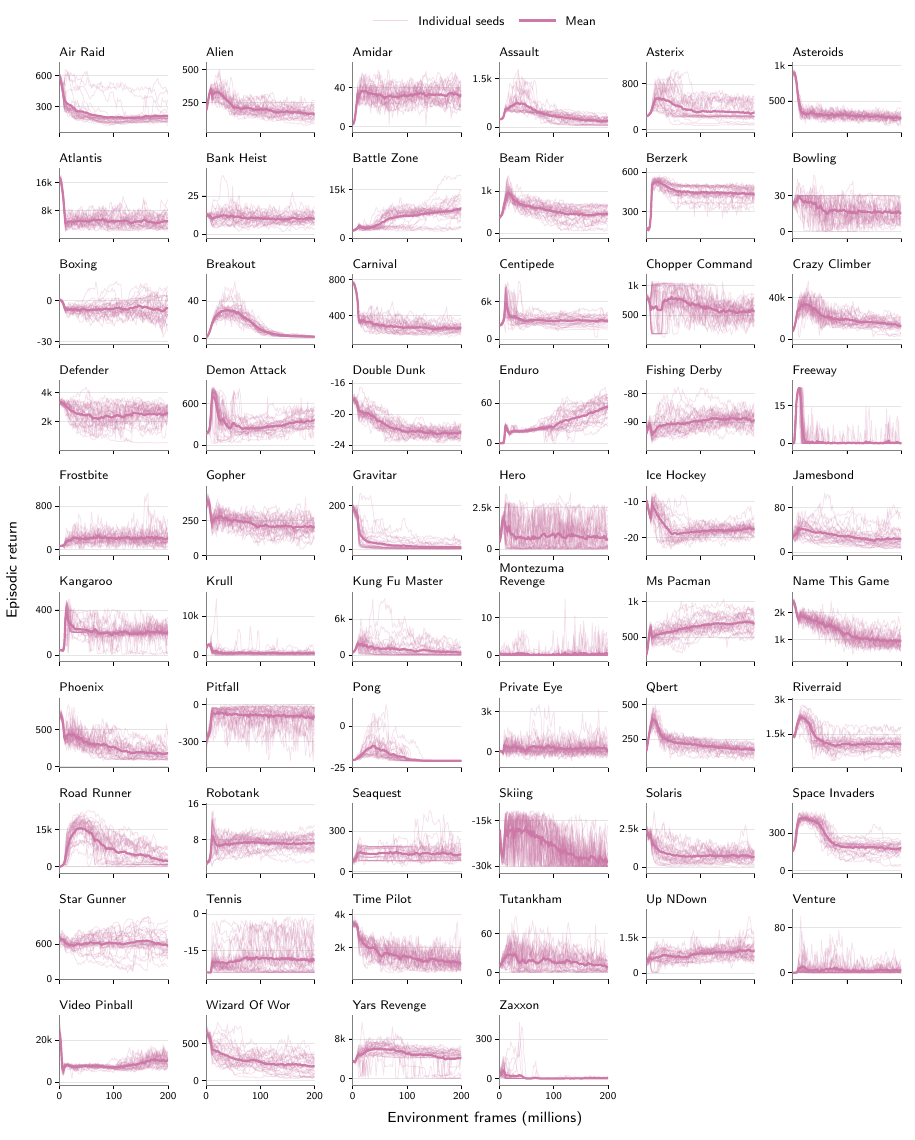}
\caption{\textbf{DQN1: individual learning curves.} Thin translucent lines show the $30$ individual runs, and the thick line shows their average. Each run is averaged in $500{,}000$-agent-step bins ($2$M frames). Returns are unnormalized and vertical scales are panel-specific.}
\end{figure}

\begin{figure}[H]
\centering
\includegraphics[width=\textwidth,height=\textheight,keepaspectratio]{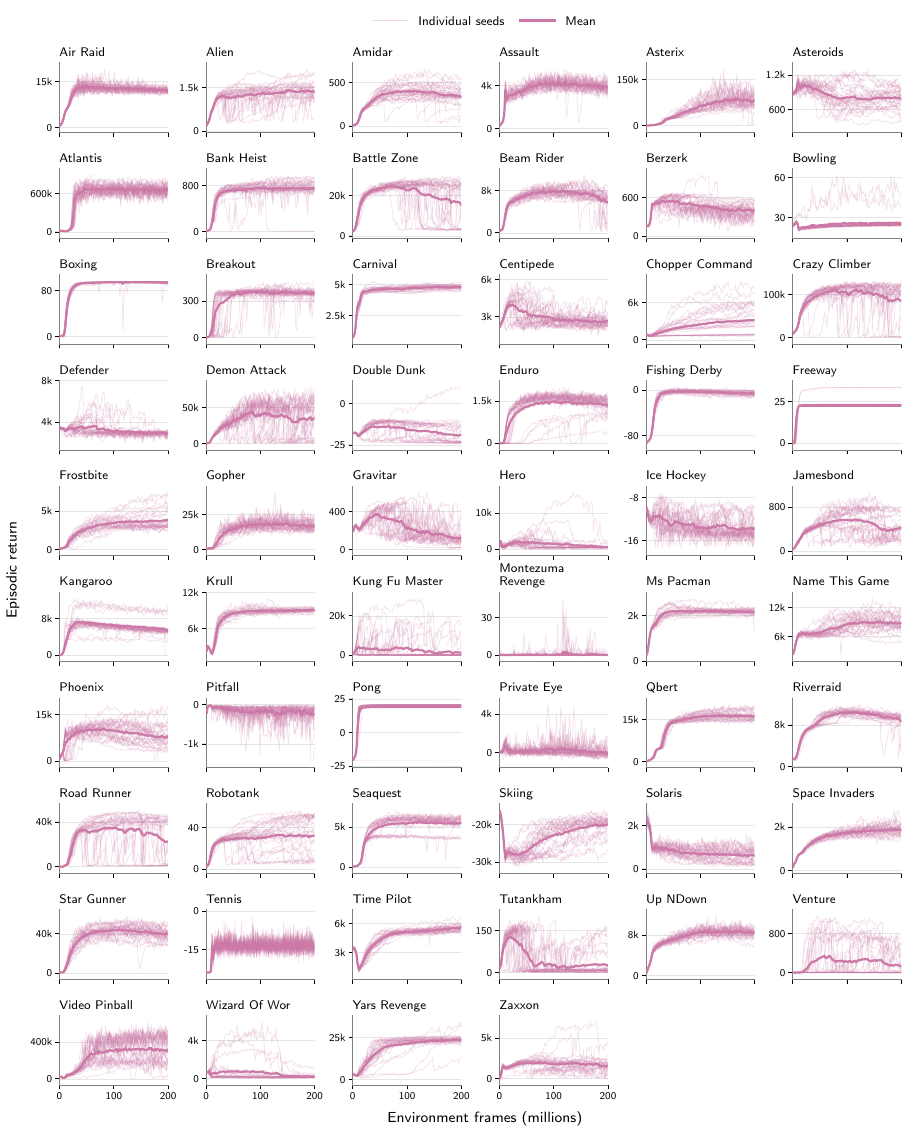}
\caption{\textbf{DQN: individual learning curves.} Thin translucent lines show the $30$ individual runs, and the thick line shows their average. Each run is averaged in $500{,}000$-agent-step bins ($2$M frames). Returns are unnormalized and vertical scales are panel-specific.}
\end{figure}

\begin{figure}[H]
\centering
\includegraphics[width=\textwidth,height=\textheight,keepaspectratio]{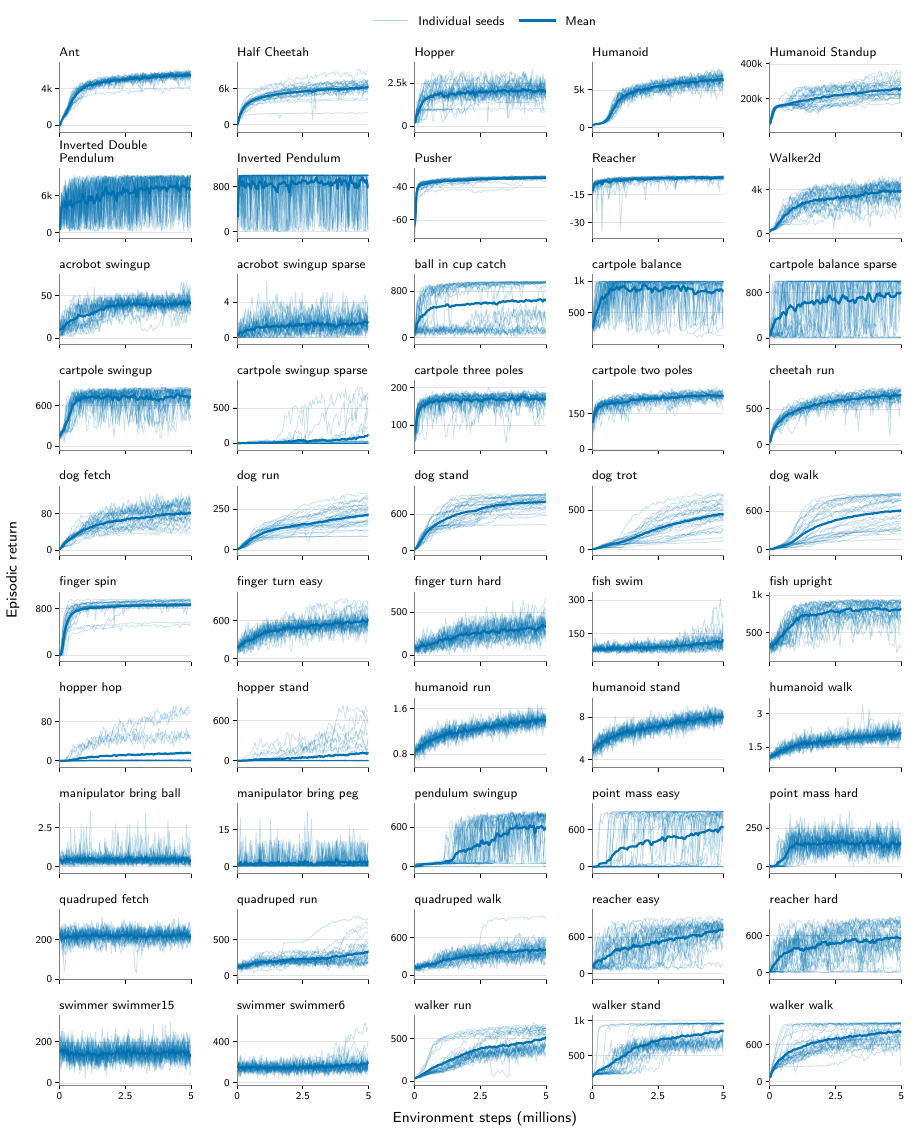}
\caption{\textbf{Stream-AC: individual learning curves across continuous control.} All $50$ tasks, with $30$ runs per task. Thin translucent lines show the individual runs and the thick line shows their average. Each run is averaged in $50{,}000$-step bins, without additional smoothing or extrapolation. Returns are unnormalized and vertical scales are panel-specific.}
\end{figure}

\begin{figure}[H]
\centering
\includegraphics[width=\textwidth,height=\textheight,keepaspectratio]{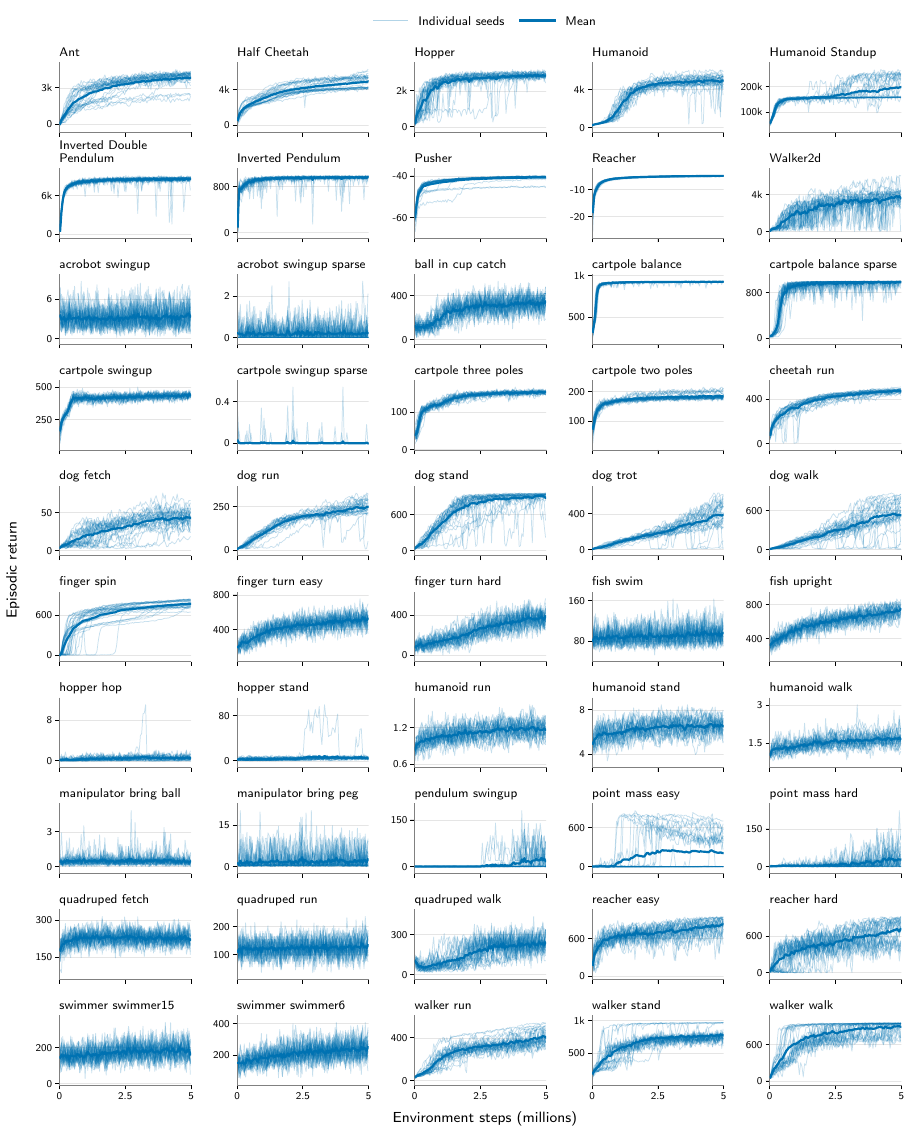}
\caption{\textbf{Stream-RAC: individual learning curves across continuous control.} All $50$ tasks, with $30$ runs per task. Thin translucent lines show the individual runs and the thick line shows their average. Each run is averaged in $50{,}000$-step bins, without additional smoothing or extrapolation. Returns are unnormalized and vertical scales are panel-specific.}
\end{figure}

\begin{figure}[H]
\centering
\includegraphics[width=\textwidth,height=\textheight,keepaspectratio]{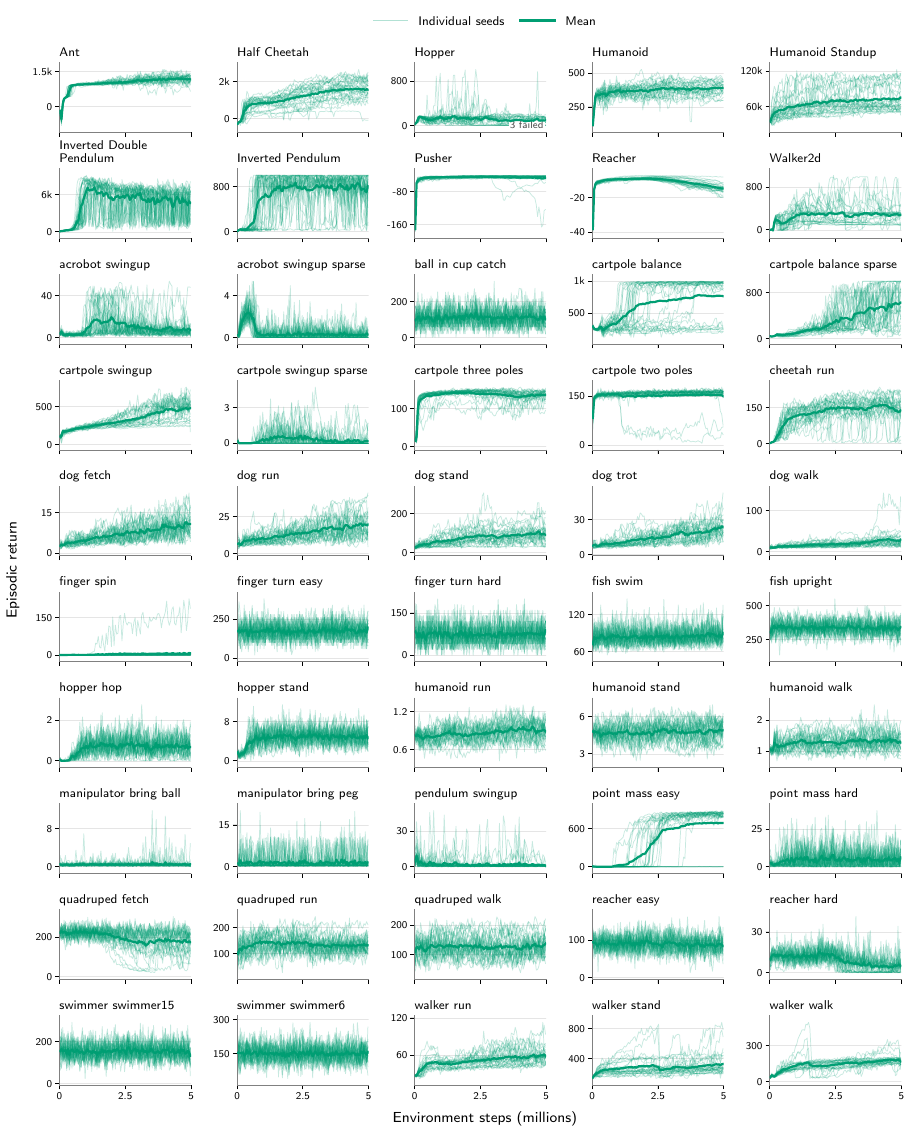}
\caption{\textbf{SAC1: individual learning curves across continuous control.} All $50$ tasks, with $30$ runs per task. Thin translucent lines show the individual runs and the thick line shows their average. Each run is averaged in $50{,}000$-step bins, without additional smoothing or extrapolation. Returns are unnormalized and vertical scales are panel-specific.}
\end{figure}

\begin{figure}[H]
\centering
\includegraphics[width=\textwidth,height=\textheight,keepaspectratio]{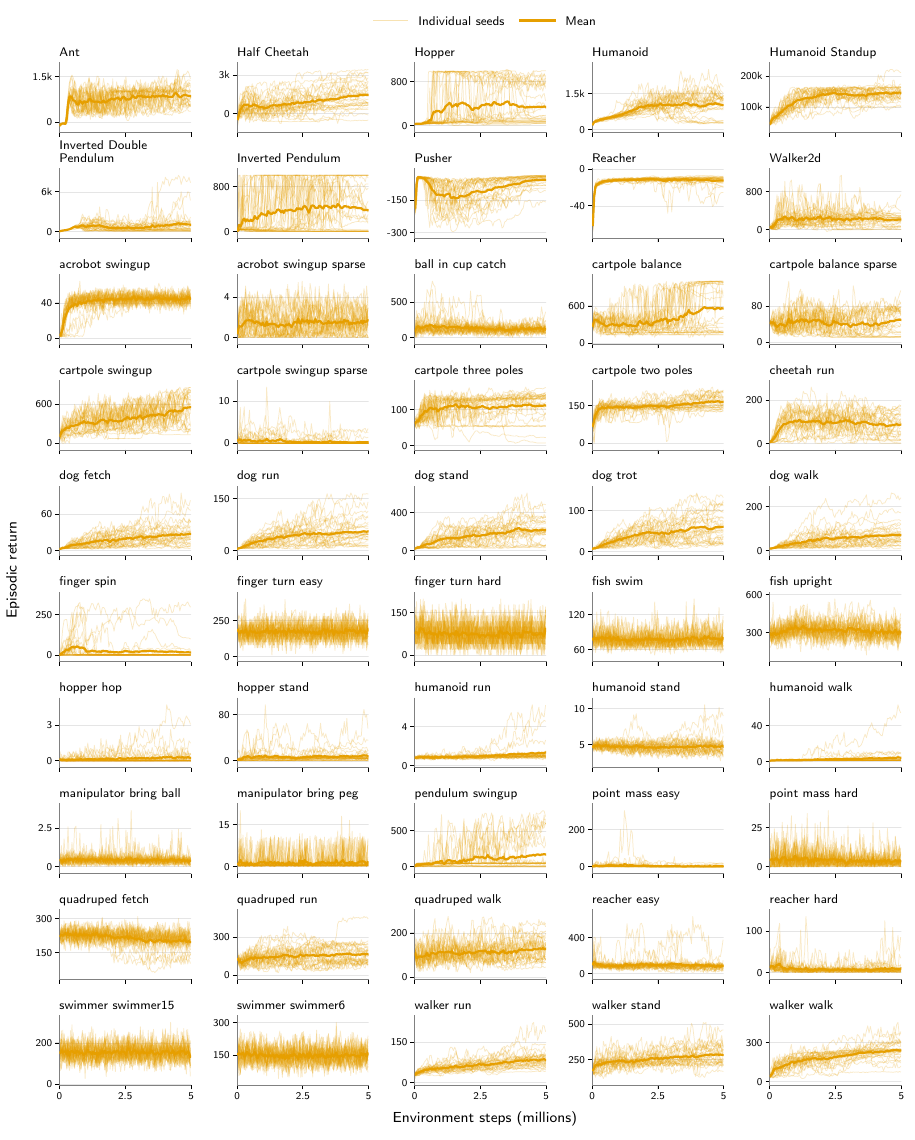}
\caption{\textbf{PPO1: individual learning curves across continuous control.} All $50$ tasks, with $30$ runs per task. Thin translucent lines show the individual runs and the thick line shows their average. Each run is averaged in $50{,}000$-step bins, without additional smoothing or extrapolation. Returns are unnormalized and vertical scales are panel-specific.}
\end{figure}

\begin{figure}[H]
\centering
\includegraphics[width=\textwidth,height=\textheight,keepaspectratio]{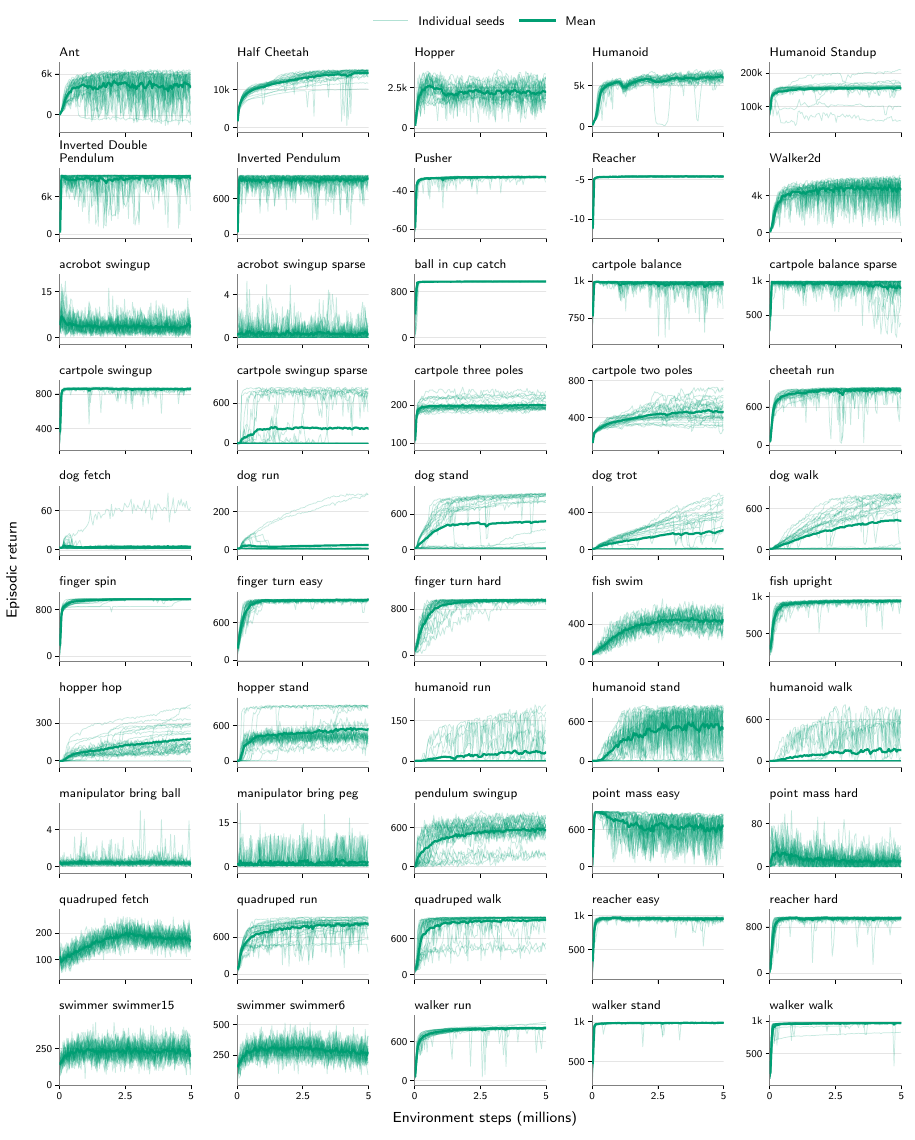}
\caption{\textbf{SAC: individual learning curves across continuous control.} All $50$ tasks, with $30$ runs per task. Thin translucent lines show the individual runs and the thick line shows their average. Each run is averaged in $50{,}000$-step bins, without additional smoothing or extrapolation. Returns are unnormalized and vertical scales are panel-specific.}
\end{figure}

\begin{figure}[H]
\centering
\includegraphics[width=\textwidth,height=\textheight,keepaspectratio]{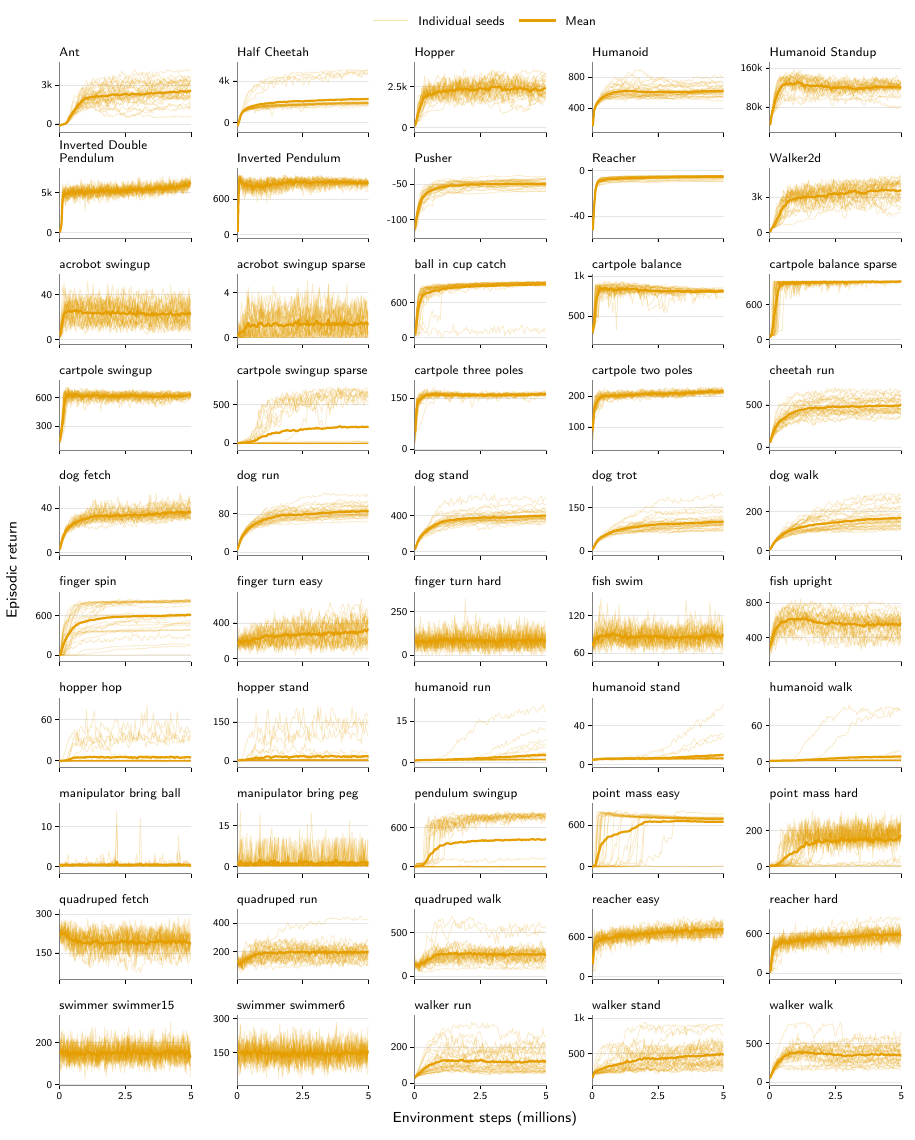}
\caption{\textbf{PPO: individual learning curves across continuous control.} All $50$ tasks, with $30$ runs per task. Thin translucent lines show the individual runs and the thick line shows their average. Each run is averaged in $50{,}000$-step bins, without additional smoothing or extrapolation. Returns are unnormalized and vertical scales are panel-specific.}
\end{figure}

\end{document}